%% file: main.tex
\documentclass[letterpaper,twocolumn,10pt]{article}
\usepackage{usenix-2020-09}

\usepackage{tikz}
\usepackage{amsmath}
\usepackage{wasysym}
\usepackage{color, pifont, comment}
\usepackage{xspace}
\usepackage{multirow}
\usepackage{algorithm}
\usepackage{algpseudocode}
\usepackage{tabularx}
\usepackage{booktabs}
\usepackage{amsthm, amssymb}
\usepackage{enumerate}
\usepackage{enumitem}
\usepackage{cases}
\usepackage[tight]{subfigure}
\usepackage[export]{adjustbox}
\usepackage{multirow}
\usepackage{stmaryrd}
\usepackage{colortbl}
\usepackage{xurl}
\usepackage[compact]{titlesec}
\usepackage[subtle,tracking=normal]{savetrees}

\newcounter{isusenixversion}
\DeclareMathOperator*{\argmax}{arg\,max}

\usepackage{eqparbox}
\newdimen{\algindent}
\algnewcommand\LeftComment[2]{%
\hspace{#1\algindent}$\triangleright$ \eqparbox{COMMENT}{#2} \hfill %
}

\newcommand{\ie}{{\it i.e.,}\xspace}
\newcommand{\eg}{{\it e.g.,}\xspace}

\newcommand{\vs}{{\it vs.}\xspace}
\newcommand{\etal}{{\it et al.}\xspace}
\newcommand{\descr}[1]{\vspace{.02in} \noindent\textbf{#1}}
\newcommand{\textcaret}{\^}
\newcommand{\persite}{per-site}

\newcommand{\fla}{FL author}

\newcommand{\appref}{App.}
\newcommand{\inittimes}{10}
\newcommand{\adhigh}{Ad Highlighter}
\newcommand{\adgraph}{AdGraph}
\newcommand{\siteruletotal}{933}
\newcommand{\sitetotal}{1042}

\newcommand{\ruletotal}{361}
\newcommand{\realtimeruletotal}{464}
\newcommand{\supportevaltotal}{272}

\newcommand{\sstimeavg}{49} %
\newcommand{\autofrravg}{13} %
\newcommand{\autofrpersite}{1.6} %
\newcommand{\task}{Task}
\newcommand{\fulldata}{\textit{Full-W09-Dataset}}
\newcommand{\wdata}{\textit{W09-Dataset}}
\newcommand{\sampledata}{\textit{Sampled-100-Dataset}}
\newcommand{\ats}{A\&T}
\newcommand{\scriptusedby}{Script-used-by}
\newcommand{\noise}{$\epsilon$}
\newcommand{\actionspace}{$\mathcal{A}_H$}
\newcommand{\rewardfunc}{$\mathcal{R}_F$}
\newcommand{\frglower}{filter rule generation}
\newcommand{\bestbin}{operating point}

\newcommand{\fl}{filter list}
\newcommand{\opsrc}{open-source}

\newcommand{\humanloop}{human-in-the-loop}
\newcommand{\finergrain}{finer-grain}

\newcommand{\snapshotcol}{1}
\newcommand{\autofrcol}{2}
\newcommand{\confirmcol}{3}
\newcommand{\eltopfivekcol}{4}
\newcommand{\reapplycol}{5} 
\newcommand{\aggonecol}{6} 
\newcommand{\aggthreecol}{7} 
\newcommand{\eltoptenkcol}{8}

\newtheoremstyle{taskstyle}%
  {\topsep}%
  {\topsep}%
  {}%
  {0pt}%
  {\bfseries}%
  {. }%
  { }%
  {\thmname{#1}\thmnumber{ #2}\textnormal{\thmnote{: \textit{#3}}}}

\theoremstyle{taskstyle}
\newtheorem{challenge}{Task}[]

\definecolor{brandeisblue}{rgb}{0.0, 0.44, 1.0}

\newcommand{\major}[1]{{\color{black}#1}}
\newcommand{\tool}{{AutoFR}}
\newcommand{\toolcontrol}{{AutoFR}}
\newcommand{\toolwild}{{AutoFR-L}}

\ifthenelse{\value{isusenixversion}>0}
{
    \pagestyle{empty} %
    \newcommand{\inusenixversion}[1]{#1}
    \newcommand{\inarxivversion}[1]{}
}
{
    \newcommand{\inusenixversion}[1]{}
    \newcommand{\inarxivversion}[1]{#1}
}

\begin{document}
\date{}

\title{\Large \bf \tool{}: Automated Filter Rule Generation for Adblocking}

\author{
{\rm Hieu Le\textsuperscript{*} \ \ \
     Salma Elmalaki\textsuperscript{*} \ \ \
     Athina Markopoulou\textsuperscript{*} \ \ \
     Zubair Shafiq\textsuperscript{\textdagger}
}
\\
\\
{\rm \textsuperscript{*}University of California, Irvine} \ \ \
{\rm \textsuperscript{\textdagger}University of California, Davis}
\inarxivversion{
\\
\\
{\rm \textcolor{Blue}{This is an extended version of our paper that appears in USENIX Security 2023}}
}
} %

\maketitle

\begin{abstract}
Adblocking relies on filter lists, which are manually curated and maintained by a community of filter list authors. 
Filter list curation is a laborious process that does not scale well to a large number of sites or over time. 
In this paper, we introduce \tool{}, a reinforcement learning framework to fully automate the process of filter rule creation and evaluation for sites of interest.
We design an algorithm based on multi-arm bandits to generate filter rules that block ads while controlling the trade-off between blocking ads and avoiding visual breakage. 
We test \tool{} on thousands of sites and we show that it is efficient: it takes only a few minutes to generate filter rules for a site of interest.
\tool{} is effective: it generates filter rules that can block 86\% of the ads, as compared to 87\% by EasyList, while achieving comparable visual breakage. 
Furthermore, \tool{} generates filter rules that generalize well to new sites.
We envision that \tool{} can assist the adblocking community in \frglower{} at scale.
\end{abstract}

\input{introduction}
\input{relatedwork}
\input{formulation}

\input{implementation}
\input{evaluation}
\input{conclusion}

\section*{Acknowledgments}
This work is supported in part by the National Science Foundation under award numbers 1956393, 1900654, 1815666, 2051592, 2102347, 2103038, 2103439, 2105084, and 2138139. We would like to thank the USENIX Security reviewers for their feedback, which helped to improve the paper. We would like to thank Stelios Stavroulakis for his help during the early stages of this work. Lastly, special thanks to the filter list community, including Ryan Brown, Arthur Kawa, and Peter Lowe, who provided valuable insight into the human process of creating and maintaining filter rules.

\bibliographystyle{abbrv}
\bibliography{master, online}

\ifthenelse{\value{isusenixversion}>0}
{
}
{
    \newpage
    \appendix
    \input{appendix}
}
\end{document}

%% file: introduction.tex
\section{Introduction}
\label{sec:introduction}

Adblocking is widely used today to improve the security, privacy, performance, and browsing experience of web users. 
Twenty years after the introduction of the first adblocker in 2002, the number of web users who use some form of adblocking now exceeds 42\% \cite{adblocker-users}.  
Adblocking primarily relies on \fl{s} (\eg{} EasyList~\cite{easylist}) that are manually curated based on crowd-sourced user feedback by a small community of \fl{} (FL) authors. 
There are hundreds of different adblocking \fl{s} that target different platforms and geographic regions \cite{AllFilterLists}. 
It is well-known that the \fl{} curation process is slow and error-prone \cite{AlrizhaErrors}, and requires significant continuous effort by the \fl{} community to keep them up-to-date \cite{cvinspectorndss}.

The research community is actively working on machine learning (ML) approaches to assist with filter rule generation  \cite{BhagavatulaResourceFiltering,Gugelmann15,alex2019generation} or to build models to replace \fl{s} altogether \cite{iqbal2018adgraph,siby22webgraph,YangWtaGraph,abi2020percival}.
There are two key limitations of prior ML-based approaches. 
First, existing ML approaches are supervised as they rely on human feedback and/or existing \fl{s} (which are also manually curated) for training.
This introduces a circular dependency between these supervised ML models and \fl{s} --- the training of models relies on the very \fl{s} (and humans) that they aim to augment or replace.
Second, existing ML approaches do not explicitly consider the trade-off between blocking ads and avoiding breakage. 
An over-aggressive adblocking approach might block all ads on a site but may block legitimate content at the same time. 
Thus, despite recent advances in ML-based adblocking, \fl{s} remain defacto in adblocking.

\begin{figure}[!t]
    \centering
    \subfigure[\textbf{Filter List Authors' (Human) Workflow.} 
    How \fl{} authors create filter rules for a site $\ell$: (1) they select a network request caused by the site; (2) they create a filter rule and apply it on the site; (3) they visually inspect whether it blocked ads without breakage; (4) they repeat the process if necessary for other network requests; and (5) they stop when they have crafted filter rules that can block all/most ads for the site without causing significant breakage.
    ]
    {\includegraphics[width=1\linewidth]{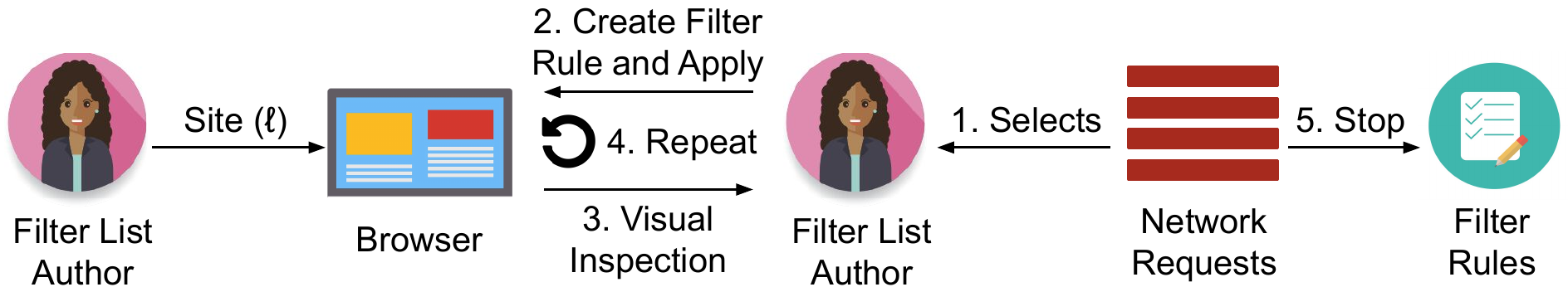}
        \label{fig:fla}
	}
	\subfigure[\textbf{\tool{} (Automated) Workflow.} \tool{} automates these steps as follows: (1) the agent selects an action (\ie{} filter rule) following a policy; (2) it applies the action on the environment; (3) the environment returns a reward, used to update the action space; (4) the agent repeats the process if necessary; and (5) the agent stops when a time limit is reached, or no more actions are available to be explored. The human filter list author only provides a site $\ell$ and configurations (\eg{} threshold $w$ and hyper-parameters).]{
		 \includegraphics[width=1\linewidth]{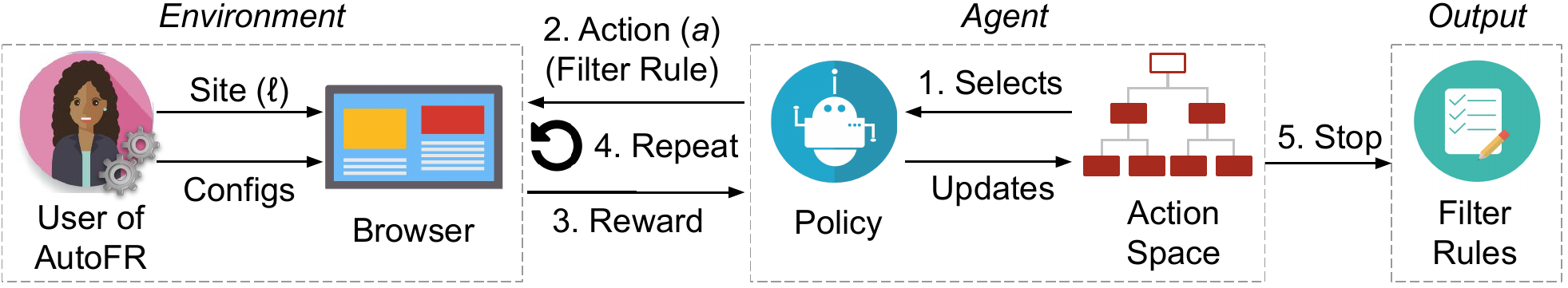}
        \label{fig:formulation}
	}
\vspace{-5pt}

\caption[width=1\linewidth]{\small \major{\tool{} automates the steps taken by \fla{s} to generate filter rules for a particular site. \fla{s} can configure the \tool{} parameters but no longer perform the manual work. Once rules are generated by \tool{}, it is up to the \fla{s} to decide when and how to deploy the rules to end-users.}}
\vspace{-5pt}
\label{fig:fla-and-formulation}
\end{figure}

Fig.~\ref{fig:fla} illustrates the workflow of a \fla{} \major{for creating rules for a particular site:} 
(1) select a network request to block;
(2) design a filter rule that corresponds to this request and apply it on the site; 
(3) visually inspect the page to evaluate if the filter rule blocks ads and/or causes breakage and;
(4) repeat for other network requests and rules; since modern sites are highly dynamic, and often more so in response to adblocking~\cite{cvinspectorndss,AlrizhaErrors,Zhu_DisruptAA,chen2021jssignatures}, the \fla{} usually revisits the site multiple times to ensure the rule remains effective;
and (5) stop when a set of filter rules can adequately block ads without causing breakage. 

We ask the question: \textit{how can we minimize the manual effort of \fla{s} by automating the process of generating and evaluating adblocking filter rules?} 
We propose \tool{} to automate each of the aforementioned steps, as illustrated in Fig.~\ref{fig:formulation},
and \major{we make the following contributions.}

First, we formulate the filter rule generation problem within a reinforcement learning (RL) framework, which enables us to efficiently create and evaluate good candidate rules, as opposed to brute force or random selection.
We focus on URL-based filter rules that block ads, a popular and representative type of rules that can be visually audited.   
An important component, which replaces the visual inspection, is the detection of ads (through a perceptual classifier, \adhigh{}~\cite{futureofadblocking}) and of visual breakage (through JavaScript [JS] for images and text) on a page. 
We design a reward function that combines these metrics to enable explicit control over the trade-off between blocking ads and avoiding breakage. 

Second, we design and implement \tool{} to train the RL agent by accessing sites in a controlled realistic environment. 
It creates rules for a site in under two minutes, which is crucial for scalability. 
We deploy and evaluate \tool{}'s efficient implementation on Top--10K websites, 
and we find that the filter rules generated by \tool{} block 86\% of the ads.
We also find that they \major{generalize well to new sites}, \eg blocking 80\% of the ads on the Top 5K--10K sites. 
The effectiveness of the \tool{} rules is overall comparable to EasyList in terms of blocking ads and visual breakage. 
Thus, we envision that the adblocking community will use \tool{} to automatically generate and update filter rules at scale. 

The rest of our paper is organized as follows. 
Sec.~\ref{sec:relatedwork} provides background and related work.
Sec.~\ref{sec:algorithm} formalizes the problem of \frglower{}, including the human process, the formulation as an RL problem, and our particular multi-arm bandit algorithm for solving it.
Sec.~\ref{sec:autofrg-tool} presents our implementation of the \tool{} framework. 
Sec.~\ref{sec:eval} provides its evaluation on the Top--10K sites. 
Sec.~\ref{sec:conclusion} concludes the paper. 
\inarxivversion{The appendices provide additional details and results.}

%% file: relatedwork.tex
\section{Background \& Related Work}
\label{sec:relatedwork}

\descr{Filter Rules.} Adblockers have relied on \fl{s} since their inception.
The first adblocker in 2002, a Firefox extension, allowed users to specify custom filter rules to block resources (e.g., images) from a particular domain or URL path~\cite{mozdev}. 
There are different types of filter rules. 
The most popular type is {\em URL-based filter rules}, which block network requests to provide performance and privacy benefits~\cite{WhoFilters}.  
Other types of filter rules are element-hiding rules (hide HTML elements) and JS-based rules (stop JS execution). 
\inarxivversion{\appref~\ref{app:filter-list-over-time} provides a longitudinal analysis and discussion of widely used filter rules.}
\major{Filter rules can also be \persite{} (\ie{} they are only allowed to trigger for particular sites) or treated as global rules (\ie{} allowed to trigger for any sites).
Popular \fl{s}, such as EasyList, support these rules. Per-site rules are denoted with the ``\$domain'' option in EasyList.
This paper focuses on URL-based, \persite{} rules.}

\descr{Filter Lists and their Curation.} 
Since it is non-trivial for lay web users to create filter rules, several efforts were established to curate rules for the broader adblocking community.
Specifically, rules are curated by \fl{} (FL) authors based on informal crowd-sourced feedback from users of adblocking tools. 
\inarxivversion{As elaborated in \appref~\ref{app:filter-list-over-time}, there is now a rich ecosystem of thousands of different \fl{s} focused on blocking ads, trackers, malware, and other unwanted web resources.}
\inusenixversion{There is now a rich ecosystem of thousands of different \fl{s} focused on blocking ads, trackers, malware, and other unwanted web resources.}
EasyList \cite{easylist} is the most widely used adblocking \fl{}.
Started in 2005 by Rick Petnel, it is now maintained by a small set of \fla{s} and has 22 language-specific versions. 
An active EasyList community provides feedback to \fla{s} on its official forum and GitHub.

The research community has looked into the \fl{} curation process to investigate its effectiveness and pain-points~\cite{WhoFilters,cvinspectorndss,walls-acceptable-ads,AlrizhaErrors}.
Snyder \etal{}~\cite{WhoFilters} studied EasyList's  evolution and showed that it needs to be frequently updated (median update interval of 1.12 hours) because of the dynamic nature of online advertising and efforts from advertisers to evade filter rules. 
They found that it has grown significantly over the years, with 124K+ rule additions and 52K+ rule deletions over the last decade.
Alrizah \etal{}~\cite{AlrizhaErrors} showed that EasyList's curation, despite extensive input from the community, is prone to errors that result in missed ads (false negatives) and over-blocking of legitimate content (false positives). 
They concluded that most errors in EasyList can be attributed to mistakes by \fla{s}. 
We elaborate further on the challenges of filter rule generation in Sec.~\ref{sec:fla-workflow}.

\descr{Machine Learning for Adblocking.} Motivated by these challenges, prior work has explored using machine learning (ML) to assist with \fl{} curation or replace it altogether. 

One line of prior work aims to develop ML models to automatically generate filter rules for blocking ads \cite{BhagavatulaResourceFiltering,Gugelmann15,alex2019generation}. 
Bhagavatula~\etal{}~\cite{BhagavatulaResourceFiltering} trained supervised ML classifiers to detect advertising URLs. 
Similarly, Gugelmann~\etal{}~\cite{Gugelmann15} trained supervised ML classifiers to detect advertising and tracking domains. 
\major{Sjosten~\etal{}~\cite{alex2019generation} is the closest related to our work. First, they trained a hybrid perceptual and web execution classifier to detect ad images~\cite{PageGraph}. 
Second, they generated adblocking filter rules by first identifying the URL of the script responsible for retrieving the ad and then simply using the effective second-level domain (eSLD) and path information of the script as a rule (similar to Table~\ref{tab:network-filter-rule-syntax} row 3). We found that 99\% of rules that they \opsrc{d} had paths. However, this overreliance on rules with paths makes them brittle and easily evaded with minor changes~\cite{cvinspectorndss}. Furthermore, the design of these rules did not automatically consider potential breakage.}  

Another line of prior work, instead of generating filter rules, trains ML models to automatically detect and block ads \cite{iqbal2018adgraph,siby22webgraph,YangWtaGraph,abi2020percival,futureofadblocking,Sentinel}.
\adgraph{}~\cite{iqbal2018adgraph}, WebGraph~\cite{siby22webgraph}, and WTAGraph~\cite{YangWtaGraph} represent web page execution information as a graph and then train classifiers to detect advertising resources. 
\adhigh{} \cite{futureofadblocking}, Sentinel \cite{Sentinel}, and PERCIVAL \cite{abi2020percival} use computer vision techniques to detect ad images. 
These efforts do not generate filter rules but instead attempt to replace \fl{s} altogether.

While promising, existing ML-based approaches have not seen any adoption by adblocking tools. 
Our discussions with the adblocking community have revealed a healthy skepticism of replacing \fl{s} with ML models due to performance, reliability, and explainability concerns. 
On the performance front, the overheads of feature instrumentation and running ML pipelines at run-time are non-trivial and almost negate the performance benefits of adblocking~\cite{moonshot}. 
On the reliability front, concerns about the accuracy and brittleness of ML models in the wild \cite{Sentinel,alex2019generation,abi2020percival}, combined with a lack of explainability~\cite{tramer2019adversarial}, have hampered their adoption.
In short, it seems unlikely that \fl{s} will be replaced by ML models any time soon, and filter rules remain crucial for adblocking tools.

\descr{ML-assisted FL Curation.} There is, however, optimism in using ML-based approaches to assist with \textit{maintenance} of \fl{s}.
For example, Brave~\cite{alex2019generation}, Adblock Plus \cite{Sentinel}, and the research community \cite{cvinspectorndss} have been using ML models to assist \fla{s} in prioritizing filter rule updates. 
However, they have two main limitations. 
\major{First, they rely on \fl{s}, such as EasyList, for training their supervised ML models causing a \textit{circular dependency}: a supervised model is only as good as the ground-truth data it is trained on.}
This also means that the adblocking community has to continue maintaining both ML models as well as \fl{s}. 
Second, existing ML approaches do not explicitly consider the trade-off between blocking ads and avoiding breakage. 
An over-aggressive adblocking approach might block all ads on a site but may block legitimate content at the same time. 
It is essential to control this trade-off for real-world deployment. 
\major{In summary, a deployable ML-based adblocking approach should be able to generate filter rules without relying on existing \fl{s} for training, while also providing control to navigate the trade-off between blocking ads and avoiding breakage. 
To the best of our knowledge, \tool{} is the only system that can generate and evaluate filter rules automatically (without relying on humans) and from scratch (without relying on existing filter lists).}

\descr{Reinforcement Learning.} 
We formulate the problem of filter rule curation {\em from scratch} (\ie{} without any ground truth or existing list) as a reinforcement learning (RL) problem; see Sec.~\ref{sec:algorithm}. 
Within the vast literature in RL~\cite{sutton2018reinforcement}, we choose the Multi-Arm Bandits (MAB) framework~\cite{auer2002finite}, for reasons explained in Sec.~\ref{sec:problem-overview}. 
Identifying the top--k arms~\cite{Bubeck2013TopKArms,Locatelli2016threshold} rather than searching for the one best arm~\cite{Gabillon2012TopArm} has been used in the problems of coarse ranking~\cite{katariya2018adaptive} and  crowd-sourcing~\cite{NIPS2015_ab233b68, heinecke2019crowdsourced}. Contextual MAB has been used to create user profiles to personalize ads and news~\cite{li2010contextual}. Bandits where arms have similar expected rewards, 
commonly called Lipschitz bandits~\cite{kleinberg2004nearly}, have also been utilized in ad auctions and dynamic pricing problems~\cite{kleinberg2003value}. In our context of \frglower{}, we leverage the theoretical guarantees established for MAB to search for ``good'' filter rules and identify the ``bad'' filter rules, while searching for opportunities of ``potentially good'' filter rules (hierarchical problem space~\cite{Wang2021ReinforcementLH}), as discussed in Sec.~\ref{sec:alg-design}. 
While RL algorithms, in general, have been applied to several  application domains~\cite{elmalaki2021fair, yu2019deep, boroushaki2021rfusion, elmalaki2018sentio}, RL often faces challenges in the real-world  ~\cite{dulacarnold2019challenges} including convergence and adversarial settings~\cite{Gleave2020Adversarial,xu2021observation, rakhlin2016bistro,immorlica2019adversarial,auer2016algorithm}.

\major{
\descr{Our Work in Perspective}.}
\major{The design of the framework is described in Sec.~\ref{sec:algorithm} and illustrated in Fig.~\ref{fig:formulation}. \tool{} is the first to fully automate the process of \frglower{} and create URL-based, \persite{} rules that block ads from scratch, using reinforcement learning. The majority of prior ML-based techniques relied on existing filter lists at some point in their pipeline, thus creating a circular dependency.  Furthermore, \tool{} is the first to choose the granularity of the URL-based rule to explicitly optimize the trade-off between blocking ads and avoiding visual breakage.} 

\major{The implementation is described in Sec.~\ref{sec:autofrg-tool}\inarxivversion{ and illustrated in Fig.~\ref{fig:autofr-impl}}. Within the \inusenixversion{general} RL framework, \tool{}'s key design contributions include the action space, the RL components (\eg{} agent, environment, reward, policy), the annotation of raw \adgraph{s} into site snapshots, and the logic and implementation of utilizing site snapshots to emulate site visits. The latter was instrumental in scaling the approach (it reduced the time for generating rules for a single site from approximately 13 hours to 1.6 minutes) and making our results reproducible.}
\major{For some individual RL components, we leverage  state-of-the-art tools: (1) we utilize one part of \adgraph{} that creates a graph representing the site (we do {\em not} use the trained ML model of \adgraph{}); and (2) we use \adhigh{} to automatically detect ads, which is used to compute our reward function. As these individual components improve over time, the \tool{} framework can benefit from new and improved versions or even incorporate newly available tools in the future.} 

%% file: formulation.tex
\section{\tool{} Framework}
\label{sec:algorithm}
We formalize the problem of \frglower{}, including the process followed by human \fla{s} (Sec.~\ref{sec:fla-workflow} and Fig.~\ref{fig:fla}), our formulation as a reinforcement learning problem (Sec.~\ref{sec:problem-overview} and Fig.~\ref{fig:formulation}), and our  multi-arm bandit algorithm for solving it (Sec.~\ref{sec:alg-design} and Alg.~\ref{alg:autofrg-algorithm}). 
\inarxivversion{Table~\ref{tab:notation-summary} in the appendix summarizes the notation used throughout the paper.}

\subsection{Filter List Authors' Workflow}
\label{sec:fla-workflow}

\descr{Scope.} Among all possible filter rules, we focus on the important case of {\em URL-based rules for \major{blocking ads}} to demonstrate our approach.
\inarxivversion{
In \appref~\ref{app:filter-list-over-time}, we provide a longitudinal analysis of \fl{s} to show that these rules are the most widely used today.}
Table~\ref{tab:network-filter-rule-syntax} shows examples of URL-based rules at different granularities: blocking by the effective second-level domain (eSLD), fully qualified domain (FQDN), and including the path. 

\begin{table}%
	\footnotesize
	\centering
	\begin{tabularx}{\columnwidth}{ r | l | l  }
	    \toprule
		\textbf{} & \textbf{Description} & \textbf{Filter Rule} \\
		\midrule
		 1 & eSLD & $||ad.com$\textcaret{}  \\
		 \midrule 
 		 2 & FQDN & $||img.ad.com$\textcaret{}  \\
        \midrule
		 3 & With Path & 
		 \parbox{7cm}{$||ad.com/banners/$ or $||img.ad.com/banners/$} 
		   \\
	    \bottomrule
	\end{tabularx}
	\caption{{\small \textbf{URL-based Filter Rules.}  They block requests, listed from coarser to \finergrain{}: eSLD (effective second-level domain), FQDN (fully qualified domain), With Path (domain and path).}} 
	\vspace{-5pt}
	\label{tab:network-filter-rule-syntax}
\end{table}

\descr{Filter List Authors' Workflow for Creating Filter Rules.}
Our design of \tool{} is motivated by the bottlenecks of \frglower{}, revealed by prior work~\cite{cvinspectorndss,AlrizhaErrors}, our discussions with \fla{s}, and our own experience in curating filter rules. 
Next, we break down the process that \fla{s} employ into a sequence of tasks, also illustrated in Fig.~\ref{fig:fla}. When \fla{s} create filter rules for a specific site, they start by visiting the site of interest using the browser's developer tools. They observe the outgoing network requests and create, try, and select rules through the following workflow. 

\begin{challenge}[Select a Network Request]
    \label{chal:what-to-block}  
    \fla{s} consider the set of outgoing network requests and treat them as candidates to produce a filter rule. The intuition is that blocking an  ad request will prevent the ad from being served. For sites that initiate many outgoing network requests, it may be time-consuming to go through the entire list. When faced with this task, \fla{s} depend on sharing knowledge of ad server domains with each other or heuristics based on keywords like ``ads'' and ``bid'' in the URL. \fla{s} may also randomly select network requests to test. 
\end{challenge}

\begin{challenge}
[Create a Filter Rule and Apply]
\label{chal:convert-and-apply} 
\fla{s} must create a filter rule that blocks the selected network request. However, there are many options to consider since rules can be the entire or part of the URL, as shown in Table~\ref{tab:network-filter-rule-syntax}. %
 \fla{s} intuitively handle this problem by trying first an eSLD filter rule because the requests can belong to an ad server (\ie{} all resources served from the eSLD relate to ads). However, the more specific the filter rule is (\eg{} eSLD $\rightarrow$ FQDN), the less likely it would lead to breakage.
 Then, the \fla{s} apply the filter rule of choice onto the site. 
\end{challenge}

\begin{challenge} [Visual Inspection] \label{chal:effective-filter-rules}  
Once the filter rule is applied on the site, \fla{s} inspect its effect, \ie whether it indeed blocks ads and/or causes breakage (\ie legitimate content goes missing or the page displays improperly).  
\fla{s} use differential analysis. They visit a site with and without the rule applied, and they visually inspect the page and observe whether ads and non-ads (\eg{} images and text) are present/missing before/after applying the rule. 
In assessing the effectiveness of a rule, it is essential to ensure that it blocks at least one request, \ie{} \textit{a hit}. Filter rules are considered ``good'' if they block ads without breakage and ``bad'' otherwise. Avoiding breakage is critical for \fla{s} because rules can impact millions of users. 
If a rule blocks ads but causes breakage, it is considered a ``potentially good'' rule. 
\end{challenge}
    
\begin{challenge}[Repeat]\label{chal:repeat} 
\fla{s} repeat the process of \task{s} \ref{chal:what-to-block}, \ref{chal:convert-and-apply}, \ref{chal:effective-filter-rules}, multiple times to make sure that the filter rule is effective. Repetition is necessary because modern sites typically are dynamic. Different visits to the same site may trigger different page content being displayed and different ads being served.
If a rule from \task{}~\ref{chal:convert-and-apply} blocks ads but causes breakage, the author may then try a more granular filter rule (\eg{} eSLD $\rightarrow$ FQDN from Table~\ref{tab:network-filter-rule-syntax}). If the rule does not block ads, go back to Task~\ref{chal:what-to-block}.
\end{challenge}
   
\begin{challenge}[Stop and Store Good Filter Rules]\label{chal:stop-condition}
\fla{s} stop this iterative process when they have identified a set of filter rules that block most ads without breakage \major{(\ie{} a best-effort approach)}. None of the considered rules may satisfy these (somewhat subjective) conditions, in which case no filter rules are produced.
\end{challenge}

\descr{Bottlenecks: Scale and Human-in-the-Loop.} 
The workflow above is labor-intensive  and does not scale well. 
There is a large number of candidate rules to consider for sites with a large number of network requests (Task~\ref{chal:what-to-block}) and long and often obfuscated URLs (Task~\ref{chal:convert-and-apply}). 
The scale of the problem is amplified by site dynamics, which requires repeatedly visiting a site (Task~\ref{chal:repeat}). 
The effect of applying each single rule must then be evaluated by the human \fla{} through visual inspection (Task~\ref{chal:effective-filter-rules}), which is time-consuming on its own. 

Motivated by these observations, we aim to automate the process of \frglower{} \persite{}. We reduce the number of iterations needed (by intelligently navigating the search space for good filter rules via reinforcement learning), and we minimize the work required by the human \fla{} in each step (by automating the visual inspection and assessment of a rule as ``good'' or ``bad''). Our proposed methodology is illustrated in Fig.~\ref{fig:formulation} and formalized in the next section.

\subsection{Reinforcement Learning Formulation}
\label{sec:problem-overview}
As described earlier and illustrated in Fig.~\ref{fig:fla}, \fla{s} repeatedly apply  different rules and evaluate their effects until they build confidence on which rules are generally ``good'' for a particular site. This repetitive action-response cycle lends itself naturally to the~\textit{ reinforcement learning (RL)} paradigm, as depicted in Fig.~\ref{fig:formulation}, where actions are the applied filter rules and rewards (response) must capture the effectiveness of the rules upon applying them to the site (environment). Testing all possible filter rules by brute force is infeasible in practice due to time and power resources. However, RL can enable efficient navigation of the action space.

More specifically, we choose the \textit{multi-arm bandit (MAB)} RL formulation.
The actions in MAB are independent \textit{k-bandit arms} and the selection of one arm returns a numerical reward sampled from a stationary probability distribution that depends on this action. The reward determines if the selected arm is a ``good'' or a ``bad'' arm. Through repeated action selection, the objective of the MAB agent is to maximize the expected total reward over a time period~\cite{auer2002finite}.

The MAB framework fits well with our problem. The \textit{MAB agent} replaces the human (\fla{}) in  Fig.~\ref{fig:fla}. The agent knows all available ``arms'' (possible filter rules), \ie the action space; see Sec.~\ref{sec:action}. The agent picks a filter rule (arm) and applies it to the \textit{MAB environment}, which, in our case, consists of the site $\ell$ (with its unknown dynamics as per \task~\ref{chal:repeat}), the browser, and a selected configuration (how we value blocking ads vs. avoiding breakage, explained in Sec.~\ref{sec:alg-design}). The latter affects the reward of an action (rule) the agent selects. Filter rules are independent of each other.
Furthermore, the order of applying different filter rules does not affect the result. In adblockers, like Adblock Plus, blocking rules do not have precedence.
Through exploring available arms, the agent efficiently learns which filter rules are best at blocking ads while minimizing breakage; see Sec.~\ref{sec:rewards}. 
Next, we define the key components of the proposed \tool{} framework, depicted in Fig.~\ref{fig:formulation}. 
It replaces the \humanloop{} in two ways: (1) the \fla{} is replaced by the MAB policy that avoids brute force and efficiently navigates the action space; and (2) the reward function is automatically computed, as explained in Sec.~\ref{sec:rewards}, without requiring a human's visual inspection. 

\begin{figure}[t!]
	\centering
	\includegraphics[width=0.9\columnwidth]{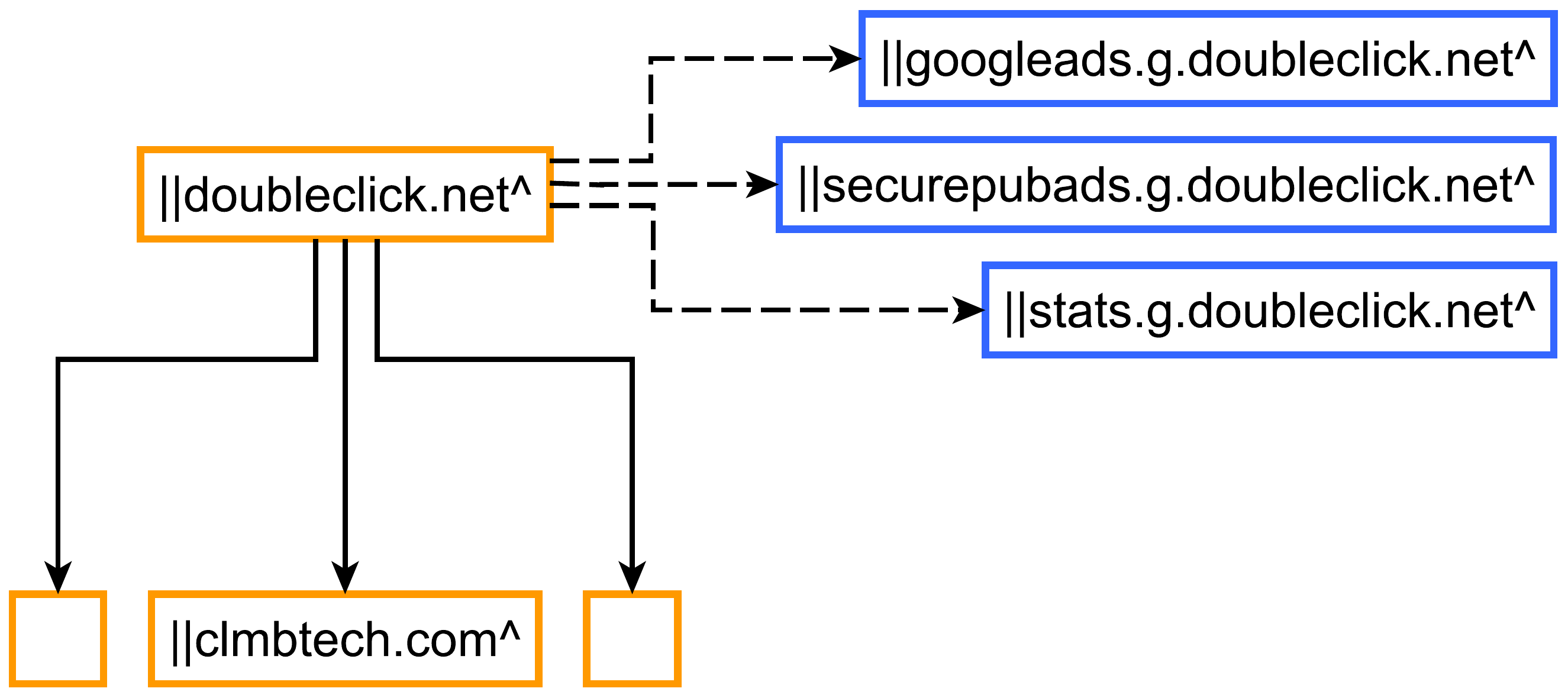}
	\caption{{\small \textbf{Hierarchical Action Space.} A node (filter rule) within the action space has two different edges (\ie{} dependencies to other rules): (1) the initiator edge, $\rightarrow$, denotes that the source node initiated requests to the target node; and (2) the \finergrain{} edge, $\dashrightarrow$, targets a request more specifically, as discussed in \task~\ref{chal:repeat} and Table~\ref{tab:network-filter-rule-syntax}. 
    \inarxivversion{An example of an entire action space is provided in \appref~\ref{app:action-space} and Fig.~\ref{fig:action-space}.} 
    }
	}
	\label{fig:action-space-node}
    \vspace{-5pt}
\end{figure}

\subsubsection{Actions}
\label{sec:action}

\descr{Action \textit{a} (Filter Rule).} An \textit{action} is a URL blocking filter rule that can have different granular levels, shown in Table~\ref{tab:network-filter-rule-syntax}, and is applied by the agent onto the environment. We use the terms action, arm, and filter rule, interchangeably.

\descr{Hierarchical Action Space $\mathbf{\mathcal{A}_H}$.}
Based on the outgoing network requests of a site $\ell$ (\task~\ref{chal:what-to-block}), there are many possible rules that can be created (\task~\ref{chal:convert-and-apply}) to block that request. Fig.~\ref{fig:action-space-node} shows an example of dependencies among candidate rules: 
\begin{enumerate}[leftmargin=*, topsep=0pt,itemsep=0pt,parsep=0pt,partopsep=0pt]
\item  We should try rules that are coarser grain first ($doubleclick.net$) before trying more \finergrain{} rules ($stats.g.doubleclick.net$) (the horizontal dotted lines). This intuition was discussed in \task{}~\ref{chal:repeat}.
\item If $doubleclick.net$ initiates requests to $clmbtech.com$, we should explore it first, before trying $clmbtech.com$ (the vertical solid lines). Sec.~\ref{sec:agent-impl} describes how we retrieve the initiator information.
\end{enumerate}

The dependencies among rules introduce a hierarchy in the \textit{action space}~\actionspace{}, which can be leveraged to expedite the exploration and discovery of good rules via pruning. If an action (filter rule) is good (it brings a high reward, as defined in Sec.~\ref{sec:rewards}), the agent no longer needs to explore its children.  
\inarxivversion{We further discuss the size of action spaces in \appref~\ref{app:top-5k-cont} and Fig.~\ref{fig:top5k-misc}; we show that they can be large.}
The creation of \actionspace{} automates \task{}~\ref{chal:convert-and-apply}.

\subsubsection{Rewards}
\label{sec:rewards}
Once a rule is created, it is applied on the site (\task~\ref{chal:convert-and-apply}). The human \fla{} visually inspects the site, before and after the application of the rule, and assesses whether ads have been blocked without breaking the page (\task~\ref{chal:effective-filter-rules}). To automate this task, we need to define a reward function for the rule that mimics the human \fla{'s} \major{assessment of whether a rule blocks ads and the breakage that could occur}.

\begin{figure}[t!]
	\centering
\includegraphics[width=0.7\columnwidth]{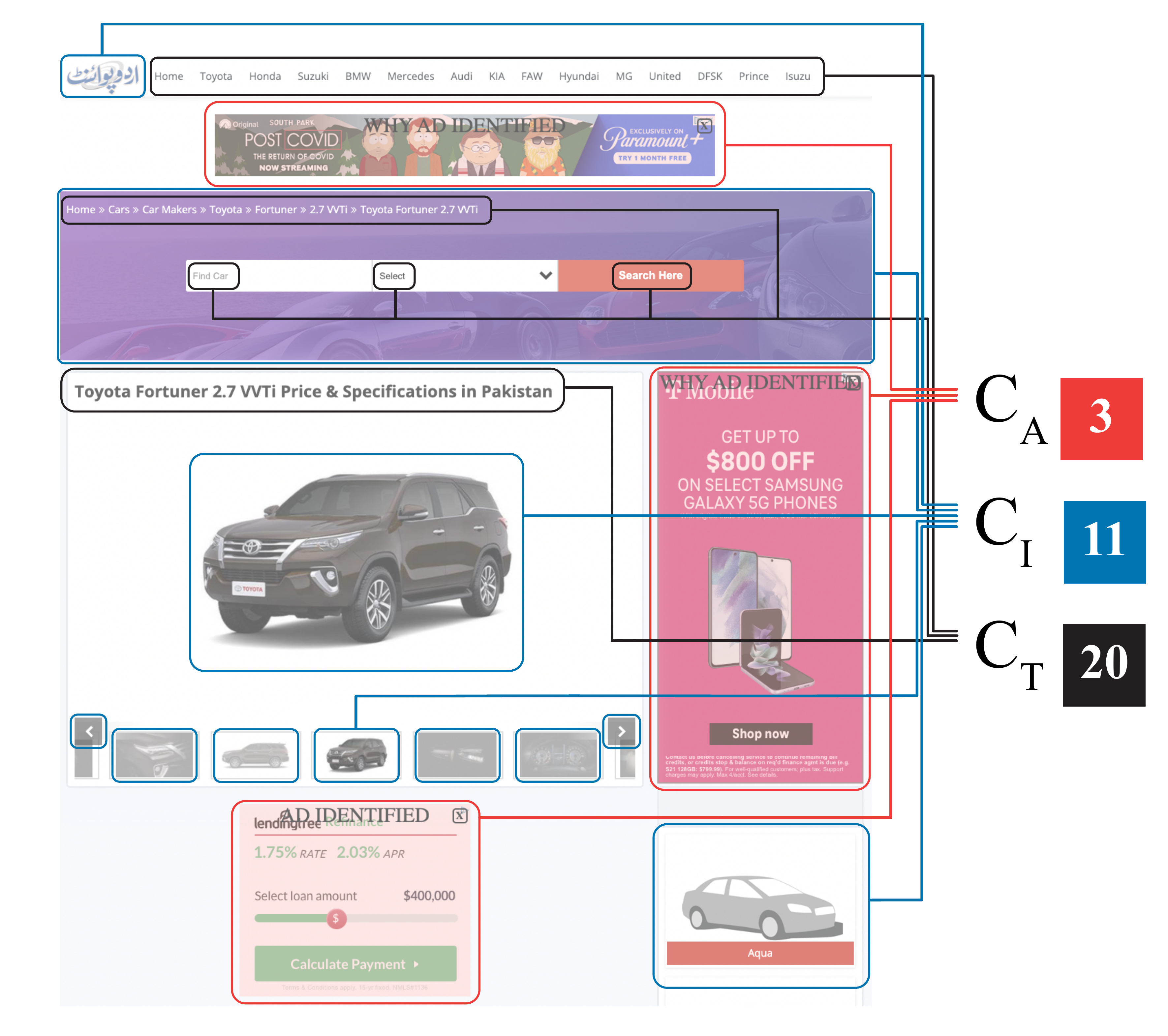}
	\caption{{\small \textbf{Site Representation.} We represent a site as counts of visible ads ($C_A$), images ($C_I$), and text ($C_T$), as explained in Sec.~\ref{sec:rewards}.
	Applying a filter rule changes them, by blocking ads (reducing $C_A$) and/or hiding legitimate content (changing $C_I$ and $C_T$, thus breakage $\mathcal{B}$).}}
	\label{fig:site-feedback-no-rule}
    \vspace{-5pt}
\end{figure}

\descr{Site Representation.} 
We abstract the representation of a site $\ell$ by counting three types of content visible to the user: we count the ads ($C_A$), images ($C_I$), and text ($C_T$) displayed.
An example is shown in Fig.~\ref{fig:site-feedback-no-rule}. 
The \textit{baseline representation} refers to the site before applying the rule. Since a site $\ell$ has unknown dynamics (\task{}~\ref{chal:repeat}), we need to visit it multiple times and average these  counters: $\overline{C}_A$, $\overline{C}_I$, and $\overline{C}_T$. 

We envision that obtaining these counters from a site can be done not only by a human (as it is the case today in \task~\ref{chal:effective-filter-rules}) but also automatically using image recognition (\eg{} \adhigh{}~\cite{futureofadblocking}) or better tools as they become available. This is an opportunity to remove the \humanloop{} and further automate the process. We further detail this in Sec.~\ref{sec:adhigh}.

\descr{Site Feedback after Applying a Rule.}
When the agent applies an action $a$ (rule), %
 the site representation will change from  $(\overline{C}_A, \overline{C}_I, \overline{C}_T)$ to ($C_A$, $C_I$, $C_T$).   The intuition is that, after applying a filter rule it is desirable to see the number of ads decrease as much as possible (ideally  $C_A=0$) and continue to see the legitimate content  (\ie no change in $C_I$, $C_T$ compared to the baseline). To measure the difference before and after applying the rule, we define the following: 
\begin{flalign}\label{eq:counter-norm}
\widehat{C}_A = \frac{\overline{C}_A - C_A}{\overline{C}_A}, && 
\widehat{C}_I = \frac{|\overline{C}_I - C_I|}{\overline{C}_I}, &&
\widehat{C}_T = \frac{|\overline{C}_T - C_T|}{\overline{C}_T} 
\end{flalign}
$\widehat{C}_A$ measures the fraction of ads blocked; the higher, the better the rule is at \emph{blocking ads}. Ideally all ads are blocked, \ie{} $\widehat{C}_A$ is $1$. In contrast, $\widehat{C}_I$ and $\widehat{C}_T$ measure the fraction of page broken. Higher values incur more breakage.
We define \textit{page breakage} ($\mathcal{B}$) as the visible images ($\widehat{C}_I$) and text ($\widehat{C}_T$), which are \textit{not} related to ads but are missing after a rule is applied: 
\begin{flalign} \label{eq:breakage}
    \mathcal{B} = \frac{\widehat{C}_I + \widehat{C}_T}{2}
\end{flalign}
\major{We take a neutral approach and treat both visual components equally and average $\widehat{C}_I$, $\widehat{C}_T$. This can be configured to express different preferences by the user, \eg{} treat content above-the-fold as more important.}
Lastly, \emph{avoiding breakage} is measured by $1 - \mathcal{B}$. It is desirable that $1-\mathcal{B}$ is $1$, and the site has no visual breakage.

\descr{Trade-off: Blocking Ads~($\widehat{C}_A$)~vs.~Avoiding Breakage~($1-\mathcal{B}$).}
The goal of a human \fla{} is to choose filter rules that block as many ads as possible (high $\widehat{C}_A$) without breaking the page (high $1-\mathcal{B}$). There are different ways to capture this trade-off. We could have taken a weighted average of $\widehat{C}_A$ and $\mathcal{B}$. However, to better mimic the practices of today's \fla{}s, we use a \emph{threshold} $w\in[0,1]$ as a design parameter to control how much  breakage a \fla{} tolerates: $1-\mathcal{B}\ge w$.  
Blocking ads is easy when there is no constraint on breakage --- one can choose rules that break the whole page. \fla{s} control this either by using more specific rules (\eg eSLD $\rightarrow$ FQDN) to avoid breakage or avoid blocking at all.
\major{We rely on this trade-off as the basis of our evaluation in Sec.~\ref{sec:eval}. 
\inarxivversion{An example is illustrated in \appref~\ref{app:top-5k-cont} and Fig.~\ref{fig:tradeoff-example}.}
It is desirable to operate where $\widehat{C}_A=1$ and $1-\mathcal{B}=1$. 
} 
In practice, \fla{}s tolerate little to no breakage, \eg{} $w \ge 0.9$. However, $w$ is a configurable parameter in our framework.  

\descr{Reward Function $\mathbf{\mathcal{R}_F}$.}
When the MAB agent applies a filter rule $F$ (action $a$) at time $t$ on the site $\ell$ (environment), this will lead to ads being blocked and/or content being hidden, which is measured by feedback ($\widehat{C}_A$, $\widehat{C}_I$, $\widehat{C}_T$) defined in Eq.~(\ref{eq:counter-norm}). We design a reward function $\mathcal{R}_{F}:\mathbb{R}^3\rightarrow [-1, 1]$ that mimics the \fla's assessment (\task~\ref{chal:effective-filter-rules}) of whether a filter rule $F$ is good ($\mathcal{R}_{F}(w,\widehat{C}_A,\mathcal{B})>0)$)  or bad ($\mathcal{R}_{F}(w,\widehat{C}_A,\mathcal{B})<0)$) \major{at blocking ads}
based on the site feedback:
\begin{subnumcases}{\label{eq:reward-function-new} \mathcal{R}_{F} (w,\widehat{C}_A,\mathcal{B})=} 
    -1 & \text{if } $\widehat{C}_A = 0$ \label{eq:reward-function-a} \\
    $0$ & \text{if } $\widehat{C}_A > 0$ , $1 - \mathcal{B} < w$ \label{eq:reward-function-b} \\
    \widehat{C}_A & \text{if } $\widehat{C}_A > 0$ , $1 - \mathcal{B} \geq w$ \label{eq:reward-function-c}
\end{subnumcases}
The rationale for this design is as follows.
\begin{enumerate}[label=\alph*),leftmargin=*, itemsep=0pt,parsep=0pt,partopsep=0pt]
    \item \emph{Bad Rules} (Eq.~(\ref{eq:reward-function-a})): If the action does not block any ads ($\widehat{C}_A = 0$), the agent receives a reward value of $-1$ to denote that this is not a useful rule to consider. 
    \item \emph{Potentially Good Rules} (Eq.~(\ref{eq:reward-function-b})): If the rule  blocks some ads ($\widehat{C}_A > 0$) but incurs breakage beyond the \fla{}'s tolerable breakage, then it is considered as ``potentially good''\footnote{\major{``Potentially'' means that the rule may have children rules within the action space that are effective at blocking ads with less breakage.}} and receives a reward value of zero. 
    \item \emph{Good Rules} (Eq.~(\ref{eq:reward-function-c})): If the rule blocks ads\footnote{\major{ Eq.~(\ref{eq:reward-function-new}) explicitly requires a rule to block at least some ads, to receive a positive reward. \tool{} can select rules that have additional side-benefits (\eg also blocks tracking requests, typically related to ads).}} and causes no more breakage than what is tolerable for the  \fla{}, then the agent receives a positive reward based on the fraction of ads that it blocked ($\widehat{C}_A$).
\end{enumerate}

\subsubsection{Policy}
\label{sec:policy}
Our goal is to identify ``good'' filter rules, \ie{} rules that give consistently high rewards. To that end, we need to refine our notion of a ``good'' rule and define a strategy for exploring the space of candidate filter rules. 

\descr{Expected Reward $\mathbf{Q_t(a)}$.}
The MAB agent selects an action $a$, following a policy, from a set of available actions $\mathbf{\mathcal{A}}$, and applies it on the site to receive a reward ($r_t=\mathcal{R}_F(w,\widehat{C}_A,\mathcal{B})$). It does this  over some time horizon $t=1,2,..,T$. 
However, due to the site dynamics as explained in \task{}~\ref{chal:repeat}, the reward varies over time, and we need a different metric that captures how good a rule is over time. In MAB, this metric is the weighted moving average of the rewards over time: $Q_{t+1}(a) = Q_t(a) + \alpha (r_t - Q_t(a))$, where $\alpha$ is the learning step size.

\descr{Policy.} 
Due to the large scale of the problem and the cost of exploring candidate rules, the agent should spend more time exploring good actions. The MAB policy utilizes $Q_t(a)$ to balance between exploring new rules in $\mathcal{A}_H$ and exploiting the best known $a$ so far. This process automates \task{}~\ref{chal:what-to-block} and~\ref{chal:convert-and-apply}. 

We use a standard Upper Bound Confidence (UCB) policy to manage the trade-off between exploration and exploitation~\cite{auer2002finite}. Instead of the agent solely picking the maximum $Q_t(a)$ at each $t$ to maximize the total reward, UCB considers an exploration value $U_t(a)$ that measures the confidence level of the current estimates, $Q_t(a)$. 
An MAB agent that follows the UCB policy selects $a$ at time $t$, such that $a_t = \argmax_a [Q_t(a) + U_t(a)]$. Higher values of $U_t(a)$ mean that $a$ should be explored more. It is updated using $U_t(a) = c \times \sqrt{\frac{\log N[a']}{N[a]}}$, where $N[a']$ is the number of times the agent selected all actions ($a'$) and $N[a]$ is the number of times the agent has selected $a$, and $c$ is a  hyper-parameter that controls the amount of exploration.

\subsection{\tool{} Algorithm}
\label{sec:alg-design}

\begin{algorithm}[t!]
\begin{algorithmic}[1]
\footnotesize
\Require
    \Statex Design-parameter:\hspace{0.10cm} $w \in [0,1]$
    
    \Statex Inputs:$\qquad\quad\quad$\hspace{0.27cm} Site ($\ell$) 
    \Statex $\qquad\qquad\quad\quad\quad$\hspace{0.25cm} Reward function ($\mathcal{R}_F:\mathbb{R}^3\rightarrow [-1, 1]$)
    \Statex $\qquad\qquad\quad\quad\quad$\hspace{0.25cm} Noise threshold (\noise{} $ = 0.05$)
    \Statex $\qquad\qquad\quad\quad\quad$\hspace{0.25cm} Number of site visits ($n = 10$)

    \Statex Hyper-parameters: \hspace{0.03cm} Exploration for UCB ($c = 1.4$)
    \Statex$\qquad\qquad\quad\quad\quad\quad$Initial Q-value ($Q_0 = 0.2$)
    \Statex$\qquad\qquad\quad\quad\quad\quad$Learning step size ($\alpha = \frac{1}{N[a]}$)
    \Statex$\qquad\qquad\quad\quad\quad\quad$Time Horizon ($T$)

    \Statex Output:\hspace{0.15cm} Set of filter rules ($\mathcal{F}$)

\State  \Procedure{Initialize}{$\ell$, $n$}
\State $\overline{C}_A, \overline{C}_I, \overline{C}_T,$ $reqs$ $\gets$ \textsc{VisitSite}($\ell$, $n$, $\emptyset$)
\State $\mathcal{A}_H \gets$ \textsc{BuildActionSpace}($reqs$) 

\State \Return $\overline{C}_A, \overline{C}_I, \overline{C}_T, \mathcal{A}_H$
\EndProcedure

\State 

\Procedure{AutoFR}{$\ell$, $w$, $c$, $\alpha$, $n$}
\State $\overline{C}_A, \overline{C}_I, \overline{C}_T, \mathcal{A}_H \gets \textsc{Initialize}(\ell, n)$
\State $\mathcal{F}\gets \emptyset$, $\mathcal{A}\gets \emptyset$
\State $\mathcal{A}\gets$ $\mathcal{A}_H$.root.children
\Repeat

    \State $Q(a) \gets Q_0$, $\forall a \in \mathcal{A}$ 
    \For{$t=1$ to $T$}
        \State $a_t \gets$ \textsc{ChooseArmUCB}($\mathcal{A}$, $Q_{t}$, $c$)
        \State  $C_{A_t}, C_{I_t}, C_{T_t},$ $hits$ $\gets$ 
        \textsc{VisitSite}($\ell$, 1, $a_t$) 
        \State $\widehat{C}_{A_t}, \widehat{C}_{I_t}, \widehat{C}_{T_t}$ $\gets$ \textsc{SiteFeedback}($C_{A_t}, C_{I_t}, C_{T_t}$)
        \State $\mathcal{B}_{t}$ $\gets$ \textsc{Breakage}($\widehat{C}_{I_t}, \widehat{C}_{T_t}$)
        \If{$a_t \in hits$}
            \State $r_t \gets$ $\mathcal{R}_F$($w$,  $\widehat{C}_{A_t}$,$\mathcal{B}_{t}$)
            
            \State $ Q_{t+1}(a_t) \gets Q_t(a_t) + \alpha (r_t - Q_t(a_t))$ 
        \Else 
        \State Put $a_t$ to sleep
        \EndIf
    \EndFor
    \State $\mathcal{A} \gets \{a$.children , $ \forall a \in \mathcal{A} \mid - $ \noise{} $ <= Q(a) <=  $ \noise$\}$
    \State $\mathcal{F} \gets \mathcal{F} \cup  \{\forall 
    a \in \mathcal{A} \mid Q(a) > $ \noise{} $\}$
    
    \Until $\mathcal{A}$ is $\emptyset$
    \State \Return $\mathcal{F}$
\EndProcedure
\end{algorithmic}
 \caption{\small \tool{} Algorithm}
 \label{alg:autofrg-algorithm}
\end{algorithm}

Algorithm~\ref{alg:autofrg-algorithm} summarizes our \tool{} algorithm. The inputs are the site $\ell$ that we want to create filter rules for, the design parameter (threshold) $w$, and various hyper-parameters\inusenixversion{.} ~\inarxivversion{(discussed in \appref~\ref{app:params}).}
In the end, it outputs a set of filter rules $\mathcal{F}$, if any.
It consists of the two procedures discussed next. 

\descr{\textsc{Initialize} Procedure.} First, we obtain the baseline representation of a site of interest $\ell$ (Sec.~\ref{sec:rewards}), when no filter rules are applied. To do so, it will visit the site $n$ times (\ie{} \textsc{VisitSite}) to capture some dynamics of $\ell$. The environment will return the average counters $\overline{C}_A, \overline{C}_I, \overline{C}_T,$ and the set of outgoing $reqs$. The average counters will be used in evaluating the reward function (Eq.~(\ref{eq:reward-function-new})). Next, we build the hierarchical action space \actionspace~ using all network requests $reqs$ (\task~\ref{chal:what-to-block},~\ref{chal:convert-and-apply}).  

\descr{\textsc{AutoFR} Procedure.} 
This is the core of \tool{} algorithm. We call \textsc{Initialize} and then traverse the action space \actionspace{} from the root node to get the first set of arms to consider, denoted as $\mathcal{A}$. 
Note that we treat every layer ($\mathcal{A}$) of \actionspace{} as a separate \textit{run} of MAB with independent arms (filter rules). 

One run of MAB starts by initializing the expected values of all ``arms'' at $Q_0$ and then running UCB for a time horizon $T$, as explained in Sec.~\ref{sec:policy}. Since the size of $\mathcal{A}$ can change at each run, we scale $T$ based on the number of arms; by default, we used  $100 \times \mathcal{A}.size$.
Each run of the MAB ends by checking the candidates for filter rules. In particular, we check if a filter rule should be further explored (down the \actionspace) or become part of the output set $\mathcal{F}$,  
 using Eq.~(\ref{eq:reward-function-new}) as a guide. A technicality is that Eq.~(\ref{eq:reward-function-b}) compares the reward $\mathcal{R}_F$ to zero, while in practice, $Q(a)$ may not converge to exactly zero. Therefore, we use a noise threshold ($\epsilon = 0.05$) to decide if $Q_t(a)$  is close enough to zero ($-\epsilon \le Q(a) \le \epsilon$). Then, we apply the same intuition  as in Eq.~(\ref{eq:reward-function-new}) but using $Q(a)$, instead of $R_F$, to assess the rule and next steps.
\begin{enumerate}[label=\alph*),leftmargin=*, itemsep=0pt,parsep=0pt,partopsep=0pt]
    \item \emph{Bad Rules: Ignore.} This case is not explicitly shown but mirrors Eq.~(\ref{eq:reward-function-a}). If a rule is $Q(a) <$~\noise{}, then we ignore it and do not explore its children.     
    \item \emph{Potentially Good Rules: Explore Further.} Mirroring Eq.~(\ref{eq:reward-function-b}), if a rule is within a range of $\pm $ \noise{} of zero, it helps with blocking ads but also causes more breakage than it is acceptable ($w$). In that case, \major{we ignore the rule but further explore its children within \actionspace.}  An example based on $doubleclick.net$ is shown on Fig.~\ref{fig:action-space-node}. In that case, $\mathcal{A}$ is reset to be the immediate children of these arms, and we proceed to the next MAB run.  
    \item \emph{Good Rules: Select.} When we find a good rule ($Q(a) > $ \noise{}), we add that rule to our list $\mathcal{F}$ and no longer explore its children. This mimicks Eq.~(\ref{eq:reward-function-c}). An example is shown in Fig.~\ref{fig:action-space-node}: if $doubleclick.net$ is a good rule, then  its children are not explored further.
\end{enumerate}

We repeatedly run MAB until there are no more potentially good filter rules to explore\footnote{When we find a rule that we cannot apply, we put it to ``sleep'', in MAB terminology. This is because they do not block any network request (\ie{} no hits, in \task{}~\ref{chal:effective-filter-rules}), and we expect them to not likely affect the site in the future, either.}. This stopping condition automates \task{}~\ref{chal:stop-condition}.
The output is the final set of good filter rules $\mathcal{F}$.

%% file: implementation.tex
\begin{figure}[t!]
	\centering
	\includegraphics[width=1\columnwidth]{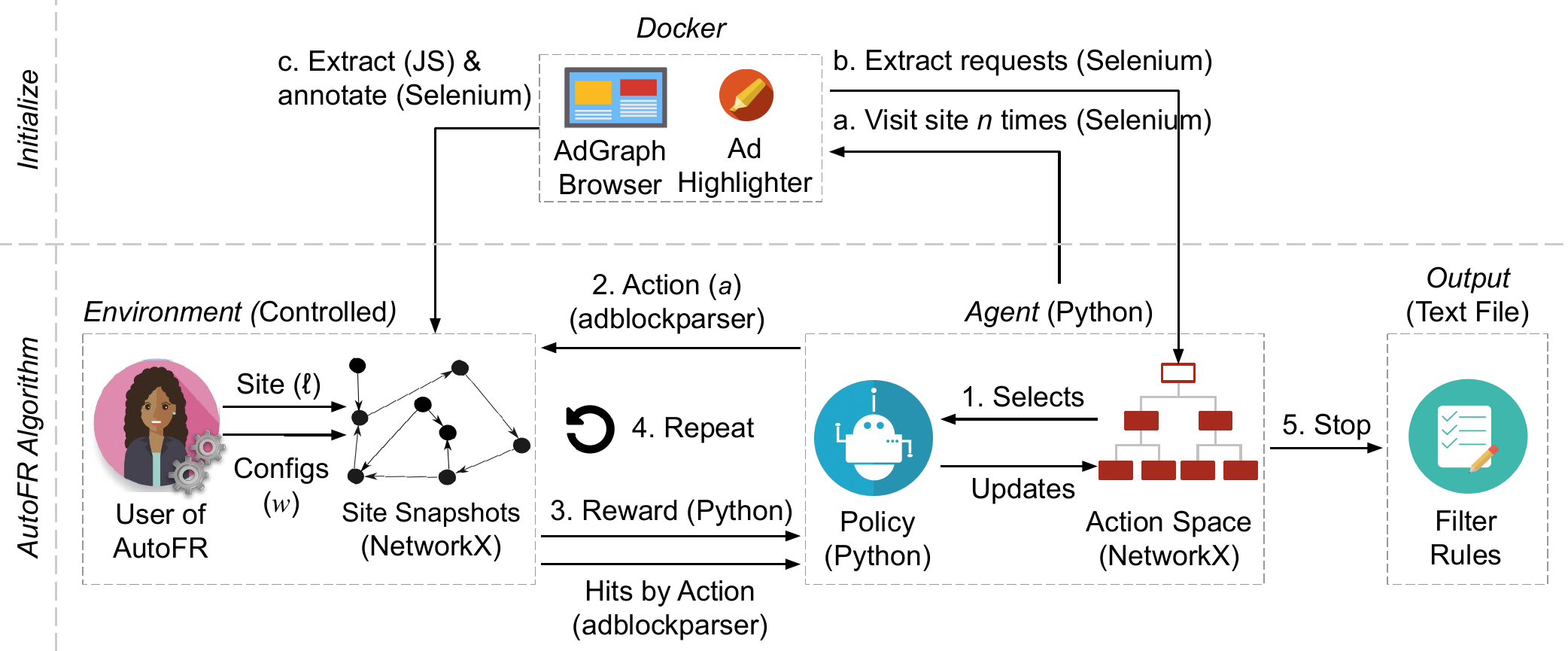}
	\caption{{\small \major{ \textbf{\tool{} Example Workflow (Controlled Environment).} \textsc{Initialize} (a--c, Alg.~\ref{alg:autofrg-algorithm}): (a) spawns $n=10$ docker instances and visits the site until it finishes loading; (b) extracts the outgoing requests from all visits and builds the action space; (c) extracts the raw graph and annotates it to denote $C_A$, $C_I$, and $C_T$, using JS and Selenium. Once all \inittimes{} site snapshots are annotated, we run the RL portion of the \textsc{AutoFR} procedure (steps 1--4). Lastly,~\tool{} outputs the filter rules at step 5, \eg{} \textit{$||$s.yimg.com/rq/darla/4-10-0/html/r-sf.html}.}}}
	\label{fig:autofr-impl}
    \vspace{-5pt}
\end{figure}

\section{\tool{} Implementation}
\label{sec:autofrg-tool}

In this section, we present the \tool{} tool that fully implements the RL framework as described in the previous section. \tool{} removes the \humanloop. The \fla{} only needs to provide their preferences (\ie how much they care about avoiding breakage via $w$) and hyper-parameters (detailed in Alg.~\ref{alg:autofrg-algorithm}), and the site of interest $\ell$. \tool{} then automates \task{s}~\ref{chal:what-to-block}--~\ref{chal:stop-condition} and outputs a list of filter rules $\mathcal{F}$ specific to $\ell$, and their corresponding values $Q$. 

\descr{Implementation Costs.} 
Let us revisit Fig.~\ref{fig:formulation} and reflect on the interactions with the site. The MAB agent (as well as the human \fla{}) must visit the site $\ell$, apply the filter rule, and wait for the site to finish loading the page content and ads (if any). The agent must repeat this several times to learn the expected reward of rules in the set of available actions $\mathcal{A}$.
First, for completeness, we implemented exactly that in a live environment 
\inusenixversion{(referred to as \toolwild{} with details in~\cite{autofr-arvix}).}
\inarxivversion{(referred to as \toolwild{}: details in \appref~\ref{app:autofrg-wild} and evaluation in \appref~\ref{app:control-vs-realtime}).} 

We employed cloud services using Amazon Web Services (AWS) to scale to tens of thousands of sites. This has high computation and network access costs and, more importantly, introduces long delays until convergence.

To make things concrete. For the delay, we found it took 47 seconds per-visit to a site, on average, by sampling 100 sites in the Top--5K. Thus, running \tool{} for one site with ten arms in the first MAB run, for 1K iterations, would take \autofrravg{} hours for one site alone! For the monetary cost, running \toolwild{} on 1K sites and scaling it using one AWS EC2 instance \persite{} (\$0.10/hour) would cost roughly \$1.3K for 1K sites, or \$1.3 to run it once \persite{}.
This a well-known problem with applying RL in a real-world setting. 
Thus, an implementation of \tool{} that creates rules by interacting with live sites is inherently slow, expensive, and does not scale to a large number of sites.

\descr{Scalable and Practical.}
Although \toolwild{} is already an improvement over the human workflow, we were able to design an even faster tool, which produces rules for a single site in minutes instead of hours. The core idea is to create rules in a realistic but controlled environment, where the expensive and slow visits to the website are performed in advance, stored once, and then used during multiple MAB runs, as explained in Sec.~\ref{sec:alg-design}. In this section, we present the design of this implementation in a controlled environment: \tool{}-C, or \tool{} for simplicity. \major{An overview of our implementation is provided in Fig.~\ref{fig:autofr-impl}. Importantly, this allows our \tool{} tool to scale across thousands of sites and, thus, utilized as a practical tool.}

\subsection{Environment}
\label{sec:env-impl}
To deal with the aforementioned delays and costs during training, we replace {\em visiting} a site live with {\em emulating} a visit to the site, using saved site snapshots.   
This provides advantages: (1) we can parallelize and speed up the collection of snapshots, and then run MAB off-line; (2) we can reuse the same stored snapshots to evaluate different $w$ values, algorithms, or reward functions while incurring the collection cost only once; and (3) we plan to make these snapshots available to the community (\ie{} it can replicate our results and utilize snapshots in its own work).

\descr{Collecting and Storing Snapshots.} Site snapshots are collected up-front during the \textsc{Initialize} phase of Alg.~\ref{alg:autofrg-algorithm} and saved locally. We illustrate this in Fig.~\ref{fig:autofr-impl}, steps a--c. We use \adgraph{}~\cite{iqbal2018adgraph}, an instrumented Chromium browser that outputs a graph representation of how the site is loaded. To capture the dynamics, we visit a site multiple times using Selenium to control \adgraph{} and collect and store the site snapshots. The environment is dockerized using Debian Buster as the base image, making the setup simple and scalable. 
For example, we can retrieve \inittimes{} site snapshots in parallel, if the host machine can handle it. In Sec.~\ref{sec:autofrg-control-rules}, we find that a site snapshot takes \sstimeavg{} seconds on average to collect. Without parallelization, this would take 8 minutes to collect \inittimes{} snapshots sequentially.

\begin{figure}[t!]
	\centering
	\includegraphics[width=0.7\columnwidth]{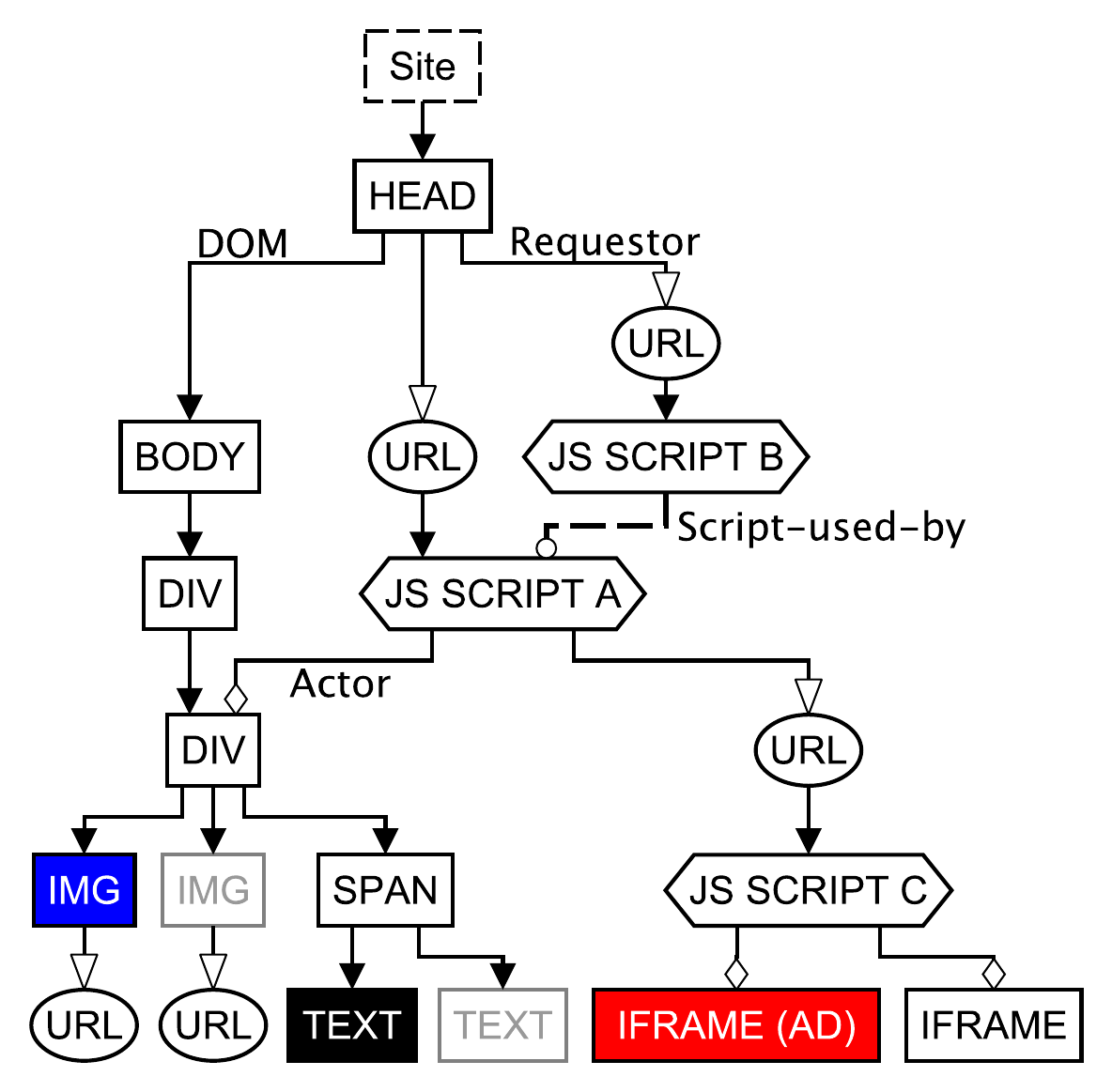}
	\caption{{\small \textbf{Site Snapshot.} It is a graph that represents how a site is loaded. %
	The nodes represent JS Scripts, HTML nodes (\eg{} DIV, IMG, TEXT, IFRAME), and network requests (\eg{} URL).  ``Actor'' edges track which source node added or modified a target node. ``Requestor'' edges denote which nodes initiated a network request. ``DOM'' edges capture the HTML structure between HTML nodes. Lastly,~``\scriptusedby'' edges track how JS scripts call each other. As described in Sec.~\ref{sec:env-impl}, nodes annotated by \tool{} have filled backgrounds, while grayed-out nodes are invisible to the user.}}
	\label{fig:sitesnapshot}
    \vspace{-5pt}
\end{figure}

\descr{Defining Site Snapshots.}
Site snapshots represent how a site $\ell$ is loaded. They are directed graphs with known root nodes and possible cycles. An example is shown in Fig.~\ref{fig:sitesnapshot}.
Site snapshots are large and contain thousands of nodes and edges\inarxivversion{; see \appref~\ref{app:top-5k-cont}, Fig.~\ref{fig:top5k-misc}}.
We use \adgraph{} as the starting point for defining the graph structure
and build upon it. 
First, we automatically identify the visible elements, \ie ads (AD), images (IMG), and text (TEXT) \major{(technical details in Sec.~\ref{sec:adhigh})}, for which we need to compute counts $C_A$, $C_I$, and $C_T$, respectively.
Second, once we identify them, we make sure that \adgraph{} knows that these elements are of interest to us. Thus, we annotate the elements with a new attribute such as ``FRG-ad'', ``FRG-image'', and ``FRG-textnode'' set to ``True''.
Annotating is challenging because ads have complex nested structures, and we cannot attach attributes to text nodes.
Third, we include how JS scripts interact with each other using ``Script-used-by'' edges, shown in Fig.~\ref{fig:sitesnapshot}. 
Lastly, we save site snapshots as ``.graphml'' files.
\inarxivversion{Due to lack of space, we defer technical details on building site snapshots to \appref~\ref{app:build-site-snapshots}.}

\descr{Emulating a Visit to a Site.} Emulation means that the agent does not actually visit the site live but instead reads a site snapshot and traverses the graph to infer how the site was loaded. 
To emulate a visit to the site, we randomly read a site snapshot into memory using NetworkX and traverse the graph in a breadth-first search manner starting from the root --- effectively replaying the events (JS execution, HTML node creation, requests that were initiated, etc.) that happened during the loading of a site. 
This greatly increases the performance of \tool{} as the agent does not wait for the \persite{} visit to finish loading or for ads to finish being served. Thus, reducing the network usage cost.
We hard-code a random seed (40) so that experiments can be replicated later. 

\descr{Applying Filter Rules.}
To apply a filter rule, we use an offline adblocker, adblockparser~\cite{adblockparser}, which can be instantiated with our filter rule. If a site snapshot node has a URL, we can determine whether it is blocked by passing it to adblockparser. We further modified adblockparser to expose which filter rules caused the blocking of the node (\ie{} hits). If a node is blocked, we do not consider its children during the traversal.

\descr{Capturing Site Feedback from Site Snapshots.} 
The next step is to assess the effect of applying the rule on the site snapshot. At this point, the nodes of site snapshots are already annotated. We need to compute the counters of ads, images, and text ($C_A$, $C_I$, $C_T$), which are then used to calculate the reward function. Its python implementation follows Sec.~\ref{sec:rewards}.

We use the following intuition. If we block the source node of edge types ``Actor'', ``Requestor'', or ``\scriptusedby'', then their annotated descendants (IMG, TEXT, AD) will be blocked (\eg{} not visible or no longer served) as well. Consider the following examples on Fig.~\ref{fig:sitesnapshot}: (1) if we block JS Script A, then we can infer that the annotated IMG and TEXT will be blocked; (2) if we block the annotated IMG node itself, then it will block the URL (\ie{} stop the initiation of the network request), resulting in the IMG not being displayed; and (3) if we block JS Script B that is used by JS Script A, then the annotated nodes IMG, TEXT, IFRAME (AD) will all be blocked.  As we traverse the site snapshot, we count as follows.
If we encounter an annotated node, we increment the respective counters $C_A$. $C_I$, $C_T$. If an ancestor of an annotated node is blocked, then we do not count it.

\major{
\descr{Limitations.} 
To capture the site dynamics due to a site serving different content and ads, we perform several visits \persite{} and collect the corresponding snapshots. We found that \inittimes{} visits were sufficient to capture site dynamics in terms of the eSLDs on the site, which is a similar approach taken by prior work~\cite{cvinspectorndss,Zhu_DisruptAA}\inusenixversion{.}
~\inarxivversion{(see \appref~\ref{app:params}).} 
However, there is also a different type of dynamics that snapshots miss. When we emulate a visit to the site while applying a filter rule, we infer the response based on the stored snapshot. In the live setting, the site might detect the adblocker (or detect missing ads~\cite{cvinspectorndss}) and try to evade it (\ie{} trigger different JS code), thus leading to a different response that is not captured by our snapshots. 
\inarxivversion{We evaluate this limitation in \appref~\ref{app:support-eval} and show that it does not greatly impact the effectiveness of our rules.}
Another limitation can be explained via Fig.~\ref{fig:sitesnapshot}. When JS Script B is used by JS Script A, we assume that blocking B will negatively affect A. Therefore,  if A is responsible for IMG and TEXT, then blocking B will also block this content;  this may not happen in the real world. When we did not consider this scenario, we found that \tool{} may create filter rules that cause major breakage. Since breakage must be avoided and we cannot differentiate between the two possibilities, we maintain our conservative approach.
}

\subsection{Agent}
\label{sec:agent-impl}

\descr{Action Space \actionspace{}.} 
\inarxivversion{During the \textsc{Initialize} procedure (Alg.~\ref{alg:autofrg-algorithm}), we visit the site $\ell$ multiple times and construct the action space, 
as explained in \appref~\ref{app:action-space} and summarized here.}
\inusenixversion{During the \textsc{Initialize} procedure (Alg.~\ref{alg:autofrg-algorithm}), we visit the site $\ell$ multiple times and construct the action space from all the visits.}
First, we convert every request to three different filter rules, as shown in Table~\ref{tab:network-filter-rule-syntax}. We add edges between them (eSLD $\rightarrow$ FQDN $\rightarrow$ With path), which serve as the \finergrain{} edges, shown in Fig.~\ref{fig:action-space-node}. We further augment \actionspace{} by considering the ``initiator'' of each request, retrieved from the Chrome DevTools protocol and depicted in solid lines in Fig.~\ref{fig:action-space-node}.  
This makes the \actionspace{} taller and reduces the number of arms to  explore per run of MAB, as described in Sec.~\ref{sec:alg-design}.
The resulting action space is a directed acyclic graph with nodes that represent filter rules; 
\inarxivversion{ see Fig.~\ref{fig:action-space-node} for a zoom-in along with  \appref~\ref{app:action-space} and Fig.~\ref{fig:action-space} for a larger example.}
\inusenixversion{Fig.~\ref{fig:action-space-node} provides a zoom-in example.}
We implement it as a NetworkX graph and save it as a ``.graphml'' file, a standard graph file type utilized by prior work~\cite{alex2019generation}.

\descr{Policy.} The UCB policy of  Sec.~\ref{sec:policy} is implemented in python. At time $t$ (Alg.~\ref{alg:autofrg-algorithm}, line 14), the agent retrieves the filter rule selected by the policy and applies it on the randomly chosen site snapshot instance.

\subsection{\major{Automating Visual Component Detection}}
\label{sec:adhigh}
A particularly time-consuming step in the human workflow is \task{}~\ref{chal:effective-filter-rules} in Fig.~\ref{fig:fla}. The \fla{} visually inspects the page, before and after they apply a filter rule, to assess whether the rule blocked ads ($\widehat{C}_A$) and/or impacted the page content ($\widehat{C}_I$, $\widehat{C}_T$).
\tool{} in Fig.~\ref{fig:formulation} summarizes this assessment in the reward in Eq.~(\ref{eq:reward-function-new}). However, to minimize the human work, we also need to replace the visual inspection and automatically detect and annotate elements as ads (AD), images (IMG), or text (TEXT) on the page. 

\descr{Detection of AD (Perceptual).}
To that end, we automatically detect ads using \adhigh{}~\cite{futureofadblocking}, a perceptual ad identifier (and web extension) that detects ads on a site.  We evaluated different ad perceptual classifiers, including Percival~\cite{abi2020percival}, and we chose \adhigh{} because it has high precision and does {\em not} rely on existing filter rules. 
\major{We utilize Selenium to traverse nested iframes to determine whether \adhigh{} has marked them as ads.}
\inarxivversion{The details of how \adhigh{} works are deferred to \appref~\ref{app:build-site-snapshots},~\ref{app:adhigh}.}

\descr{Detection of IMG and TEXT.}
We automatically detect visible images and text by using Selenium to inject our custom JS that walks the HTML DOM and finds image-related elements (\ie{} ones that have background-urls) or the ones with text node type, respectively. To know if they are visible, we see whether the element's or text container's size is $>$ 2px~\cite{cvinspectorndss}. 

\descr{Discussion of the Visual Components.}  It is important to note that our framework is agnostic to how we detect elements on the page. For detecting ads, this can be done by a human, the current \adhigh{}, future improved perceptual classifiers, heuristics, or any component that identifies ads with high precision. This also applies to detecting the number of images and text. Images can be counted using an instrumented browser that hooks into the pipeline of rendering images~\cite{abi2020percival}. Text can be extracted from screenshots of a site using Tesseract~\cite{futureofadblocking}, an OCR engine. Therefore, the \tool{} framework is modular and dependent on how well these components perform. 

\major{
\descr{Discussion of Blocking Ads vs. Tracking.}
We focus on detecting ads and generating filter rules that block ads for two reasons. First, they are the most popular type of rules in \fl{s}\inarxivversion{ (\appref~\ref{app:filter-list-over-time}, Fig.~\ref{fig:filterlists-per-year})}. 
Second, ads can be visually detected, enabling a human (\fla{}) or a visual detection module (such as \adhigh{}) to assess if the rule was successful (the ad is no longer displayed) or not at blocking ads.  
Although tracking is related to ads, it is impossible to detect visually, and assessing the success of a rule that blocks tracking is more challenging, \eg involves JS code analysis~\cite{chen2021jssignatures}. Extending \tool{} for tracking is a direction for future use.
}

%% file: evaluation.tex
\section{Evaluation}
\label{sec:eval}
In this section, we evaluate the performance of \tool{} (\ie{} the trade-off between blocking ads and avoiding breakage) and compare it to EasyList as a baseline. \major{In addition, we characterize properties of the filter rules produced by \tool{}: how they can be controlled via parameter $w$, how they compare to EasyList rules, how fast they need to be updated, and how well they generalize across sites.}
\inarxivversion{Parameter selection, automated evaluation workflow, and more can be found in \appref~\ref{app:eval}.}

\begin{table}[t!]
	\footnotesize
	\centering
	\begin{tabularx}{\linewidth}{X r r r}
	    \toprule
		\textbf{Datasets} $w=0.9$ & \textbf{Sites} & \textbf{Filter Rules} & \textbf{Snapshots}
		\\
		\midrule
        \parbox{8cm}{\wdata~(Sites $\geq$ 1 rule)} & \siteruletotal & \ruletotal & 9.3K \\
        \midrule
        \parbox{8cm}{\fulldata~(All sites)}  & \sitetotal & \ruletotal & 10.4K \\

	    \bottomrule
	\end{tabularx}
	\caption{\small \textbf{\tool{} Top--5K Results.} 
	}
	\label{tab:top-site-results}
	\vspace{-5pt}
\end{table}

\subsection{Filter Rule Evaluation Per-Site}
\label{sec:autofrg-control-rules}
We apply \tool{} on the Tranco Top--5K sites ~\cite{LePochat2019,TrancoList} to generate rules using the breakage tolerance threshold of $w=0.9$. \major{All other \tool{} parameters are the same as in Alg.~\ref{alg:autofrg-algorithm}.}

\descr{\tool{} Results.}
Table~\ref{tab:top-site-results} summarizes our results. 
Overall, \tool{} generated \ruletotal{} filter rules for \siteruletotal{} sites. 
For some sites, \tool{} did not generate any rules since none of the potential rules were viable at the selected $w$ threshold. 

\descr{Efficiency.} %
\tool{} is efficient and practical: it can take \autofrpersite--9 minutes to run \persite{}\inarxivversion{ (see \appref~\ref{app:top-5k-cont})}, which is an order of magnitude improvement over the 13 hours \persite{} of live training in Sec.~\ref{sec:autofrg-tool}. 
During each \persite{} run, we explore tens to hundreds of potential rules and conduct up to thousands of iterations within MAB runs\inusenixversion{.} \inarxivversion{(see Fig.~\ref{fig:top5k-misc}).}
This efficiency is key to scaling \tool{} to a large number of sites and over time.

\begin{figure*}[t!]
\centering
    \subfigure[\small \tool{} (Snapshots)]{
		 \includegraphics[width=.66\columnwidth]{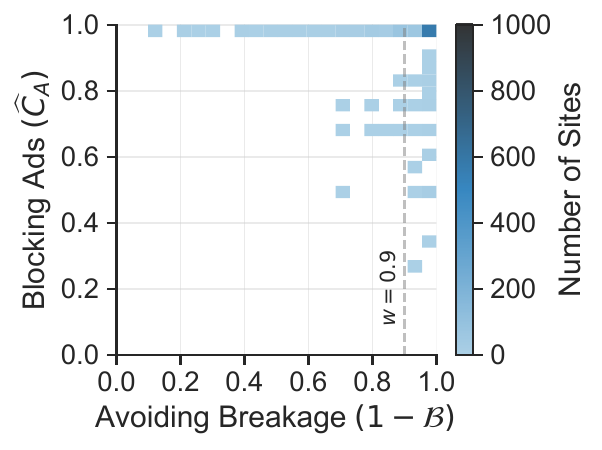}
        \label{fig:top5k-control-autofrg-snapshots}
	}
    \subfigure[\small \tool{} (In the Wild)]{
		 \includegraphics[width=.66\columnwidth]{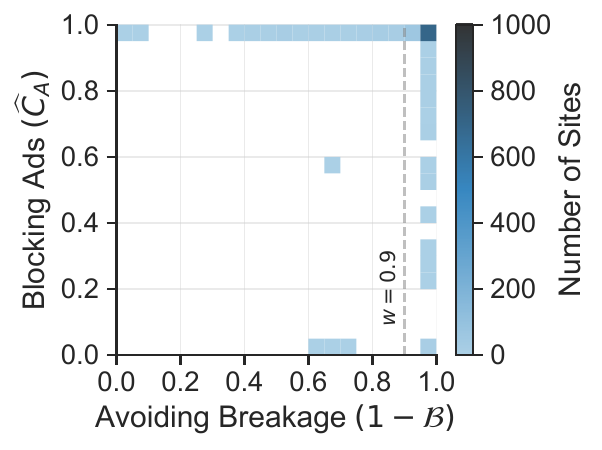}
        \label{fig:top5k-control-autofrg}
	}
	\subfigure[\small EasyList (In the Wild)]{
		 \includegraphics[width=.66\columnwidth]{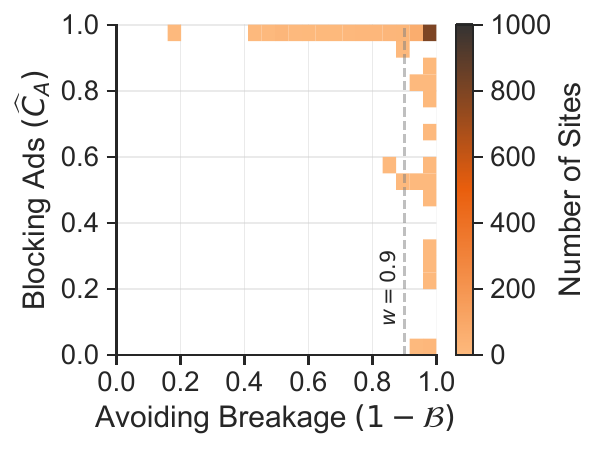}
        \label{fig:top5k-control-easylist}
	}
\caption{\small \textbf{\tool{} (Top--5K).} 
All sub-figures exhibit similar patterns. First, the filter rules were able to block ads with minimal breakage for the majority of sites. Thus, the top-right bin (the \bestbin) is the darkest. Second, there are edge cases for sites with partially blocked ads within the $w$ threshold (right of $w$ line) and sites below the $w$ threshold (left of $w$ line). \inarxivversion{Fig.~\ref{fig:tradeoff-example} explains how to read these plots.}
See Table~\ref{tab:all-results}, col.~\snapshotcol, \autofrcol, and \eltopfivekcol, for additional information.}
\label{fig:autofrg-control-vs-easylist}
\end{figure*}

\begin{figure*}[t!]
\centering
	\subfigure[\small\tool{} (All Rules)]{
		 \includegraphics[width=.66\columnwidth]{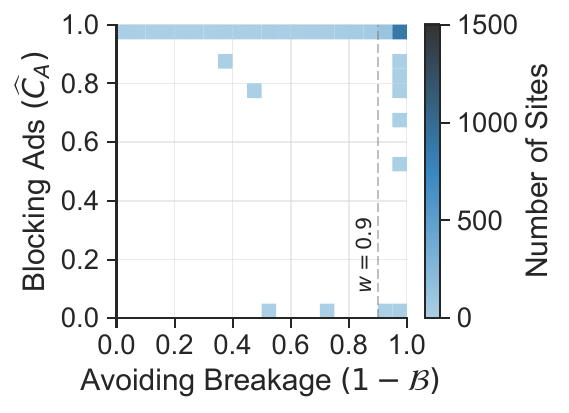}
        \label{fig:autofrg-top10k-threshold1}
	}
    \subfigure[\small \tool{} (Rules from $\geq$ 3 sites)]{
		 \includegraphics[width=.66\columnwidth]{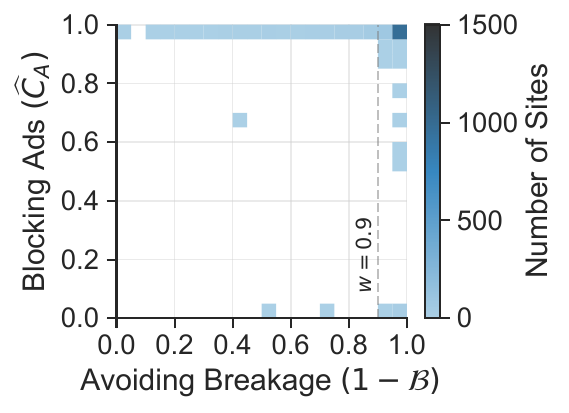}
        \label{fig:autofrg-top10k-threshold3}
	}
	\subfigure[\small EasyList (In the Wild)]{
		 \includegraphics[width=.66\columnwidth]{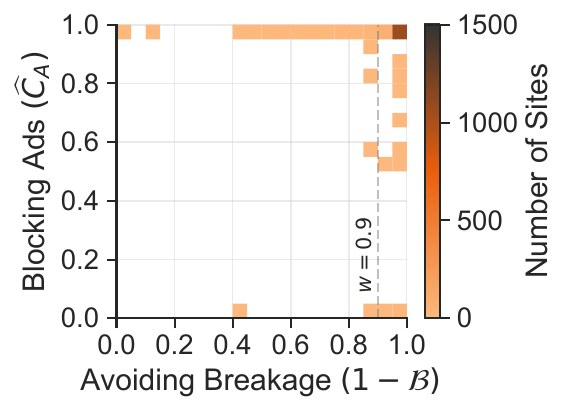}
        \label{fig:autofrg-top10k-easylist}
	}
\caption{\small \textbf{Testing Filter Rules on New Sites (Top 5K--10K, In the Wild).} We create two \fl{s}, Fig.~\ref{fig:autofrg-top10k-threshold1} with all rules from \wdata{} and Fig.~\ref{fig:autofrg-top10k-threshold3} that contains rules that were created for $\geq 3$ sites. We test them in the wild on the Top--5K to 10K sites (new sites) and show their effectiveness along with EasyList (Fig.~\ref{fig:autofrg-top10k-easylist}). We observe that Fig.~\ref{fig:autofrg-top10k-threshold3} performs better,  blocking 8\% more ads than Fig.~\ref{fig:autofrg-top10k-threshold1}. 
\inarxivversion{Fig.~\ref{fig:tradeoff-example} explains how to read these plots.}
Table~\ref{tab:all-results}, col.~\aggonecol--\eltoptenkcol, contains additional information.}
\label{fig:autofrg-top10k}
\end{figure*}

\begin{table*}[t!]
	\footnotesize
	\centering
	\begin{tabularx}{\linewidth}{l X c c c c c c c c}
	    \toprule
	    &  & 
	    \multicolumn{4}{c}{Sec.~\ref{sec:autofrg-control-rules}, Fig.~\ref{fig:autofrg-control-vs-easylist}, Top--5K} & \multicolumn{1}{c}{Sec.~\ref{sec:long-study}} & \multicolumn{3}{c}{Sec.~\ref{sec:filter-lists-top10k}, Fig.~\ref{fig:autofrg-top10k}, Top--5K to 10K}
	    \\
	    \cmidrule(lr{1em}){3-6} \cmidrule(lr{1em}){7-7} \cmidrule(lr{1em}){8-10}
	     &  &
	    \rotatebox{70}{\parbox{1.4cm}{ \textbf{\tool{}} \\ (Snapshots)\\(Jan. 2022)}} &
	    \rotatebox{70}{\parbox{1.4cm}{\textbf{\tool{}} \\ (In the Wild)\\(Jan. 2022)}} &
	    \rotatebox{70}{\parbox{1.4cm}{\textbf{\tool{}}\\(*Confirm)\\(In the Wild)}} &
        \rotatebox{70}{\parbox{1.4cm}{\textbf{EasyList} \\(In the Wild)\\(Jan. 2022))}} &
        \rotatebox{70}{\parbox{1.4cm}{\major{\textbf{\tool{}} \\(In the Wild)\\(July 2022) }}} &
        \rotatebox{70}{\parbox{1.4cm}{\textbf{\tool{}} \\(All rules) \\(In the Wild)}} &
        \rotatebox{70}{\parbox{1.4cm}{\textbf{\tool{}} \\($\geq$ 3 sites)\\(In the Wild)}} &
        \rotatebox{70}{\parbox{1.4cm}{\textbf{EasyList} \\ (In the Wild)}}
        \\
        \cmidrule(lr{1em}){3-10}
         & \parbox{5cm}{\textbf{Description ($w=0.9$)}} & \snapshotcol & \autofrcol & \confirmcol & \eltopfivekcol & \reapplycol & \aggonecol & \aggthreecol & \eltoptenkcol
        \\
		\midrule
        1 & \parbox{7cm}{Sites in \bestbin: \\
        $\widehat{C}_A \geq 0.95$, $1-\mathcal{B} \geq 0.95$} 
        & 62\%  & 74\% & 85\% & 79\% & 72\% &
        67\% & 73\% & 80\% \\
        \midrule
        2 &\parbox{7cm}{Sites within $w$: \\$\widehat{C}_A > 0$, $1-\mathcal{B} \geq 0.9$}  
        & 77\% & \textbf{86\%} & \textbf{85\%} & 87\% & 82\% & 
        76\% & \textbf{80\%} & 87\% \\
        \midrule
        3 & \parbox{7cm}{Ads blocked within $w$:\\
        $\sum_{\ell} (\overline{C}_A \times \widehat{C}_A)$ / $\sum_{\ell} \overline{C}_A$; $1 - B \geq 0.9$} 
        & 70\% & \textbf{86\%} & \textbf{84\%} & 87\% & 78\% & 
        72\% & \textbf{80\%} & 86\% \\
        
	    \bottomrule
	\end{tabularx}
	\caption{\small \textbf{Results.} We provide additional results to Fig.~\ref{fig:autofrg-control-vs-easylist} and ~\ref{fig:autofrg-top10k}, within their respective sections. We explain the meaning of each row: (1) the number of sites that are in the \bestbin{} (top-right corner of the figures), where filter rules were able to block the majority of ads with minimal breakage; (2) the number of sites that are within $w$; and (3) the fraction of ads that were blocked across all ads within $w$. 
	\textit{*Confirming via Visual Inspection (In the Wild)} (Sec.~\ref{sec:autofrg-control-rules}): col.~\confirmcol{} is based on a binary evaluation. As it is not simple for a human to count the exact number of missing images and text, we evaluate each site based on whether the rules blocked all ads or not (\ie{} $\widehat{C}_A$ is either 0 or 1) and whether they caused breakage or not (\ie{} $\mathcal{B}$ is either 0 or 1). \major{For col.~\reapplycol{} (Sec.~\ref{sec:long-study}), we repeat the same experiment of col.~\autofrcol{} during July 2022 for a longitudinal study of \tool{} rules.}
	}
	\label{tab:all-results}
\end{table*}

\descr{\tool{}: Validation with Snapshots.}
\major{Since \tool{} generates rules for each particular site (\ie{} \persite{}), we first apply these rules to the site for which they have been created. To that end, we first apply the rules to the stored site snapshots,} 
and we report the results in Fig.~\ref{fig:top5k-control-autofrg-snapshots} and Table~\ref{tab:all-results} col.~\snapshotcol.
We see that the rules block ads on 77\% of the sites within the $w=0.9$ breakage threshold. 
As we demonstrate next, this number is lower due to the limitations of traversing snapshots (Sec.~\ref{sec:env-impl}) and the rules are more effective when tested on sites in the wild.

\descr{\tool{} \vs{} EasyList: Validation In The Wild.}
\major{Next, we apply the rules from \tool{} to the same sites they have been created for, but this time on the real site (``in the wild''), not on the site snapshots. For comparison, we also apply
EasyList\footnote{For a fair comparison, we parse EasyList and utilize delimiters (\eg{} ``\$'', ``$||$'', and ``\textcaret{}'') to identify URL-based filter rules and keep them.} 
to the same set of Top--5K sites and we report our results in Fig.~\ref{fig:top5k-control-autofrg} and Table~\ref{tab:all-results} col.~\autofrcol~and \eltopfivekcol.}
\tool{}'s rules block 95\% (or more) of ads with less than 5\% breakage for 74\% of the site (\ie{} within the \bestbin{}) as compared to 79\% for EasyList. 
For sites within the $w$ threshold, \tool{} and EasyList perform comparably at 86\% and 87\%, respectively (row 2). 
Overall, our rules blocked 86\% of all ads \vs{} 87\% by EasyList, within the $w$ threshold (row 3).
\inusenixversion{Some sites fall below the $w$ threshold partly due to the limitations of \adgraph{}~\cite{iqbal2018adgraph}.}
\inarxivversion{Some sites fall below the $w$ threshold partly \major{due to limitations discussed in \appref~\ref{app:top-5k-cont}, including limitations of \adgraph{}~\cite{iqbal2018adgraph}.}}

To further confirm our results for \tool{} and EasyList, we randomly selected \supportevaltotal{} sites (a sample size out of 933 sites to get a confidence level of 95\% with a 5\% confidence interval), and we visually inspected them. \major{In particular, we looked for breakage not perfectly captured by automated evaluation.} 
Table~\ref{tab:all-results} col. \confirmcol{} summarizes the results and confirms our results obtained through the automated workflow. 
We find that 3\% (7/\supportevaltotal{}) of sites had previously undetected breakage. 
For instance, the layout of four sites was broken (although all of the content was still visible), and one site's scroll functionality was broken. 
Note that this kind of functionality breakage is currently not considered by \tool{}. 
We observed two sites that intentionally caused breakage (the site loads the content, then goes blank) after detecting their ads were blocked. 
\tool{}'s implementation currently does not handle this type of adblocking circumvention.

\begin{figure}[t!]
\centering
    \subfigure[\small Rule Types]{
		 \includegraphics[width=0.47\columnwidth]{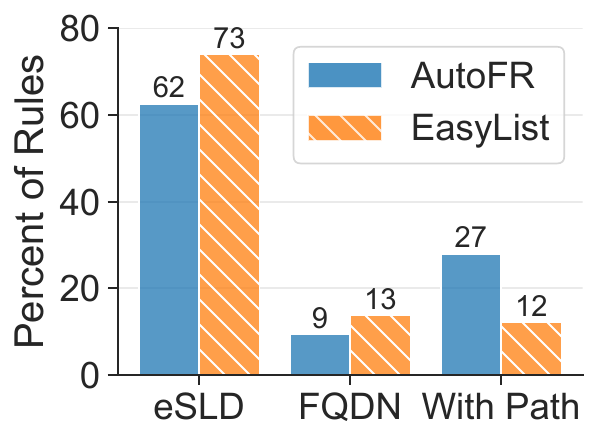}
        \label{fig:rule-type-comparison}
	}
    \subfigure[\small Grouped by eSLD]{
		 \includegraphics[width=.47\columnwidth]{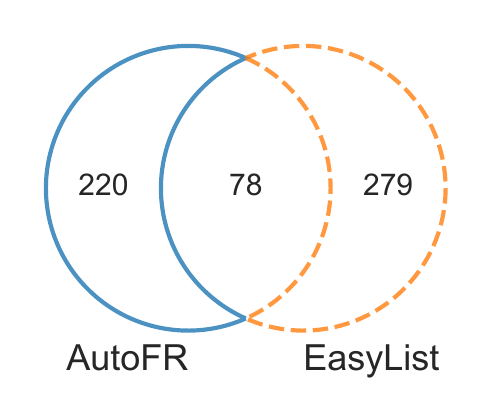}
        \label{fig:sld-venn-with-el}
	}
\vspace{-5pt}
\caption{\major{\small 
\textbf{Comparing \tool{} Rules to EasyList}. Some rules are common and some are unique to each approach. When comparing rules, one must consider the right granularity.}}
\label{fig:autofrg-vs-easylist-rules-high-level}
\end{figure}

\major{\descr{Tuning \tool{} via Threshold $\mathbf{w}$.}} 
\tool{} is the first approach that \major{can be tuned \persite{} and explicitly allows to express a preference.}  The FL author that uses \tool{} must select the site to create rules for and express their preference by tuning a knob (threshold $w$)\inusenixversion{.}
\inarxivversion{, which controls the trade-off between blocking ads \vs{} avoiding breakage. \major{Results are provided in \appref~\ref{app:effects-of-w}.}}

\subsection{\tool{} \vs{} EasyList: Comparing Rules}
\label{sec:autofr-vs-el-rules}

\major{
We compare the rules generated \persite{} by \tool{} and EasyList from Sec.~\ref{sec:autofrg-control-rules}. 
For a fair comparison, we only consider EasyList rules that are \textit{triggered} when visiting sites.

\subsubsection{Rule Type Granularity}
\label{sec:rule-type-comparison}

An important aspect to consider when comparing rules is the suitable granularity of the rules that block ads while limiting breakage. 
Fig.~\ref{fig:rule-type-comparison} breaks down the granularity of rules by \tool{} and EasyList. 
We note that both exhibit a similar distribution: eSLD rules are the most common, while the other rule types are less common.
Across all granularities, there are 59 identical rules (\eg{} $||pubwise.io$\textcaret{}, $||adnuntius.com$\textcaret{}, and $||deployads.com$\textcaret{}) between \tool{} and EasyList, which represents 
15\% of EasyList rules.

Next, we focus on rules that are \textit{related}, \ie{} they share a common eSLD but may differ in subdomain or path, to understand why AutoFR generates rules that are coarser or \finergrain{} than EasyList rules.
In Fig.~\ref{fig:sld-venn-with-el}, we show that when we group rules by eSLD, there are 78 common eSLDs, 60 (77\%) of which have at least one identical rule. 
For example, for $mail.ru$, both \tool{} and EasyList have $||ad.mail.ru$\textcaret{}. 

For 26 eSLD groups, \tool{} and EasyList rules differ in granularity. 
First, 18 eSLDs have \tool{} rules that are coarser-grained than EasyList. 
For instance, \tool{} has $||cloudfront.net$\textcaret{} but EasyList has 15 different rules based on FQDNs like $||d2na2p72vtqyok.cloudfront.net$\textcaret{}. 
CloudFront is a CDN that can serve resources for legitimate content, ads, and tracking. 
As \tool{} generates per-site rules, it can afford to be more coarse-grained because a particular site may only use CloudFront for ads and tracking. 
However, since EasyList rules that target CloudFront are not \persite{}, they are more \finergrain{} to avoid breakage on other sites.

Second, six eSLDs have \tool{} rules that are \finergrain{} than EasyList. 
For instance, for $moatads.com$, \tool{} has $||z.moatads.com$\textcaret{} when EasyList has $||moatads.com$\textcaret{}.
Recall in Sec.~\ref{sec:env-impl} that \tool{} generates rules with a conservative approach when using site snapshots, and thus will consider \finergrain{} rules for some cases to avoid breakage. 
Whereas \fla{s} manually verify rules for EasyList and will know that $||moatads.com$\textcaret{} is more appropriate.

Lastly, four eSLDs share the same granularity but contain rules that are not identical. 
For example, for site $pastemagazine.com$, \tool{} has $||pastemagazine.com/common/js/ads\text{-}gam\text{-}a9\text{-}ow.js$, while EasyList has $pastemagazine.com/common/js/ads\text{-}$. 
Partial paths within EasyList may extend the life of a filter rule over time for some sites. 
We further evaluate this in Sec.~\ref{sec:long-study}. 
\tool{} can extend to partial paths in the future.
}

\major{
\subsubsection{Understanding Unique Rules}
\label{sec:missed-rules}

We investigate why \tool{} generates rules that are not present in EasyList and vice versa. 
We found that when grouped by eSLD (Fig.~\ref{fig:sld-venn-with-el}), unique rules
are due to the design and implementation of our framework, as well as due to site dynamics. 

\descr{Methodology.}
To investigate each unique rule (either from \tool{} or EasyList), we apply the rule to its corresponding site snapshots (\persite{}) and extract the requests that were blocked. 
We manually investigate these requests as follows. 
For images, we visually decide whether it is an ad. 
For scripts, we use our domain knowledge and keywords (\eg{} ``advertising'', ``bid'') to examine the source code to discern whether they affect ads, tracking, functionality, or legitimate site content.
When we cannot determine the nature of the request (\eg{} due to obfuscated JS code), we fall back to applying the rule and evaluating its effectiveness via visual inspection, following the methodology in Sec.~\ref{sec:autofrg-control-rules}.

\descr{Findings.} 
Depicted in Fig.~\ref{fig:sld-venn-with-el}, the differences in rules when grouped by eSLDs are due to three main reasons. 

\textit{1. \tool{} Framework:} Our framework exhibits several strengths when generating rules. 
48\% (105/220) of the unique eSLDs for \tool{} have rules that are valid but seem challenging for a \fla{} to manually craft. 
Within this set, 19\% (20/105) are first-party (\eg{} $||kidshealth.org/\text{...}/inline\_ad.html$), 52\% (55/105) block resources that involve both ads and tracking (\eg{} $||snidigital.com$\textcaret{}), 23\% (24/105) block ad-related resources served by CDNs (\eg{} $||cdn.fantasypros.com/realtime/media\_trust.js$), and 42\% (44/105) block ad-related resources served through seemingly obfuscated URLs. %
We conclude that \tool{} can create rules that are not obviously ad-related (\eg{} by looking at keywords in the URL) but are effective nonetheless. 

Next, we explain how certain design decisions behind \tool{}'s framework can lead to missed EasyList rules. 
First, \tool{} focuses on rules that block at least some ads (due to Eq.~(\ref{eq:reward-function-a})), which is why \tool{} ignored 10\% (28/279) of unique eSLDs from EasyList that are responsible for purely tracking requests.
Second, we choose to generate rules that block ads across all \inittimes{} site snapshots of a site, not just one site snapshot, to be robust against site dynamics. 
In addition, we choose to stop exploring the hierarchical action space when we find a good rule following the intuition from Sec.~\ref{sec:action}, which improves the efficiency of \tool{}. 
Of course, these design decisions can be altered depending on the user's preference. 
When we do so, we find that the overlap in Fig.~\ref{fig:sld-venn-with-el} goes from 22\% (78/357) to 35\% (124/357). For example, $adtelligent.com$ and $adscale.de$ are new common eSLDs found when we remove these design decisions.

\textit{2. \tool{} Implementation:} Our implementation of Alg.~\ref{alg:autofrg-algorithm} 
focuses on visual components (\eg{} using \adhigh{} to detect ads) and how filter rules affect them. 
The rules generated are as good as the components that we utilize. %
First, \tool{} misses 28\% (78/279) of unique eSLDS from EasyList because \adhigh{} can only detect ads that contain transparency logos.
However, \tool{} rules are still effective when compared to EasyList, as shown in Sec.~\ref{sec:autofrg-control-rules} and Table~\ref{tab:all-results}. 
This demonstrates that we do not necessarily need to replicate all rules from EasyList to be effective. 
Second, 18\% of unique eSLDs from \tool{} can affect both ads and functionality (\eg{} cdn.ampproject.org/v0/amp-ad-0.1.js for ads, amp-accordion-0.1.js for functionality). 
\tool{} balances the trade-off between blocked ads and breakage, see Sec.~\ref{sec:autofrg-control-rules}.

\textit{3. Site Dynamics}  can also lead to differences in the site resources between site snapshots \vs{} the in the wild evaluation. 
Due to this, 18\% (50/279) of unique eSLDs on the EasyList side did not appear in our \wdata{}. 
Thus, \tool{} did not get an opportunity to generate these rules. 
Conversely, 5\% (11/220) of unique eSLDs from \tool{} appear in EasyList but were not triggered during the evaluation of EasyList rules. 
This can be mitigated by increasing the number of site snapshots used in \tool{'s} rule generation or applying EasyList more times during our in the wild evaluation. Although, recall that we already do these steps for \inittimes{} times.

\descr{Takeaways.} 
The difference in the granularity of related rules generated by \tool{} and EasyList is mainly because \tool{} creates rules \persite{}.
Unique rules to \tool{} or EasyList are due to the design and implementation of our framework and site dynamics. 
These differences are acceptable because the effectiveness of the rules from \tool{} and EasyList is comparable. This is crucial from a practical standpoint.
}

\begin{figure}[t!]
    \centering
    \includegraphics[width=1\columnwidth]{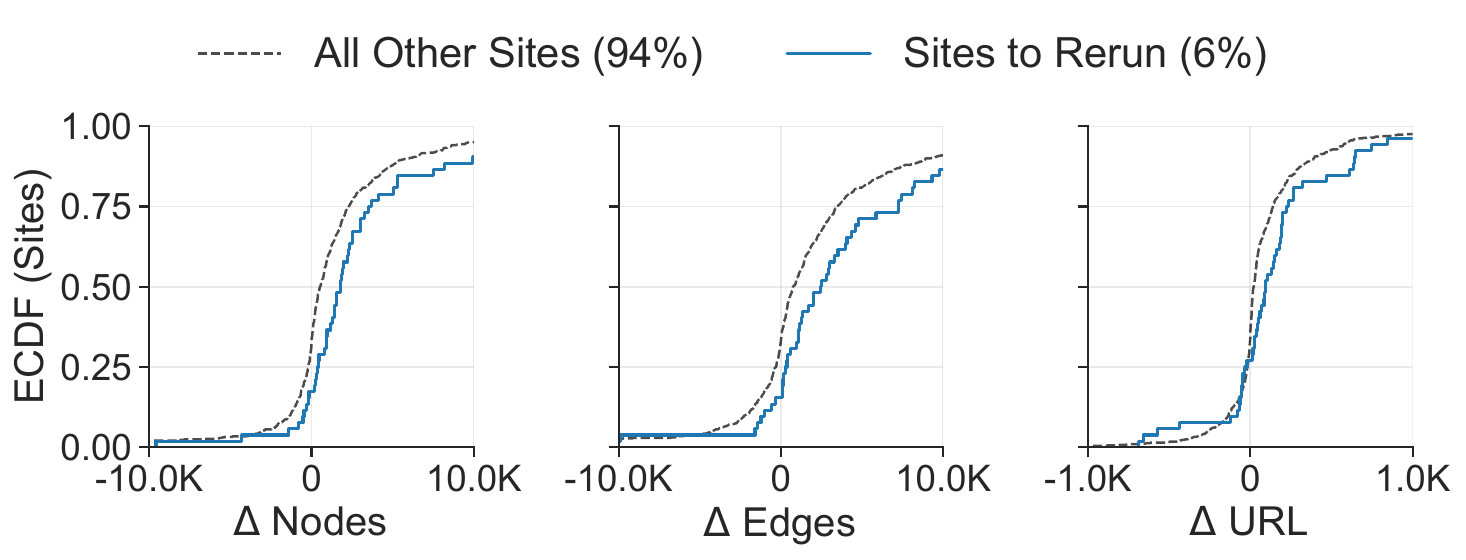}
    \vspace{-15pt}
    \caption{\major{{\small \textbf{$\Delta$ Site Snapshots between July vs. January 2022.} The differences in site snapshots for nodes, edges, and URLs. A positive change in the x-axis denotes that July had more of the respective factor, while a zero denotes no change.}}}
\label{fig:site-snapshot-stability-julyjan-allsites}
\end{figure}

\subsection{Robustness of \tool{} Filter Rules}
\label{sec:generalize-rules}
\major{\tool{} generates rules for a particular site and uses snapshots collected at a particular time. Next, we investigate and discuss how well these rules perform over time, across different sites, and in adversarial scenarios.}

\subsubsection{\major{How Long-lived are \tool{} Rules?}}
\label{sec:long-study}

\major{Sites change naturally over time, which may result in changes in the site snapshots, and eventually into changes in the filter rules. We show that \tool{} rules remain effective for a long time and can be rerun fast when needed to update.}

\major{
\descr{Efficacy of Rules Over Time.} 
We re-apply \persite{} rules generated in January 2022 (Sec.~\ref{sec:autofrg-control-rules}) to the same sites in July 2022 and summarize the results in Table~\ref{tab:all-results} (col.~\reapplycol). 
We find that the majority of \tool{} rules are still effective after six months. 72\% of sites (down only by 2\%) still achieve the operating point (row 1), and 82\% (down by 4\%) achieve $1-\mathcal{B} \geq 0.9$ (row 2). 
Even more interestingly, we found only 6\% of the sites now no longer have all or any ads blocked in July. For those few sites, which we refer to as ``sites to rerun'', we can  rerun \tool{}; this takes \autofrpersite{} min-\persite{} on average.

\descr{Site Snapshots Over Time.} 
We recollect site snapshots for our entire~\wdata{} in July 2022 and associate them with the results of re-applying the rules above. 
For the 6\% of sites that \tool{} needs to rerun, we report the changes in their corresponding snapshots.
Fig.~\ref{fig:site-snapshot-stability-julyjan-allsites} reports the changes in snapshots of the same site between January and July in terms of different nodes, edges, and URLs. It also compares the differences for all sites, with those 6\% sites to rerun \tool{}.
For all other sites, 50\% and 70\% of sites have more than $\pm$1K changes in nodes and edges, respectively; while 40\% of sites have more than $\pm$100 changes in URL nodes. 
Compared to sites to rerun, 75\% of sites have more than $\pm$1K changes in nodes and edges, while 65\% of sites have more than $\pm$100 changes in URL nodes. 
As expected, the snapshots of the sites to rerun indeed change more than other sites.
However, \tool{}'s rules remain effective on the vast majority of sites whose snapshots do not significantly change. 

\descr{Why do Rules become Ineffective?}
For the sites that need to be rerun, we conduct a comparative analysis of how rules change by rerunning \tool{} on those sites. 
We find that 23\% of these sites have completely new rules than before,
which is typically due to a change in ad-serving infrastructure on the site. 
40\% of the sites need some additional rules
(some older rules still work), which is due to additional ad slots on the site.
In addition, 9\% of the sites have changes in their paths.
Lastly, 29\% of these sites have the same rules as before. 
We deduce that this is because the rules are the best we can do without pushing breakage beyond the acceptable threshold $w$.

\descr{Takeaways.} 
\tool{} rules need to be updated for a small fraction of sites (6\% of Top--5K in six months), which demonstrates that \tool{} generates robust rules over time. \tool{} can be rerun for these sites at an average of \autofrpersite{} min-\persite{}.
} 

\begin{figure}[t!]
	\centering
	\includegraphics[width=1\columnwidth]{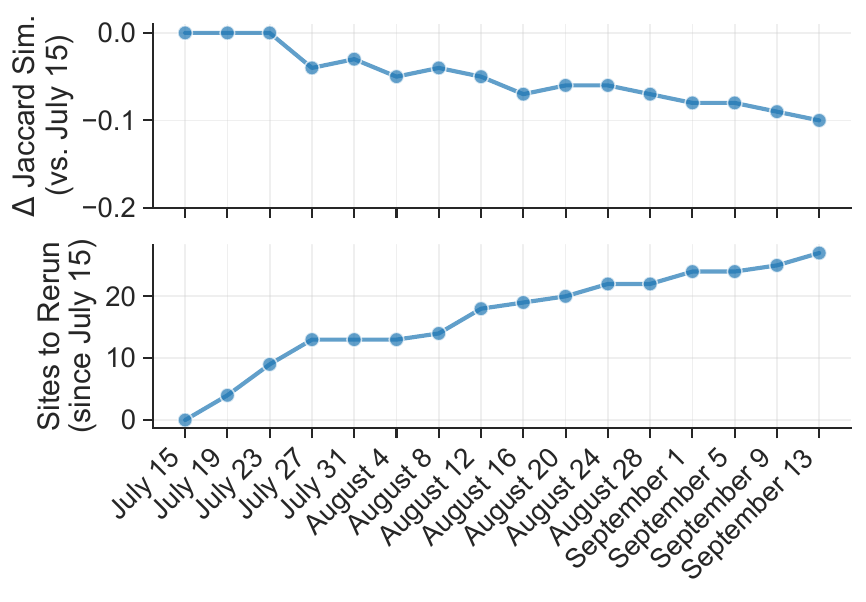}
	\vspace{-15pt}
	\caption{{\small \major{\textbf{Longitudinal Study Every Four Days.} We conduct a \finergrain{} longitudinal study of 100 sites over a two-month period. We find that over time, site snapshots will become less similar (\ie{} negative $\Delta$ Jaccard similarity), often denoting that rules may be less effective. \fla{s} can rerun \tool{} on these sites that change more frequently to output effective rules.
	}}}
	\label{fig:long-study-four-days}
    \vspace{-5pt}
\end{figure}

\major{
\subsubsection{How Frequently Should We Run \tool{}?}
\label{app:long-study-four-days}
Next, to understand how often \fla{s} should run \tool{} over time, we provide a \finergrain{} longitudinal study of every four days for two months to study how site snapshots change and the sites that need \tool{} to be rerun. We choose every four days because this is how often EasyList is updated and deployed to end-users. In addition, we choose to focus on 100 sites, two-thirds of which are sampled from \wdata{} and one-third is sampled from the set of 6\% of sites that need to rerun in July (from Sec.~\ref{sec:long-study}).
Fig.~\ref{fig:long-study-four-days} illustrates our two-month results, using July 15, 2022, as our baseline. In this study, using Jaccard similarity, our comparison considers the relationship between HTML, JS, and CSS (different nodes within site snapshots). To do so, we retrieve the path from the root to every URL node for every site snapshot. We then convert these paths to strings and use them to calculate the Jaccard similarity between the site snapshots of July 15 to subsequent dates shown in the figure. 

As expected, we arrive at the same conclusion as Sec.~\ref{sec:long-study}. As time passes, the similarity between site snapshots will naturally decrease, which denotes that there are sites where our rules are no longer effective, and we need to rerun \tool{} on them. For our 100 sites, we ran \tool{} on 13 sites only once (\eg{} $weheartit.com$, $legit.ng$), three sites twice (\eg{} $buzzfeednews.com$), and two sites three or more times (\eg $npr.org$), within two months. In terms of the time between the reruns of \tool{}, we find that one site (\eg{} $charlotteobserver.com$) varied between four to 10 days from August 12 to September 13. This was due to path changes that would evade our rules like $||charlotteobserver.com/.../0a086549941921c9ac8e.js$. 
Similarly, one site (\eg $npr.org$) varied from two weeks to one month.
In addition, two sites had runs that were 1--2 weeks apart (\eg{} \tool{} found additional rules for $amarujala.com$). Lastly, one site had runs that were one month apart (\eg{} $liputan6.com$ went from $||googlesyndication.com$\textcaret{} to a new rule, $||infeed.id$\textcaret{}).
By the end of this study, the similarity of site snapshots decreased by 10\% (compared to site snapshots of July 15), and we ran \tool{} 27 times on 18 unique sites within two months. 

\descr{Takeaways.} We find that each site will naturally change over time, causing site snapshots to be less similar. More changes often denote a higher possibility of rules being evaded. Overall, 18\% of 100 sites needed a rerun of \tool{}.  \fla{s} can periodically rerun \tool{} on sites that tend to change frequently in terms of weekly to monthly reruns. \tool{} minimizes the human effort for updating rules over time.
}

\subsubsection{\major{From Per-Site Rules To Global Filter Lists}}
\label{sec:filter-lists-top10k}

\major{\tool{} generates URL-based filter rules for a particular site. Similarly, EasyList supports \persite{} rules as well. It currently contains $\sim$800 \persite{} rules. Although these rules are guaranteed to perform well on the sites that they have been designed for (as demonstrated in Sec.~\ref{sec:autofrg-control-rules}), it is not guaranteed that the same rules are as effective when applied to other sites, \ie{} used as ``global'' rules. 
}

\begin{figure}[t!]
	\centering	\includegraphics[width=0.8\columnwidth]{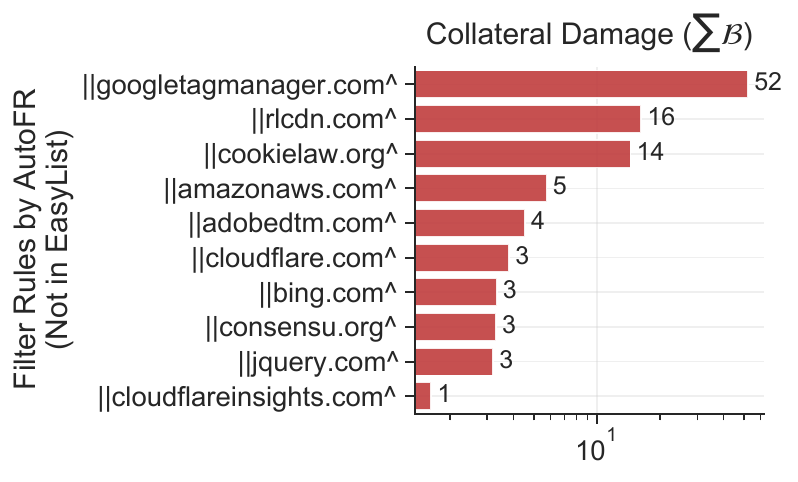}
	\vspace{-10pt}
	\caption{\major{\small \textbf{Collateral Damage of Global Rules.} \tool{} rules are generated \persite{} and can potentially cause breakage when applied to other sites (\ie{} treated as a global rule). We report the rules that are unique to \tool{} (\ie not part of EasyList), ordered by decreasing total collateral damage ($\sum \mathcal{B}$) that they cause to site snapshots within \fulldata{}. We can see that most of these rules (93\%) cause negligible collateral damage (below 10 on the x-axis). Note that the possible max $\sum \mathcal{B}$ of each rule is the size of the dataset.}
	}
	 \vspace{-5pt}
	\label{fig:collat-dmg}
\end{figure}

\major{
\descr{Collateral Damage.}
In Fig.~\ref{fig:collat-dmg}, we report the potential collateral damage, defined as the sum of breakage ($\sum \mathcal{B}$), caused when \tool{} rules are treated as global rules.
Rules are considered global when applied to sites other than the ones they have been created for.
We observe that they tend to block tag managers
(\eg{} $||googletagmanager.com$\textcaret{}, $||adobedtm.com$\textcaret{}), CDNs or cloud storage services (\eg{} $||cloudflare.com$\textcaret{}, $||amazonaws.com$\textcaret{}, $||rlcdn.com$\textcaret{}), third-party libraries (\eg{} $||jquery.com$\textcaret{}), and cookie consent forms
(\eg{} $||cookiekaw.org$\textcaret{}, $||consensu.org$\textcaret{}). 
These rules target domains that can serve legitimate content and ads across different sites.
Thus, adopting a \persite{} rule into a global rule is nontrivial because the rule may not block as many ads or may cause more breakage (\ie collateral damage). 
It is not a problem distinct to \tool{}. Our discussions with EasyList authors confirmed that  new rules are created \persite{}. They become global rules when \fla{s} know that the same rules are effective for other sites. \fla{s} rely on feedback from users to know when global rules either are ineffective or cause collateral damage on unknown sites\cite{AlrizhaErrors}. 
}

\begin{figure}[t!]
	\centering
	 \includegraphics[width=0.8\columnwidth]{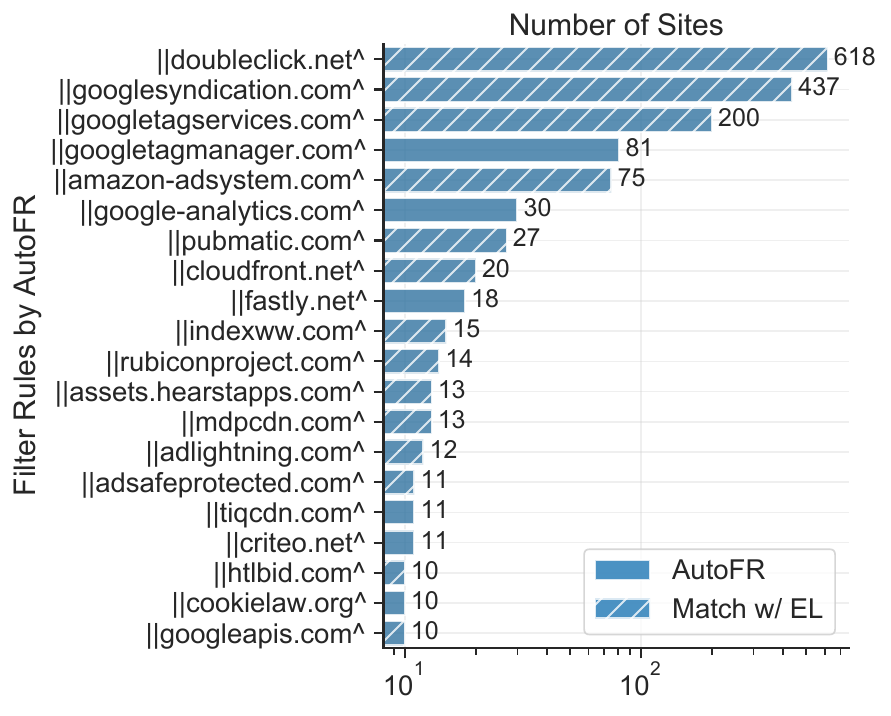}
	\vspace{-10pt}
	\caption{\small \major{\textbf{Top--20 Filter Rules by \tool{} for Top--5K Sites.} They include the main advertising and tracking services, such as Alphabet ($doubleclick.net$), Amazon ($amazon\text{-}adsystem.com$), and PubMatic ($pubmatic.com$). Thus, they are likely to generalize well.}}
	 \vspace{-5pt}
	\label{fig:top5k-rules-control-autofrg}
\end{figure}

\begin{figure}[t!]
	\includegraphics[width=1\columnwidth]{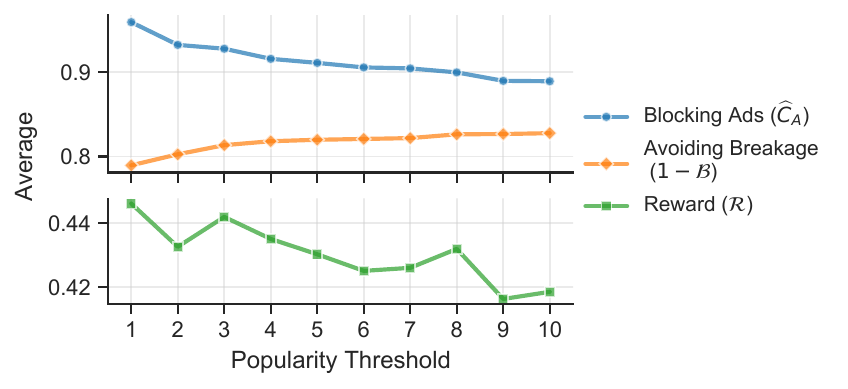}
	\vspace{-10pt}
	\caption{{\small \major{\textbf{Selecting Per-Site Rules into Global Filter Lists.} After creating the per-site \tool{} rules for each site (with $w=0.9$), we create 10 global \fl{s}. ``Popularity  1'' means that a rule is selected into the global list if it was generated in at least one site; ``popularity 10'' means that a rule is selected if it was generated for at least 10 sites. Once selected, the rules are now treated as global rules. We apply these global \fl{s} on our \fulldata{} site snapshots and plot the average blocking ads, avoiding breakage, and reward.}}}
	\label{fig:aggregation-threshold}
    \vspace{-5pt}
\end{figure}

\descr{Towards Global Filter Lists.} 
Although we cannot guarantee, in advance, how well \persite{} rules will perform on other sites, we can try heuristics and assess their performance. Intuitively, if the same filter rule is generated by \tool{} across multiple sites, then it has a better chance of generalizing to new sites. We denote this as the ``popularity'' of a rule. 
Fig.~\ref{fig:top5k-rules-control-autofrg} shows the Top--20 \tool{} most common rules across sites. They intuitively make sense as they belong to widely used advertising and tracking services.
Therefore, we utilize these heuristics as criteria to select \tool{} rules to include in \fl{s}. Once selected, we now treat them as global rules. As the popularity increases, the global \fl{} contains fewer global rules, resulting in fewer blocked ads but less breakage. We show the results in  Fig.~\ref{fig:aggregation-threshold}.

\major{We analyze in detail two global \fl{s}. First, ``popularity 1'' treats all \tool{} \persite{} rules as global rules, which serves as a baseline for comparison. Second, ``popularity 3'' denotes \tool{} rules that were generated from $\geq$ 3 sites. Fig.~\ref{fig:aggregation-threshold} reveals that this has the highest average reward. Note that selecting the popularity threshold based on the average reward implicitly considers collateral damage because it encompasses breakage (Eq.~(\ref{eq:reward-function-new})).}
We apply these \major{global \fl{s}} on the Tranco Top 5K--10K sites in the wild.
Fig.~\ref{fig:autofrg-top10k} and Table~\ref{tab:all-results} col. 5--6 show the results. 
As expected, we see that the global \fl{} created from rules that appeared in $\geq$ 3 sites perform better than the list with all rules. 
Moreover, Fig.~\ref{fig:autofrg-top10k-threshold3} compares relatively well against Fig.~\ref{fig:autofrg-top10k-easylist} (EasyList): 73\% of sites are in the desired \bestbin{} (top-right corner), \vs{} 80\% by EasyList (row 1, col.~\aggthreecol--\eltoptenkcol).
Overall, the rules generated from the Top--5K sites were able to block 80\% of ads on the Top 5K--10K sites. 
\major{This shows good generalization of \tool{} rules across unseen sites, which agrees with Fig.~\ref{fig:top5k-rules-control-autofrg}}. 

\major{
\subsubsection{Evading URL-based Filter Rules}
\label{sec:evading-url-rules}
\tool{} generates URL-based filter rules, which EasyList also supports.  
Well-known evasion techniques for URL-based filter rules, such as randomizing URL components, affect both \tool{} rules and EasyList rules~\cite{cvinspectorndss}. The strength of \tool{} is that new rules can be learned automatically and quickly (\eg in \autofrpersite{} min-per-site on average) when old ones are evaded.
Publishers and advertisers can also try to specifically evade \tool{}~\cite{cvinspectorndss,tramer2019adversarial}. For example, they can put ads outside of iframes, use different ad transparency logos, or split the logo into smaller images, preventing \adhigh{} from detecting ads~\cite{tramer2019adversarial}. This impacts our reward calculations. Defense approaches include the following. At the component level, we can try to improve \adhigh{} to handle new logos or look beyond iframes, replace \adhigh{} with a better future visual perception tool, or pre-process the logos to remove adversarial perturbations~\cite{ComDefend}. At the system level, as an adversarial bandits problem, where the reward received from pulling an arm comes from an adversary~\cite{auer2016algorithm}.
}

%% file: conclusion.tex
\section{Conclusion \& Future Directions}
\label{sec:conclusion}

\descr{Summary.}
The \fl{} curation follows a human-in-the-loop approach: (1) the rules are manually created, visually evaluated, and maintained; and (2) the \fla{} has to carefully balance between blocking ads \vs{} avoiding breakage.
We introduced the \tool{} framework to automate the process of generating \major{URL-based filter rules to block ads} from scratch. Our implementation of the framework allows it to learn rules without relying on existing rules created by humans.
Our evaluation showed that \tool{} is efficient and performs comparably to EasyList.
Thus, we envision that \tool{} will be used by the adblocking community to automatically generate and update filter rules at scale.
\inusenixversion{An extended version of this paper, including appendices, can be found at ~\cite{autofr-arvix}.}

\descr{Future Directions.} \tool{} provides a general framework for automating \frglower{}. In this paper, we focused specifically on the commonly used URL-based rules for blocking ads on browsers, but we envision several extensions and applications.
The \tool{} framework can be extended to include: (1) the creation of global rules, in addition to site-specific rules, (2) rules that block tracking; 
(3) other types of filter rules, such as element hiding rules, \eg using the concept of CSS specificity to leverage the hierarchy;
(4) functionality (beyond visual) breakage, \eg{} by testing click functionality for buttons and links; 
(5) new visual detection modules for images and ads on sites as these become available.
\tool{} can also be applied to other platforms, such as mobile, smart TVs, and VR devices, as there is a need for better platform-specific \fl{s}, in terms of coverage and breakage~\cite{shuba2018nomoads,varmarken2020tv,trimananda2022ovrseen}. On mobile and smart TVs specifically, one could leverage existing tools to automatically explore apps or mobile browsers~\cite{shuba2018nomoads,varmarken2020tv,droidbot,omnicrawl}.

\descr{Availability.}
The \tool{} implementation, generated filter rules, and the dataset are available at ~\cite{autofr-project}.

%% file: appendix.tex
\vfill
\newpage
\newpage

\section*{\scshape APPENDICES}
Due to space constraints, many additional implementation details, results, and discussions are deferred to these appendices.

\begin{table}
	\footnotesize
	\centering
	\begin{tabularx}{\columnwidth}{ p{3.5em} | p{5.5em} | X }
	    \toprule
		\textbf{Ref.} & \textbf{Notation} & \textbf{Definitions} \\
		\midrule
		\multicolumn{3}{l}{\textit{Site Feedback (\ie{} information received from the environment)}} \\
		\midrule
		\task~\ref{chal:what-to-block} & $reqs$ & Set of outgoing requests \\
		\task~\ref{chal:effective-filter-rules} & $hits$ & The filter rules that blocked requests\\
		Sec.~\ref{sec:rewards} &  $C_A$, $C_I$, $C_T$  & Counters for ads, images, text  \\
		\task~\ref{chal:repeat} & Site dynamics & Changes in page content (images, text), ads, and outgoing network requests  \\	
		\midrule
		\multicolumn{3}{l}{\textit{Derived from Site Feedback}} \\
		\midrule
		Sec.~\ref{sec:action} & \actionspace & Hierarchical action space, action = rule \\
		Sec.~\ref{sec:rewards} & $\overline{C}_A$, $\overline{C}_I$, $\overline{C}_T$  & Base representation of site $\ell$ \\
		Eq.~(\ref{eq:counter-norm}) & $\widehat{C}_A$, $\widehat{C}_I$, $\widehat{C}_T$ & Difference between base representation of site $\ell$ and when a filter rule is applied\\
		Eq.~(\ref{eq:breakage}) & $\mathcal{B}$ $\in [0, 1]$& The fraction of page that is broken. 1 = entire page is visually broken\\
		\midrule
		\multicolumn{3}{l}{\textit{Policy: Multi-arm bandits with Upper Bound Confidence}} \\
		\midrule
		Sec.~\ref{sec:policy} & $Q_t(a)$ & Expected reward of action $a$ \\
		Sec.~\ref{sec:policy} & $U_t(a)$ & Confidence of $Q(a)$, higher values mean we should explore action $a$ more. \\
		Sec.~\ref{sec:policy} & $c > 0$ & Exploration factor for UCB \\
		Sec.~\ref{sec:policy} & $\alpha \in (0, 1]$ & Learning step size for updating $Q(a)$  \\
		Sec.~\ref{sec:policy} & $Q_0(a)$ & Initial $Q(a)$ for arms \\
		\midrule
		\multicolumn{3}{l}{\textit{Trade-off: Blocking ads \vs{} Avoiding breakage} (Fig.~\ref{fig:tradeoff-example}) } \\
		\midrule
		Sec.~\ref{sec:rewards} & $\widehat{C}_A$ & Blocking ads, 1 = blocked all ads \\
		Sec.~\ref{sec:rewards} & $1-\mathcal{B}$ & Avoiding breakage, 1 = no breakage \\
		Sec.~\ref{sec:rewards} & $w \in [0, 1]$ & User preference to avoiding breakage \\
		Eq.~(\ref{eq:reward-function-new}) & \rewardfunc{}  & Reward function \\
            Sec.~\ref{sec:filter-lists-top10k} & $\sum \mathcal{B}$ & Potential collateral damage (global rule) \\
		\midrule
		\multicolumn{3}{l}{\textit{Effectiveness of Filter Rules} (\task~\ref{chal:effective-filter-rules}, Fig.~\ref{fig:tradeoff-example})} \\
		\midrule
		Eq.~(\ref{eq:reward-function-a})  & Bad & Rules that don't block ads\\
		Eq.~(\ref{eq:reward-function-b})  & Potential & Blocked ads below $w$ threshold\\
		Eq.~(\ref{eq:reward-function-c})  & Good & Blocked ads w/in. $w$ threshold\\
	    \bottomrule
	\end{tabularx}
	\vspace{-5pt}
	\caption{\small Notations and their references within the paper.}
	\label{tab:notation-summary}
\end{table}

\begin{figure}
	\centering
	\includegraphics[width=1\columnwidth]{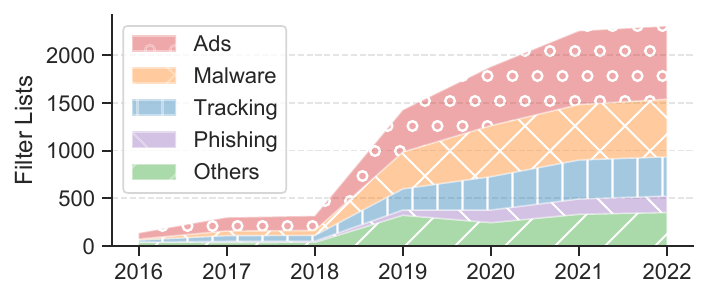}
	\caption{\small \textbf{Filter Lists over Time.} We illustrate the growth of open-source \fl{s} between 2016--2022, categorized by different purposes like ads, malware, and tracking. The most popular purpose is to block ads. }
	\label{fig:filterlists-per-year}
    \vspace{-5pt}
\end{figure}

\begin{figure*}
	\centering
	\includegraphics[width=1\linewidth]{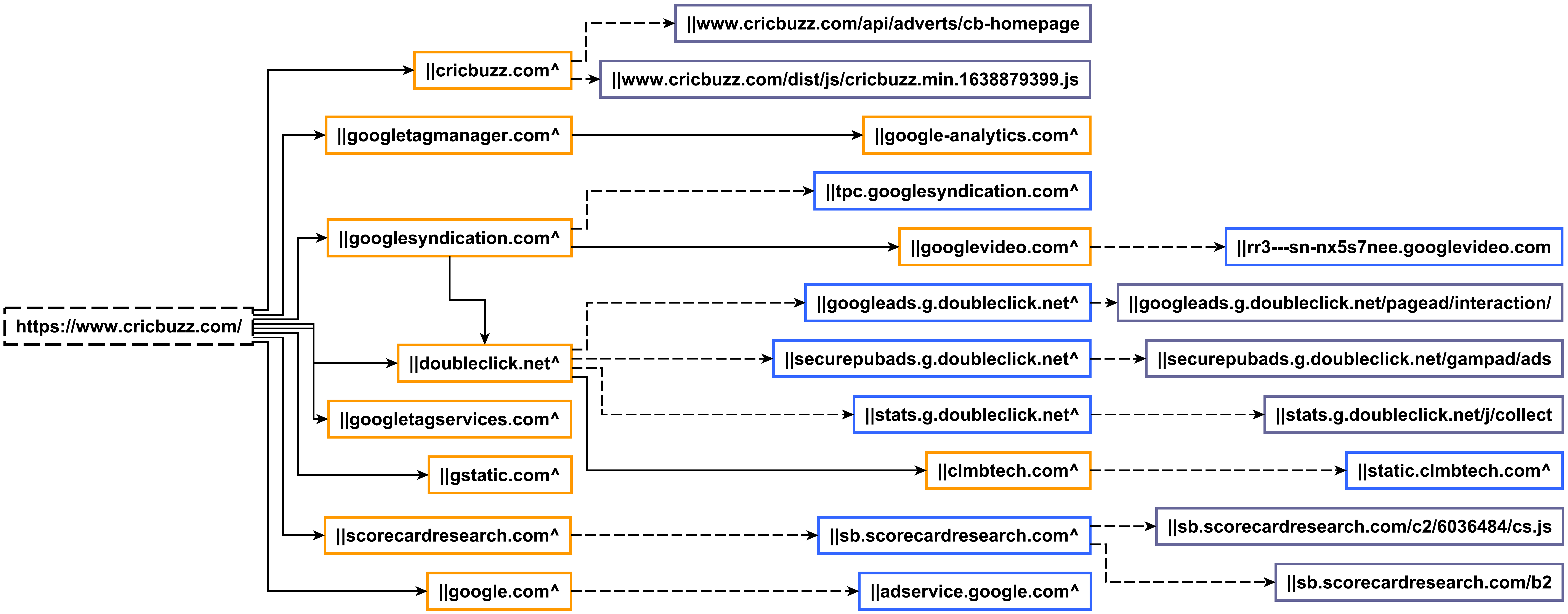}
	\caption{\small \textbf{Action Space \actionspace{}.} We illustrate an truncated example of \actionspace{} with a root (dashed border). Solid $\rightarrow$ are initiator edges. Dashed $\dashrightarrow$ are \finergrain{} edges between eSLD to FQDN filter rules, etc... \textcolor{orange}{Orange}, \textcolor{blue}{blue}, and black solid borders represent eSLD, FQDN, and With Path filter rules, respectively. They align directly with Table~\ref{tab:network-filter-rule-syntax}. As shown in Fig.~\ref{fig:top5k-misc}, action spaces can be quite large.}
	\label{fig:action-space}
\end{figure*}

\section{Filter Lists}
\label{app:filter-list-over-time}

\descr{Filter Rules.} Filter rules for the web (on desktop and mobile) have grown out of the need to block ads and tracking (\ats{}), without causing breakage, as they affect user-facing software (\eg{} browsers and apps). They typically block at the HTTP-level and are distributed to millions of users using web extensions and built-in browser adblocking.
There are different filter rules for blocking A\&T. Some block outgoing network requests based on URLs, while others hide HTML elements (such as DOM elements) and abort JS execution. 

Beyond browsers, OS-wide filter rules are increasingly adopted on mobile \cite{ikrammobile,shuba2018nomoads} and smart TV \cite{varmarken2020tv}.
In addition, DNS-based lists like Pi-Hole~\cite{pihole-homepage}, AdGuard Home~\cite{AdGuardHome}, and NextDNS \cite{NextDNS}, block based on hostnames. They can be applied network-wide; IP-based rules are compatible with firewall software like pfBlockerNG~\cite{pfBlockerNG}.

\descr{Filter Lists.} Filter rules are compiled into open-sourced {\em \fl{s}} (FL) that are maintained by domain experts or individual power users such as Peter Lowe~\cite{PeterLowe}. These lists are curated for different languages, devices, and purposes. For example, the most popular \fl{} for blocking ads, EasyList~\cite{easylist}, has 22 language-specific versions, while AdGuard provides a mobile-specific \fl{} to block ads~\cite{adguardmobilelist}. 
Similarly, the most popular tracking \fl{} for the web is EasyPrivacy~\cite{EasyPrivacy}.

\descr{Filter Lists over Time.} To motivate our focus on URL-based filter rules \major{to block ads} by \tool{}, we conduct a longitudinal analysis of \fl{s} over the years. 
We discover that filter rules for blocking ads are the most prevalent. To do so, we analyze \textit{filterlists.com}, a repository of open-source \fl{s}~\cite{AllFilterLists}. Using Wayback Machine~\cite{Wayback}, we extract information about the \fl{s} within the past six years (up to when the site was established). Fig.~\ref{fig:filterlists-per-year} depicts the growth of \fl{s} and categorized to showcase their Top--4 purposes: ads, malware, tracking, and phishing. Other purposes include blocking annoyances (\eg{} popups and cookie consent dialogs), crypto-mining scripts, and social tracking. By 2022, there are over 2.1K different \fl{s} for 42 diverse purposes that are supported by 44 software tools (\eg{} Adblock Plus (ABP), uBlock Origin, Pi-Hole, AdGuard Home, Opera). 
In addition, when we consider popular tools based on their compatibility with filter lists, we observe that 80\% of the Top--10 tools all support URL-based filter rules.

\section{\tool{} in a Controlled Environment}

In this section, we provide additional details on the implementation of \tool{} that complements Sec.~\ref{sec:autofrg-tool}. \appref~\ref{app:site-done-loading} describes how we detect when a site is finished with loading its content and ads. \appref~\ref{app:action-space} details how we build our action space \persite{}. \appref~\ref{app:build-site-snapshots} explains how we build site snapshots. 
Lastly, Fig.~\ref{fig:autofr-impl} illustrates an example of how \tool{} works end-to-end within a controlled environment.

\subsection{Knowing When a Site is Done Loading}
\label{app:site-done-loading}
We explain how we can provide a best-effort approach when visiting a site, which affects our Alg.~\ref{alg:autofrg-algorithm} every time we visit a site for real.
A challenge to ad-related web measurements is knowing when sites are finished loading to end the visit (including when ads are done loading). Prior work commonly imposes a time limit to the visit for all sites~\cite{iqbal2018adgraph, cvinspectorndss}. 

We address this challenge by using heuristics to minimize and adapt the waiting time necessary for each site. Specifically, using Selenium~\cite{Selenium}, we access the browser's performance logs~\cite{PerformanceLogs}, which are events for life cycles for network requests. 
In addition, we enable ``Page.setLifecycleEventsEnabled'' from Chrome DevTools Protocol to consider the events from the page's (and iframes) life cycle~\cite{PageLifeCycle}. 
We wait until there are less than four events within one second before injecting JS to capture counters of ads, images, and texts. This approach is similar to how other tools (\eg Puppeteer~\cite{puppeteer-networkidle}) know when a page is done loading. If it exceeds 45 seconds, we timeout to deal with sites with automatic videos playing.

\subsection{Building the Action Space}
\label{app:action-space}

We provide additional details on how we construct the action space, as discussed in Sec.~\ref{sec:agent-impl}, and showcase a big-picture example in Fig.~\ref{fig:action-space}.
It is constructed from the set of outgoing network requests collected from the \textsc{Initialize} procedure (Alg.~\ref{alg:autofrg-algorithm}), which comprises multiple visits to the site. We collect the outgoing network requests using Selenium~\cite{PerformanceLogs}. 
Its edges are based on rule dependencies from Sec.~\ref{sec:action}.

\descr{Finer-Grain Dependency.} For every request, we use tldextract~\cite{tldextract} to convert the request into its eSLD, FQDN, and With Path formats (if applicable), and then convert them into URL-based filter rule syntax as shown in Table~\ref{tab:network-filter-rule-syntax}. 
We add edges from eSLD $\rightarrow$ FQDN $\rightarrow$ With Path. We refer to these edges as ``\finergrain{}'' edges. 

\descr{Initiator Dependency.} We further improve \actionspace{} by considering the ``initiator'' of each request. We follow the intuition that if only \textit{ads.com} initiates requests to \textit{bids.com}, then the agent should try \textit{ads.com} first. We do not need to try \textit{bids.com} if it is effective. Consequently, this makes \actionspace{} taller and reduces the number of arms the agent needs to explore per run of the multi-arm bandit, as described in Sec.~\ref{sec:alg-design}. 

We use the initiator call stack provided by Chrome DevTools Protocol~\cite{network-initiator}, which is already part of the collected web requests. The top of this stack is the script that initiates the request. When no call stack is available, we fallback to the ``documentURL'' of the request, which is the URL of the frame from which the request was made. For example, if the request was made within an iframe, then the documentURL will be the URL of the iframe. Recall that the initiator request should already have three nodes that correspond to it (\eg{} eSLD, FQDN, and With Path nodes), and the request itself should have the same. We then add edges from the initiator eSLD node to the request eSLD node, and so forth for other granular nodes. We refer to these edges as ``initiator'' edges. Finally, for all nodes that do not have a parent, we put the root node as their parent. Note that we avoid adding duplicate nodes and edges that would cause cycles.

\begin{figure}[t!]
\centering
    \subfigure[Ads on \textit{gulfnews.com}]{
		 \includegraphics[width=.46\columnwidth]{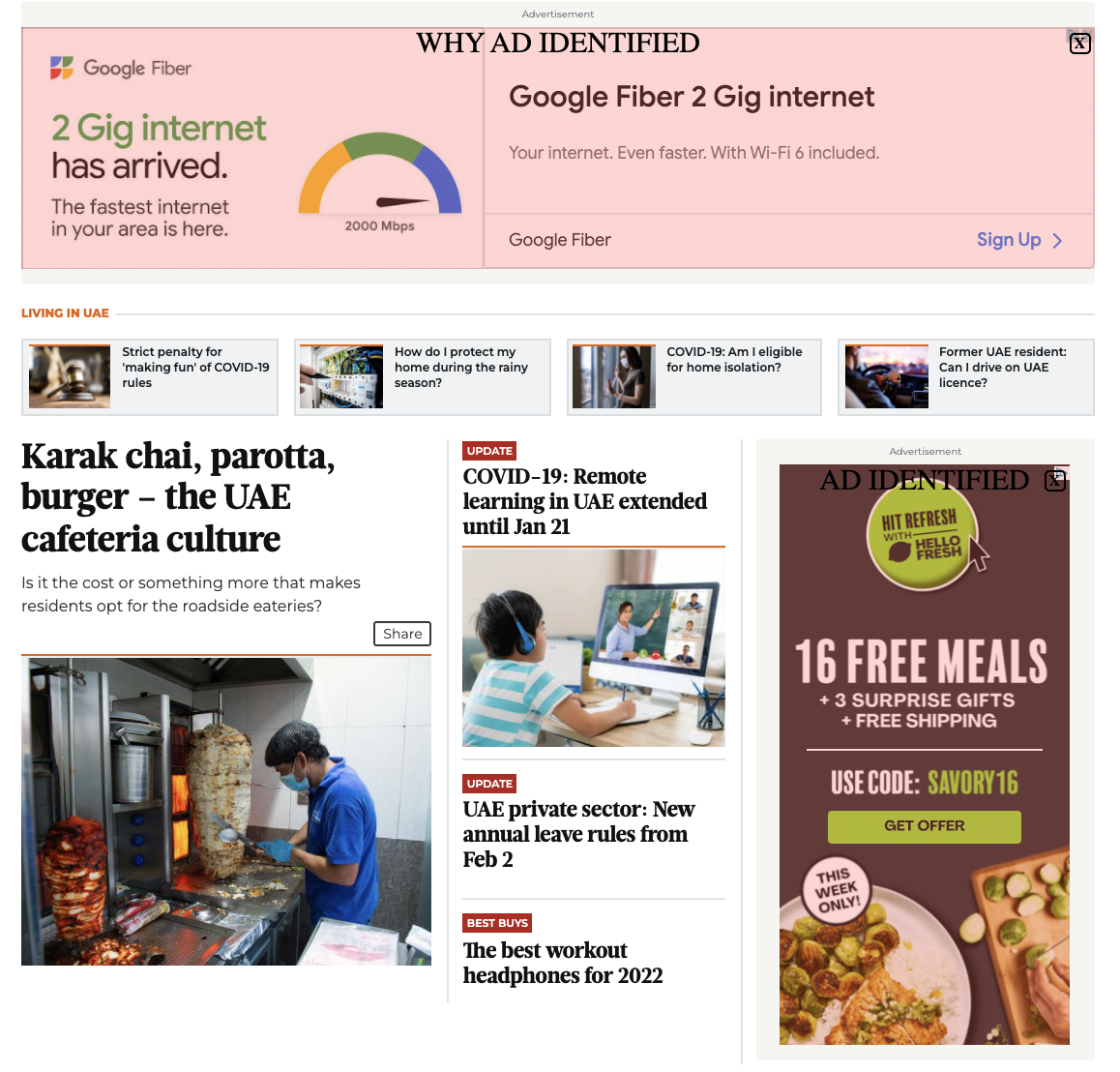}
        \label{fig:ad-high-with-ads}
	}
	\subfigure[Ads on \textit{urdupoint.com}]{
		 \includegraphics[width=.46\columnwidth]{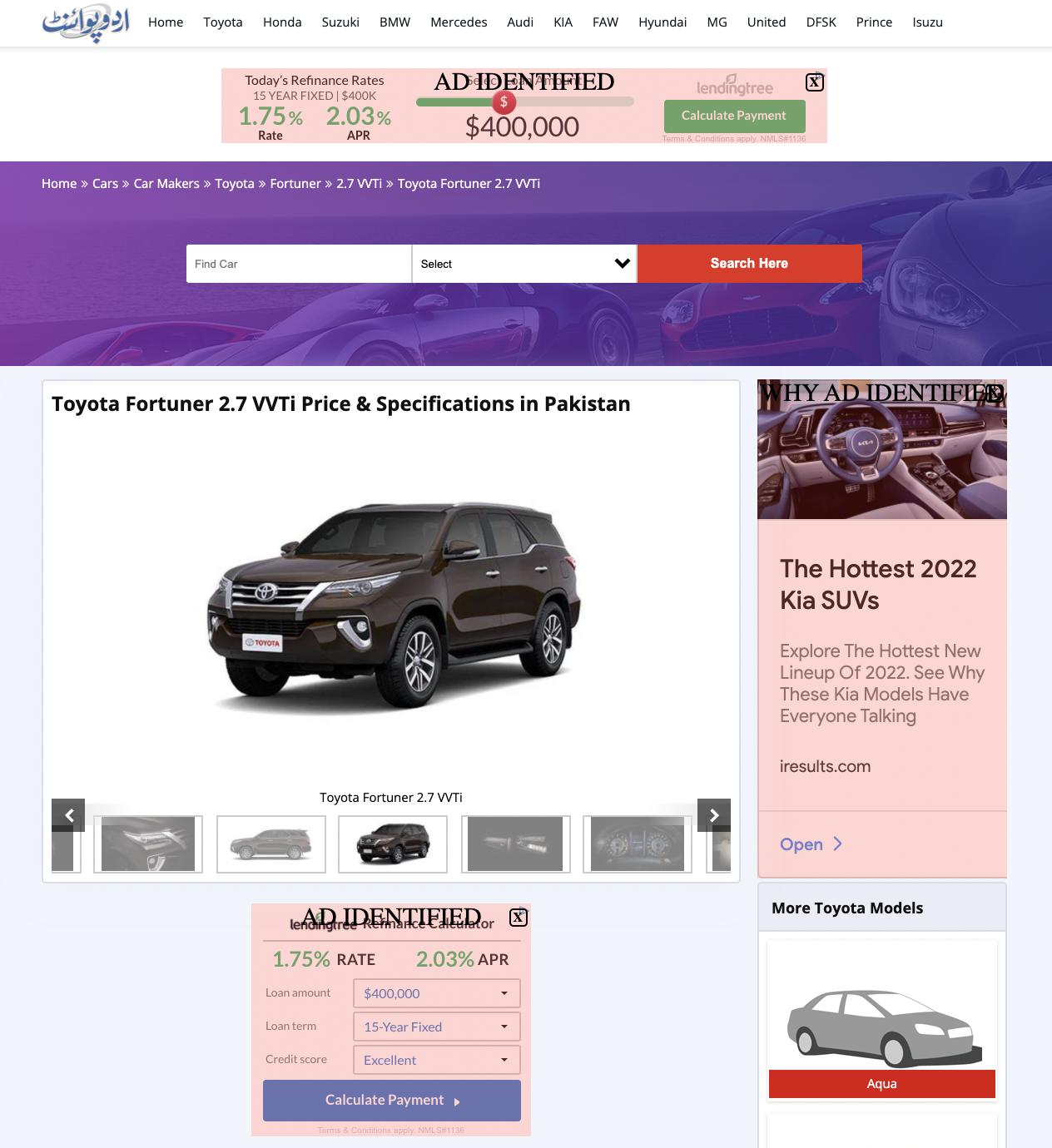}
        \label{fig:ad-high-with-ads-2}
	}

\caption{\small \textbf{\adhigh{}.} We illustrate two examples of how \adhigh{} overlays ads that it identifies using a red background and the label ``Ad Identified.''}
\label{fig:adhighliter}
\end{figure}

\subsection{Building Site Snapshots}
\label{app:build-site-snapshots}
We provide details on how to build site snapshots, which is complementary to Sec.~\ref{sec:env-impl}.

\descr{1. Collecting Raw \adgraph{s}.} We utilize the \adgraph{} browser~\cite{iqbal2018adgraph} with \adhigh{} to determine the number of ads.
We use XVFB~\cite{XVFB} to simulate a virtual display, which makes it possible to run our environment in a headless server (and cloud services). We use a display size of 2560px width and 3240px height, which is wider than the most popular screen size for desktops in 2021~\cite{desktopsize} and three times as tall.
At the end of each visit, we take a screenshot for audibility and save the \adgraph{} file. We repeat this step until we collect the desired number of snapshots, and each must contain at least one ad.

\descr{2. Annotating Site Snapshots.} Recall that we are only concerned with the visual components of a site, such as ads, images, and texts ($C_A$, $C_I$, $C_T$). However, \adgraph{} does not know which nodes are visible to the user, and thus we must annotate nodes that are important. 
We build upon the methodology in \appref~\ref{app:visit-site-real}, where we simply capture the counters and return them. Here, we need to make \adgraph{} know specifically where we got those counters from.
This annotation is illustrated in Fig.~\ref{fig:sitesnapshot}.
The core idea is to use JS to make changes to the HTML so that \adgraph{} can capture them, allowing us to associate site snapshot nodes with three attributes ``FRG-image=true'', ``FRG-textnode=true'', or ``FRG-ad=true''. This annotation is illustrated in Fig.~\ref{fig:sitesnapshot} with filled backgrounds. 

For \textit{visible images}, we add a new attribute ``FRG-image=true'' to the HTML element. This translates to a new node in our site snapshot that references the image node, allowing us to now mark the image node (within the site snapshot) with ``FRG-image=true'' as well.

For \textit{visible text}, since they are not HTML elements, we cannot add an attribute to them. We devise a different approach. We find that each \adgraph{} node keeps track of its previous sibling node in the HTML structure. Therefore, for every visible textnode that we identify using JS, we add a new HTML ``<FRG-TextNode>'' element right after the textnode. This causes \adgraph{} to pick this up as a new site snapshot node with a reference to the previous sibling node (the actual textnode we want). We post-process the raw file to mark the real textnode within the snapshot with ``FRG-textnode=true'' using the previous sibling node information as a guide.

For \textit{visible ads}, annotating is much more challenging because \adhigh{} overlays an ad iframe inside the iframe and not the actual HTML tag itself. The core idea is to figure out if the iframe content contains this overlay and then annotate the top-level element (its tag) as an ad. However, there are several limitations: (1) JS injection cannot access content within third-party iframes as it is sand-boxed (first-party cannot access content within the iframe and vice versa); and (2) ads are complex and can have many nested iframes and the \adhigh{} overlay can be deep inside one of the nested iframes. To address these limitations, we utilize the following methodology. First, we start at the top level (site) and find all visible iframes. Second, we conduct a depth-first search traversal of these iframes using Selenium's ``switchTo'' feature that allows us to switch into the context of individual iframes. Third, once inside the context of each iframe, we inject JS to find the \adhigh{} overlay element (bypassing the limitation from before). If the element is found, we mark the very top iframe element with ``FRG-ad=true'', otherwise, we continue with the traversal.

\descr{3. JS Dependency.} \adgraph{} does not keep track of how scripts depend on each other. For instance, if JS script A calls a method from JS Script B before adding an HTML element, it will only capture JS Script A $\rightarrow$ DIV. Recall that this information is available in the initiator call stack of a web request in \appref~\ref{app:action-space}. Thus, we transfer this information to the site snapshot as well. 
This relationship is shown in Fig.~\ref{fig:sitesnapshot}: JS Script B $\rightarrow$ JS Script A $\rightarrow$ DIV.

\descr{4. Save the Site Snapshot.} We use NetworkX~\cite{NetworkX} to save site snapshots as ``.graphml'' files to be used by \tool{}.

\subsubsection{Site Snapshots \vs{} Action Spaces} 
\label{app:sitesnapshot-vs-actionspace}
Although both site snapshots (\appref~\ref{app:build-site-snapshots} and Fig.~\ref{fig:sitesnapshot}) and action spaces (\appref~\ref{app:action-space} and Fig.~\ref{fig:action-space}) are graph-based, they contain different information and are constructed differently. First, site snapshots represent one visit to a site and capture more than just information about network requests, \eg{} HTML structure and JS script usage. Thus, there can be multiple site snapshots for a given site. Site snapshots have no notion of eSLD and FQDNs, they only consider the entire URLs. Conversely, action spaces are constructed from the set union of network requests based on multiple visits to a site, which results in one action space \persite{}. Nodes are created by converting URLs to different granular filter rules, as described in Table~\ref{tab:network-filter-rule-syntax}. In addition, action spaces do not have any information about HTML structure and JS.

\section{\tool{} in a Live Environment (\toolwild{})}
\label{app:autofrg-wild}

\begin{figure}[t!]
	\centering
	\includegraphics[width=1\columnwidth]{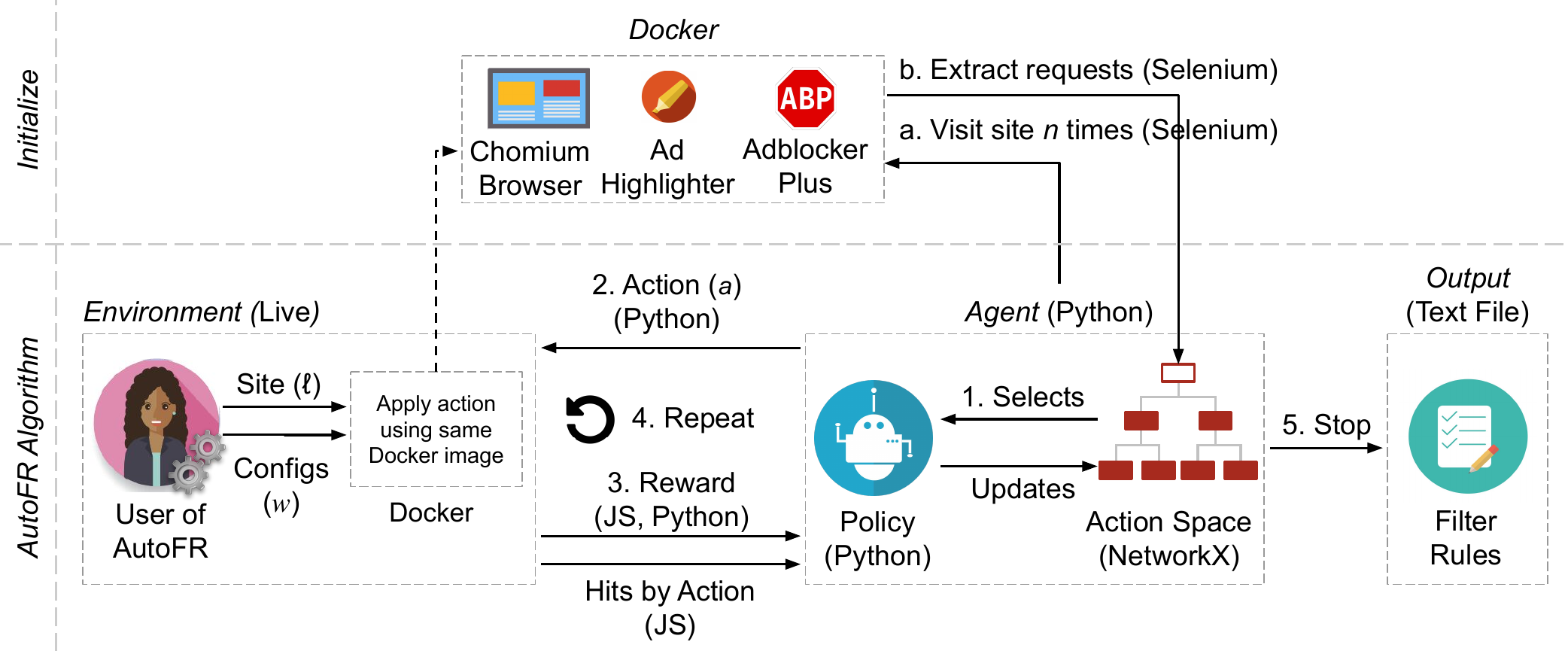}
	\caption{\small \major{\textbf{\toolwild{} Example Workflow (Live Environment).} \textsc{Initialize} (a--b, Alg.~\ref{alg:autofrg-algorithm}): (a) spawns $n=10$ docker instances and visits the site until it finishes loading; (b) extracts the outgoing requests from all visits and builds the action space. We run the RL portion of \textsc{AutoFR} procedure (steps 1--4). Lastly,~\tool{} outputs the filter rules at step 5, \eg{} \textit{$||$s.yimg.com/rq/darla/4-10-0/html/r-sf.html}. Note that we do not use \adgraph{} or site snapshots in this version.}}
	\label{fig:tool-wild}
    \vspace{-5pt}
\end{figure}

In this section, we provide the full details of how we implemented \tool{} in a live setting, referred to as \toolwild{} (\ie{} it visits the site for real during the initialize phase or when an arm is pulled).
\appref~\ref{app:preamble-c-vs-r} explains the differences between \tool{} (that runs using site snapshots) \vs{} \toolwild{}. In \appref~\ref{app:realtime-impl}, we detail the implementation and evaluate \toolwild{} \vs{} \tool{} using a sample of sites. \major{Fig.~\ref{fig:tool-wild} illustrates an example of how \toolwild{} works end-to-end.}

\subsection{\toolcontrol{} vs. \toolwild{} (Implementation)}
\label{app:preamble-c-vs-r}

This section complements Sec.~\ref{sec:autofrg-tool}, where we presented the implementation of \toolcontrol{} in a controlled environment (\ie based on site snapshots). There, we argued that an implementation of \tool{} that exactly mimics the human process, would need to interact with sites and test different rules in a live environment, which would be slow and expensive, albeit strictly better than the human \fla{} process. In this appendix, we describe the implementation of this live version, which we refer to as \toolwild{}. 
It is worth emphasizing that the distinction between controlled (\ie based on snapshots) and ``live'' in the implementation of \tool{} applies only to the {\em training} phase, \ie during the trial and evaluation of candidate filter rules. Once the filter rules are generated with either version of the implementation, they can be {\em applied or tested} on any site in the wild. This is the case in the evaluation of \toolcontrol{} in the main paper (Sec.~\ref{sec:eval}), as well as in the evaluation of \toolwild{} in \appref~\ref{app:control-vs-realtime}.

Fig.~\ref{fig:tool-wild} outlines how we implement \tool{} in a live environment (\toolwild{}). This means that it visits the site for real at every time step $t$ of the algorithm.
It corresponds to our formulation of the problem in Fig.~\ref{fig:formulation}. To simplify our explanation, we follow the same outline as Sec.~\ref{sec:autofrg-tool}. 

\begin{figure*}[t!]
\centering
	\subfigure[\toolwild{} (Sample--100)]{
		 \includegraphics[width=.66\columnwidth]{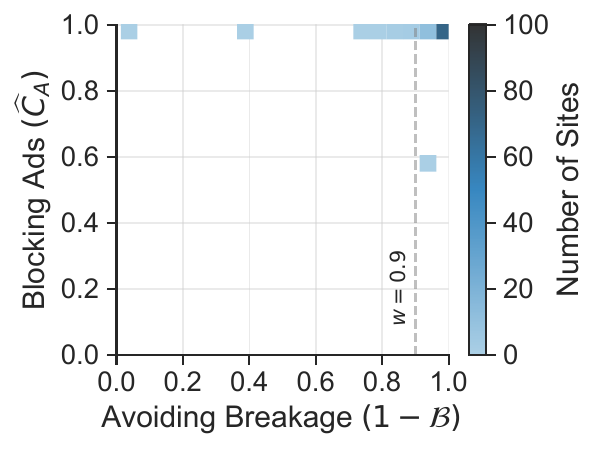}
        \label{fig:realtime-autofrg-100}
	}
    \subfigure[\toolcontrol{} (Sample--100)]{
		 \includegraphics[width=.66\columnwidth]{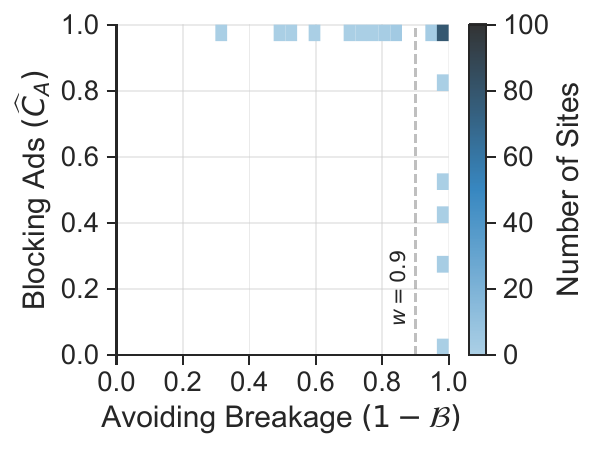}
        \label{fig:control-autofrg-100}
	}
	\subfigure[EasyList (Sample--100)]{
		 \includegraphics[width=.66\columnwidth]{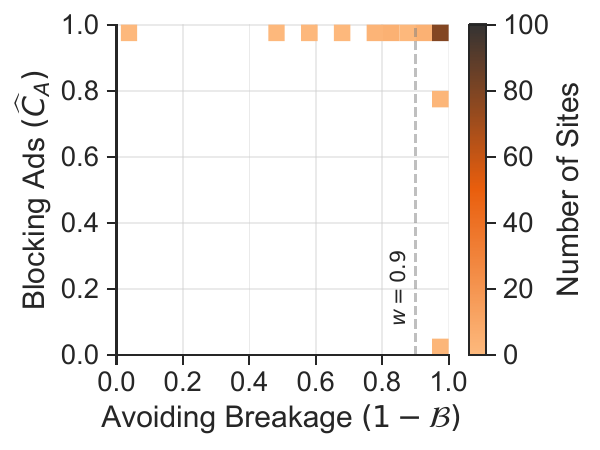}
        \label{fig:easylist-100}
	}
\caption{\small \textbf{\toolwild{} \vs{} \toolcontrol{}.} We sample 100 sites from our \wdata{} and compare the results between \toolwild{}, \toolcontrol{}, and EasyList. \toolcontrol{} performs slightly better than \toolwild{}: 86\% \vs{} 83\%, for sites within $w$ threshold of \toolcontrol{} and \toolwild{}, respectively.
Fig.~\ref{fig:tradeoff-example} explains how to read these plots. See Table~\ref{tab:autofrg-r-vs-c}, col. 1--3, for additional information.}
\label{fig:autofrg-control-vs-autofrg-realtime}
\end{figure*}

\begin{table*}[th!]
	\footnotesize
	\centering
	\begin{tabularx}{\linewidth}{l l X X X }
	    \toprule
	    & & 
	    \multicolumn{3}{c}{\appref~\ref{app:control-vs-realtime}, Fig.~\ref{fig:autofrg-control-vs-autofrg-realtime} Sampled--100} 
	    \\
	    \cmidrule(lr{1em}){3-5} 
	     &  &
        \rotatebox{0}{\parbox{1.7cm}{\textbf{\toolwild{}} \\(Sampled)}} & 
        \rotatebox{0}{\parbox{1.7cm}{\textbf{\toolcontrol{}} \\(Sampled)}} & 
        \rotatebox{0}{\parbox{1.7cm}{\textbf{EasyList} \\(Sampled)}} 
        \\
        \cmidrule(lr{1em}){3-5}
         & \parbox{5cm}{\textbf{Description ($w=0.9$)}} & 1 & 2 & 3 
        \\
		\midrule
        1 & \parbox{7cm}{Sites in \bestbin:  
        $\widehat{C}_A \geq 0.95$, $1-\mathcal{B} \geq 0.95$} 
        & 73\% (73/100) & 80\% (80/100) & 79\% (79/100)
        \\
        \midrule
        2 &\parbox{7cm}{Sites within $w$: $\widehat{C}_A > 0$, $1-\mathcal{B} \geq 0.9$}  
        & \textbf{83\%} (83/100) & \textbf{86\%} (86/100) & 85\% (85/100) 
        \\
        \midrule
        3 & \parbox{7cm}{Ads blocked within $w$:
        $\sum_{\ell} (\overline{C}_A \times \widehat{C}_A)$ / $\sum_{\ell} \overline{C}_A$; $1 - B \geq 0.9$} 
        & \textbf{83\%} (320/387) & \textbf{86\%} (321/375) & 86\% (324/375) 
        \\
	    \bottomrule
	\end{tabularx}
	\vspace{5pt}
	\caption{\small \textbf{\toolwild{} \vs{} \tool{}.}   We provide the same information as Table~\ref{tab:all-results} but for the comparison between \toolwild{} and \tool{} in \appref~\ref{app:control-vs-realtime}. Note that row 3 has different totals because \toolwild{} was run as a different experiment, while \toolcontrol{} and EasyList is a subset of the results from Sec.~\ref{sec:autofrg-control-rules}.
	}
	
	\label{tab:autofrg-r-vs-c}
\end{table*}

\subsection{\toolwild{} Implementation}
\label{app:realtime-impl}

\descr{Agent.} The implementation of the agent, policy, and action space is the same as Sec.~\ref{sec:agent-impl} with details in \appref~\ref{app:action-space}.

\descr{Environment (Live).} The environment allows the agent to apply an action in a live setting. In particular, it visits a site $\ell$ for real and applies the filter rule using Adblock Plus~\cite{AdblockPlus}. It then captures the necessary site feedback (\eg{} ads, images, text, hits) using JS injection and calculates the reward, and returns it back to the agent. A visit to a site for real is explained in \appref~\ref{app:visit-site-real}. Importantly, it has a high cost and we deem it impractical, as discussed in Sec.~\ref{sec:autofrg-tool}.

\subsubsection{\adhigh{}}
\label{app:adhigh}

As discussed in Sec.~\ref{sec:adhigh}, we rely on \adhigh{}~\cite{futureofadblocking} to capture the number of ads during a visit to a site; a real-life example is shown in Fig.~\ref{fig:adhighliter}.
\adhigh{} works in the following ways. Within every iframe, it finds all images and SVGs, or HTML elements that contain the background-url style (\eg{} spans, a tags, and divs), and calculates their 625-bit image hash. It calculates a Jaccard similarity score between the image hash and the set of known hashes (hard-coded), if the similarity is above 0.8, it marks it as an ad by overlaying it with the word ``AdChoice Identified.'' We modify \adhigh{} to listen to a custom event so that we can extract the number of ads it identified using JS injection. \adhigh{} is easily extendable using JS. It also allows us to audit its effectiveness visually using a browser. In addition, the similarity matching threshold is easily tuned to improve the precision of the tool. \adhigh{} has high precision; we explain this in \appref~\ref{app:support-eval}.

\subsubsection{Visiting a (Live) Site}
\label{app:visit-site-real}
Visiting a live site involves the same setup as described in Sec.~\ref{app:build-site-snapshots} when collecting raw \adgraph{s}. However, we use the Chromium browser this time with \adhigh{} and a customized ABP (taken from~\cite{cvinspectorndss}). We use Selenium to toggle off any \fl{s} that are by default loaded. As a result, ABP will start off with no filter rules loaded.

\descr{Applying Action $a$.} To apply our action (filter rule), we do the following methodology as ~\cite{cvinspectorndss}. First, we use flask, a python web server to locally serve our custom filter rules. We then utilize a customized ABP extension with the browser~\cite{cvinspectorndss}. This allows us to load the custom filter rules being served by flask. In addition, we can retrieve the filter rules that blocked any outgoing requests (\ie{} hits). To do so, we trigger a custom event that ABP responds to. ABP will add the $hits$ information in JSON format into the body of the page. We then retrieve it from the page by injecting JS using Selenium.

\descr{Capturing Site Feedback.} 
To capture the number of ads ($C_A$), we rely on \adhigh{}, as described in Sec.~\ref{sec:adhigh}.
For $C_I$, we inject custom JS to retrieve images, similar to ~\cite{cvinspectorndss}. It considers all visible images (\eg{} with height and width $>$ 2px and opacity $>$ 0.1) or HTML elements that have a background-url set~\cite{backgroundurl}. Similarly, for $C_T$, we inject custom JS to find all visible textnodes, which are locations of texts and not the individual word count~\cite{TextNodes}. This allows us to deal with dynamic sites that can serve personalized content. For example, a news site can display the same layout for five articles, each article contains one title and one description. The articles may change upon separate visits, but our approach allows us to still retrieve the same number for text (\eg{} $C_T = 5 \times 2$) without worrying about the change in the content itself. Using Selenium, we are able to collect the outgoing network requests ($reqs$) from the browser, including the initiator information, a call stack that connects how scripts call each other and also which script initiated the request, as discussed in \appref~\ref{app:action-space}.

\descr{Rewards.} We calculate rewards with python, as explained in Sec.~\ref{sec:rewards}.

\subsubsection{Evaluating \toolwild{} vs. \toolcontrol{}}
\label{app:control-vs-realtime}

This section provides evaluation for \toolwild{} that complements Sec.~\ref{sec:autofrg-control-rules}.
\toolwild{} runs Alg.~\ref{alg:autofrg-algorithm} in a live environment by visiting the site for real at each iteration of the multi-arm bandit. It learns directly from the response of the site, while \toolcontrol{} can only infer the response through the use of site snapshots (Sec.~\ref{sec:env-impl}). Our objective is to show that Alg.~\ref{alg:autofrg-algorithm} can be implemented with a live environment, \toolwild{}, and to explore how \toolwild{} differs from \toolcontrol{}.

\descr{Real-world Scenario.} Since \toolwild{} learns in a live environment, we opt for a more realistic experiment setup. 
Recall in Sec.~\ref{sec:alg-design} that the agent traverses the action space \actionspace{}, using one run of multi-arm bandit per layer of the hierarchy. We consider the scenario where the agent is a \fla{} and has only one hour per run of multi-arm in our Alg.~\ref{alg:autofrg-algorithm}; thus, treating time as a budget. The number of iterations within the multi-arm bandit run is now dependent on how long a visit to the site takes. Once the hour expires, the agent moves on to the next multi-arm bandit run, if necessary.

\descr{Experiment Setup.}
Due to the high cost of \toolwild{}, we rely on sampling 100 sites from our \wdata{} in Table~\ref{tab:top-site-results}. 
We denote this as the \sampledata{}.
For each site, we apply \toolwild{} using the same parameters as \appref~\ref{app:params}. Recall that this entails the agent visiting the site for real. Once \toolwild{} is done, we evaluate its filter rules using the automated process in \appref~\ref{app:support-eval}.

\descr{\toolcontrol{} \vs{} \toolwild{} (In the Wild).} Our results are displayed in Fig.~\ref{fig:realtime-autofrg-100} for \toolwild{}, Fig.~\ref{fig:control-autofrg-100} for \toolcontrol{}, and Fig.~\ref{fig:easylist-100} for EasyList, including additional results in Table~\ref{tab:autofrg-r-vs-c} columns 1--3, respectively. Overall, we find that \toolwild{} achieves the same visual pattern of effectiveness as \toolcontrol{} and EasyList. It can block the majority of sites within the $w$ threshold. Using Table~\ref{tab:autofrg-r-vs-c}, we find 73\% of sites fall into the \bestbin{} (row 1, col. 1) and 83\% of sites are within the threshold $w$ (row 2). \toolwild{} and \toolcontrol{} blocks similar proportions of ads: 83\% \vs{} 86\% of ads from \toolwild{} and \toolcontrol{}, respectively (row 3, col. 1 and 2).

We observe that \toolwild{} creates \realtimeruletotal{} filter \vs{} 62 filter rules from \toolcontrol{} for the \sampledata{}. The rules from \toolwild{} cover 67\% (42/62) of rules from \toolcontrol{}. This is not surprising as \toolwild{} is limited by the time budget. As a result, it does not have time to converge to a more precise (or minimal) set of effective rules. For instance, for large action spaces, it may only have time to select certain arms once and will not learn that some arms are non-critical to blocking ads for the site, as discussed in Sec.~\ref{sec:alg-design}. Although, as noted, this did not stop \toolwild{} from being effective overall. Lastly, we find that 42\% (195/\realtimeruletotal) of rules are within EasyList, 41\% (189/\realtimeruletotal) are in EasyPrivacy.

\begin{figure}[t!]
	\centering
	\includegraphics[width=0.66\columnwidth]{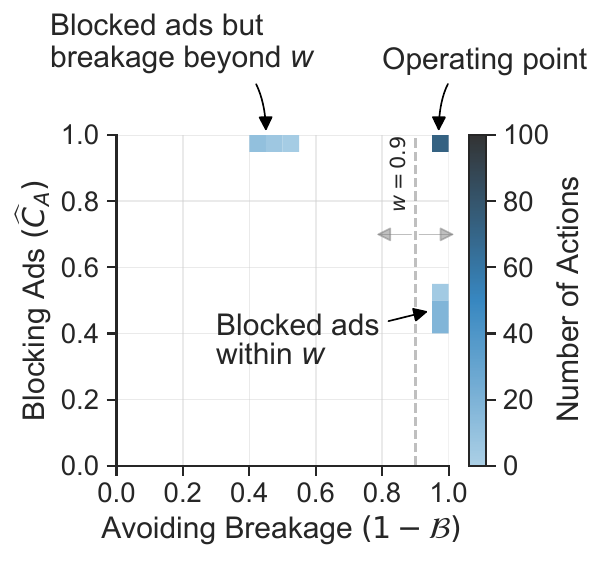}
	\caption{{\small \textbf{Trading-off  Blocking Ads vs. Breakage, via Parameter $w$.} This figure depicts a heatmap of rules applied on a site and the ($\widehat{C}_A$, $1-\mathcal{B}$) they achieve. It is desirable to select rules that block many ads within the breakage threshold $w$ ($1-\mathcal{B}\ge w$); ideally, filter rules close to the top-right corner \bestbin{} ($\widehat{C}_A=1$, $1-\mathcal{B}=1$). We set the threshold $w=0.9$ as an example. Rules to the left of $w$ ($1-\mathcal{B} < w$) block ads but cause more breakage than is tolerable, while ones to the right of $w$ ($1-\mathcal{B}\ge w$) are acceptable.}}
	\label{fig:tradeoff-example}
    \vspace{-5pt}
\end{figure}

\section{Evaluation}
\label{app:eval}

This section provides additional details, results, and evaluation that complement Sec.~\ref{sec:eval}.
\major{
\appref~\ref{app:per-site-eval} continues our evaluation of \tool{'s} \persite{} rules.
}

\subsection{Filter Rules Evaluation Per-Site Cont'd}
\label{app:per-site-eval}
This section is complementary to Sec.~\ref{sec:autofrg-control-rules}.
In \appref~\ref{app:params}, we discuss our parameter selection and experiment setup. \appref~\ref{app:top-5k-cont} and \appref~\ref{app:support-eval} provide additional evaluations of \tool{} in the wild. \appref~\ref{app:eval-annotation} describes how we validated the capturing of images and text. Lastly, in \appref~\ref{app:effects-of-w}, we discuss how $w$ affects the output of \tool{}.

\subsubsection{Parameter Selection for \tool{}}
\label{app:params}

\descr{Selecting $\mathbf{w}$.}
We select $w=0.9$ to represent a user who has similar interests to \fla{s} but has a slightly higher tolerance for breakage. This user wants filter rules that block ads with minimal breakage. We further explore how changing $w$ affects the output of \tool{} in Sec.~\ref{sec:autofrg-control-rules} and \appref~\ref{app:effects-of-w}.

\descr{Hyper-parameters Selection.}
As explained in Sec.~\ref{sec:alg-design}, we have several hyper-parameters that need to be tuned. We list the choices of these hyper-parameters as follows:  
\begin{itemize}[leftmargin=*,noitemsep,]
    \item \textit{Initial estimates ($Q_0$)}: We use the optimistic initial value approach for MAB~\cite{sutton2018reinforcement}. Every filter rule may block ads if the MAB selects it. However, we do not want to go too far above zero, as the rules that are ``potentially good'' need to converge near zero (see Sec.~\ref{sec:alg-design}). Hence, we chose $Q_0=0.2$ as an initial value. This allows every filter rule to be tested by the MAB agent. 
    \item \textit{Learning rate ($\alpha$)}: We use an adaptive learning rate $\alpha=\frac{1}{N[a]}$ to update the $Q(a)$ values. Here, $N[a]$ is the number of times the action $a$ has been selected. We adopted this approach over a constant learning rate to capture the fact that rules can vary in their effectiveness. For example, on one extreme, if the rule $googlesyndication.com$ gets $r=1$ for the first $10$ pulls by the MAB but then does not work at all on the $11th$ pull ($r=-1$), then $Q(a)$ would be dramatically affected with a high constant learning step. 
    \item \textit{Exploration rate for UCB ($c$):} We set the exploration rate for UCB $c=1.4$, to encourage \tool{} to explore the arms without prolonging the convergence of the algorithm greatly (\eg{} $c=2$ causes the convergence to take twice as long).
\end{itemize}

\begin{figure}[t!]
	\centering
	 \includegraphics[width=0.9\columnwidth]{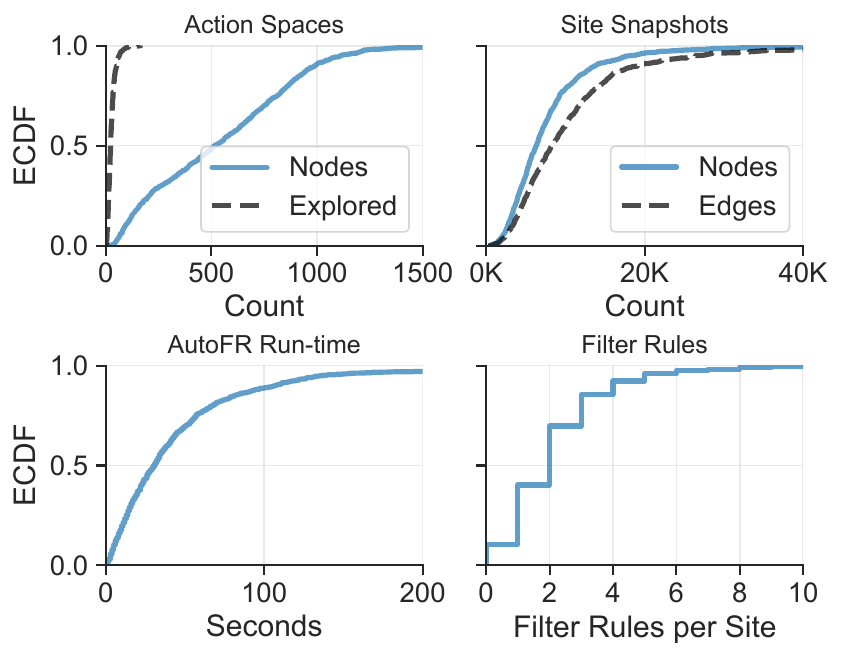}
	\vspace{-10pt}
	\caption{\small \textbf{\toolcontrol{} Information.}: \textit{Action Space (top--left):} 75\% of action graphs have 800 nodes or fewer. \toolcontrol{} only needs to explore a fraction of the action space to find effective rules. \textit{Site Snapshots (top--right):} 75\% of site snapshots contain 10K nodes or fewer. \textit{\toolcontrol{} Run--time (bottom--left):} 75\% of sites take a minute or less to execute the multi-arm bandit portion of Alg.~\ref{alg:autofrg-algorithm}. \textit{Filter Rules (bottom--right):} For 75\% of sites, \toolcontrol{} generated three filter rules or fewer.}
	\label{fig:top5k-misc}
     \vspace{-5pt}
\end{figure}
\begin{figure}[t!]
\centering
	 \includegraphics[width=0.6\columnwidth]{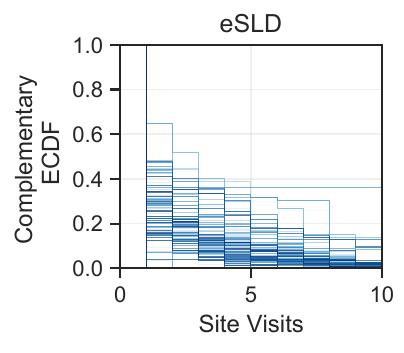}
	\vspace{-10pt}
	\caption{\small \textbf{Site Dynamics based on eSLDs.} We show the number of eSLDs we capture with multiple visits to each site; here, each line represents one site (for a sample of 100 sites from the Top--5K). By the fifth visit, we see that we already capture about 80\% of eSLDs (from the set union of all eSLDs by the \inittimes{} visits \persite{}) for the majority of sites.}
    \label{fig:site-dynamics}
\end{figure}

\descr{Capturing Site Dynamics.} We capture site dynamics by visiting the site multiple times, as discussed in Sec.~\ref{sec:alg-design} and Fig.~\ref{fig:site-dynamics}. To figure out how many visits we will need, we take motivation from prior work~\cite{cvinspectorndss}, which encapsulated the site dynamics in terms of how many eSLDs are captured after every new visit. To guide our selection, we visit 100 random sites within the Tranco Top--5K for \inittimes{} visits \persite{}. For each site, we looked at how many new eSLDs are captured after every visit incrementally. We find that by the fifth visit, we have already captured 80\% of the possible eSLDs from the 10 visits. Thus, we pick \inittimes{} visits for a site to capture its dynamics, which more than doubles prior work~\cite{cvinspectorndss,Zhu_DisruptAA}.

\descr{Making EasyList Comparable.}
EasyList is a state-of-the-art \fl{} for adblocking on the web~\cite{easylist}. However, it contains rules beyond URL-based filter rules, such as element hiding rules. As a result, we make EasyList more comparable to our URL-based filter rules: we parse the list and utilize delimiters (\eg{} ``\$'', ``$||$'', and ``\textcaret{}'') to identify URL-based filter rules and keep them. As a result, our results in Sec.~\ref{sec:eval} for EasyList are based only on its URL-based rules. 

\descr{Automated Evaluation of Filter Rules for a Site.}
For Sec.~\ref{sec:autofrg-control-rules}--~\ref{sec:generalize-rules}, we will utilize an automated approach to evaluate the effectiveness of filter rules and plot the results in a trade-off figure like Fig.~\ref{fig:tradeoff-example}. We discuss the limitations of this automated approach and confirm our results with an independent visual inspection experiment by the authors in Sec.~\ref{sec:autofrg-control-rules} and \appref~\ref{app:support-eval}. 
\begin{itemize}[leftmargin=*,noitemsep,]
    \item \textit{In the Wild:} If we are testing rules in the wild, we apply rules to a site (for real) \inittimes{} times and capture the site feedback $C_A$, $C_I$, $C_T$, as described in Sec.~\ref{sec:env-impl}. We then average the values and use that to calculate our trade-off terms of blocking ads $\widehat{C}_A$ (Eq.~(\ref{eq:counter-norm})) and avoiding breakage $1-\mathcal{B}$ (Eq.~(\ref{eq:breakage})). 
    \item \textit{Site Snapshots:} If we are testing rules in a controlled environment, we will apply the filter rules to each set of \inittimes{} site snapshots, representing a site, to emulate the visits to the site. We proceed with the same calculations for the trade-off terms. 
\end{itemize}

\descr{Cloud-based Experiments}
For our large-scale experiments in Sec.~\ref{sec:autofrg-control-rules} (Top--5K) and~\ref{sec:filter-lists-top10k} (Top 5K--10K), we use the Tranco list~\cite{LePochat2019,TrancoList}. Since we run our experiments in the US region, we customize the list with popular sites for the US only. Next, we set up \toolcontrol{} and \toolwild{} using Amazon's Web Services (AWS) and EC2. Specifically, we use EC2 instance type \textit{m5.2xlarge} that has eight vCPU, 32GB memory, and 35GB storage. For each site, we make sure that \adhigh{} can detect at least one ad before applying \toolcontrol{} or \toolwild{}. Specifically, we visit each site three times and only consider sites with an average number of ads larger than zero.

\begin{figure}[t!]
	\centering
     \includegraphics[width=0.9\columnwidth]{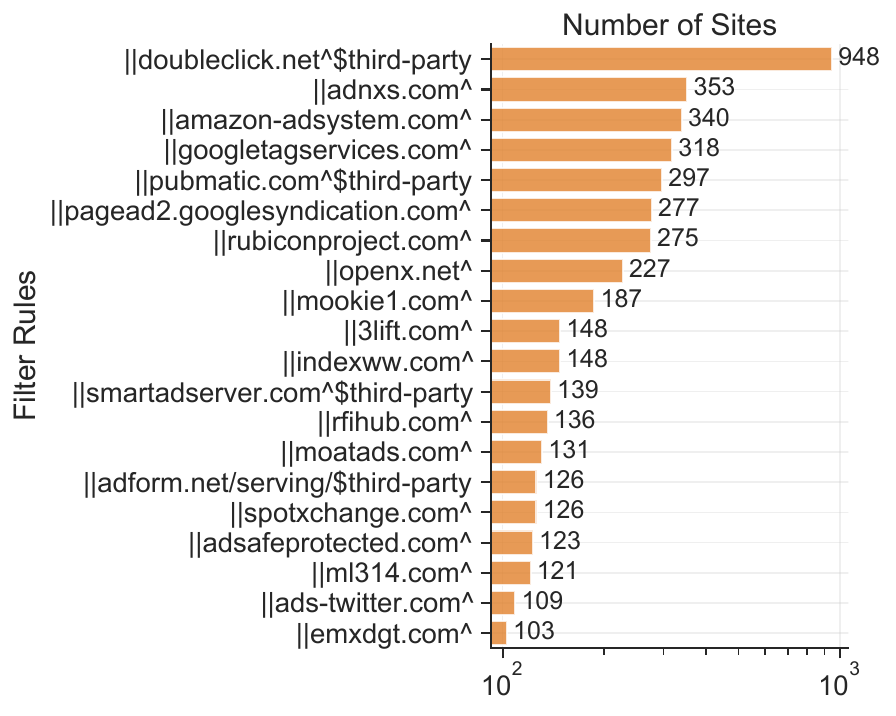}
	\caption{\small \textbf{Top--20 Filter Rules by EasyList for Top--5K Sites.} We apply EasyList to the same Top--5K sites in Sec.~\ref{sec:autofrg-control-rules} and show the popular filter rules by the number of sites that they ``hit'' on (\task~\ref{chal:effective-filter-rules}). This plot is complementary to Fig.~\ref{fig:top5k-rules-control-autofrg}.}
	\label{fig:easylist-top20}
\end{figure}

\subsubsection{Additional Per-Site Evaluation}
\label{app:top-5k-cont}

We provide an additional evaluation of our results from Sec.~\ref{sec:autofrg-control-rules}. \major{In addition, Fig.~\ref{fig:tradeoff-example} illustrates an example of how to read our trade-off plots.}

\descr{Cost of Running \tool{} (Controlled \vs{} Live Environment).}
Recall our discussion in Sec.~\ref{sec:autofrg-tool}, where we estimated that it would take 13 hours to run \tool{} in a live environment and \$1.3 to run in the cloud \persite{}. To explore the cost of running \tool{} in a controlled environment, we leverage our \fulldata{}. We discover that it takes 49 seconds to save a snapshot on average (across 10.4K snapshots). Since each site contains \inittimes{} site snapshots, that takes between 49 to 490 seconds (with or without parallelism using Docker). In addition, it takes 47 seconds on average to run \tool{} after the collection of site snapshots. Altogether, it takes 1.6 to 9 minutes to run \tool{} \persite{}. Compared with our live estimate, the controlled \tool{} is 87 to 488 times faster. In terms of cost, this would cost up to \$0.015 \vs{} \$1.3 \persite{} (pro-rate cost).

\descr{Characterization of Filter Rules.}
\toolcontrol{} generated \ruletotal{} distinct rules for the Top--5K and we feature the Top--20 in Fig.~\ref{fig:top5k-rules-control-autofrg}. 
Furthermore, the filter rules belong to a diverse set of ads and tracking (\ats) domains and resources. To determine whether our rules belong to \ats{} domains, we adopt the approach from prior work~\cite{iqbal2018adgraph,varmarken2020tv} and compare the eSLD version of our filter rules to popular EasyList (advertising) and EasyPrivacy (tracking) \fl{s}~\cite{easylist, EasyPrivacy}. We observe that 31\% (111/\ruletotal) of the rules appear in EasyList. In fact, 70\% of our Top--20 rules in Fig.~\ref{fig:top5k-rules-control-autofrg} also occur in EasyList. When we compare our Top--20 directly to EasyList's Top--20, shown in Fig.~\ref{fig:easylist-top20}, we see an overlap of 40\% in terms of eSLD.
From the remaining rules, 30\% (108/\ruletotal) appear in EasyPrivacy. The appearance of tracking-related rules highlights the relationship between ads and tracking; and we surmise that blocking tracking scripts may stop (or interfere) with the serving of ads (\eg{} $||intergient.com$\textcaret{}, $||t1.daumcdn.net/adfit/adunit\_style/$).

We rely on keywords (\eg{} ``ads'', ``track'', ``bid'', ''advert'') and source code inspection to classify the remaining 142 filter rules. We discover 27\% (38/142) of filter rules are \ats{}. For instance, we find \ats{} domains (\eg{} \textit{$||adhouse.pro$\textcaret{}}), ad resources (\eg{} $||ad.gmw.cn/html/indexbanner\_800\_5.htm$), and \ats{} JS scripts. Some scripts are patently ad-related based on the URL (\eg{} $||www.yourtango.com/prebid\_25062018.js$) and some are not (\eg{} $||www.star\-telegram.com/static/yozons\-lib/core.js$). 
We find 10\% (15/142) may cause functionality breakage, as they target jQuery, Cloudflare's Rocket Loader, and Learn Dash related scripts\footnote{These scripts provide JS features and page loading optimizations.}. However, this is a known limitation since \toolcontrol{} utilizes visual components of a site. We consider the remaining unknown and address this limitation by evaluating the effectiveness of our filter rules in Sec.~\ref{sec:autofrg-control-rules} and \appref~\ref{app:support-eval}.

\descr{Additional Insights.}
Fig.~\ref{fig:top5k-misc} provides insights into different aspects of \toolcontrol{} and Alg.~\ref{alg:autofrg-algorithm}. We observe that the size of action graphs vary across the Top--5K and that their hierarchical structure ensures that \toolcontrol{} only needs to explore a fraction of the action space to find filter rules --- considerably reducing the work necessary for filter rule generation. Next, site snapshots are large graphs which emphasize the complex dynamics of how sites are loaded. However, \toolcontrol{} remains efficient, only needing a minute to execute for 75\% of the sites in the \fulldata. Lastly, we find that \toolcontrol{} generates three filter rules or fewer for 75\% of sites. This is partly due to the dominance of a few players in the ad ecosystem within the Top--5K sites, explained in Sec.~\ref{sec:filter-lists-top10k}.

\descr{Limitations.}
Recall that we have a fixed width and height for our virtual display, as explained in Sec.~\ref{sec:env-impl}. We find that sites that have lazy loading images (\ie{} images that load only within view of the virtual display) and infinite scrolling (\ie{} more content can appear if the user scrolls down) can affect the way we extract the counters of images and texts.  For example, we capture different $C_I$ values when we visit \textit{harpersbazaar.com} without filter rules and with filter rules. This is because blocking ads can bring content upwards and cause lazy loading of images to be displayed. When we compare this with a visual inspection of that site, we find that the rules blocked all ads, and no discernible images were missing. \tool{} treats this unexpected content as breakage, as described in Sec.~\ref{sec:rewards}. This limitation affects how we generate filter rules for \tool{}. In addition, it affects our automated evaluation process unilaterally and does not bias our results towards \tool{} or EasyList for Sec.~\ref{sec:eval}.

A possible way to address this limitation is to scroll down to force all content to load. However, we find that this is a challenge to do consistently so that additional dynamics are not introduced, especially across different sites and across multiple visits to the same site. For example, how do we deal with sites that infinitely load more content as you scroll down? Thus, as a good starting point, we chose to stick with no scrolling while using a relatively large virtual display, as mentioned in \appref~\ref{app:build-site-snapshots}.

\descr{\adgraph{} Limitations.}
We find that \adgraph{} may not attribute images to the JS script that created them when a site uses React~\cite{Fusionjs}. For instance,~\toolcontrol{} created rules that blocked resources \textit{militarytimes.com/pf/dist/components/combinations/default.js}, a JS script for both first-party functionality and ads upon inspection. 
However, our snapshots did not capture that blocking would affect images. 
Thus, upon applying the rule in the wild, we find that it blocked all ads but also blocked legitimate images. 
Using visual inspection, we identify six such sites that used React-related scripts.

\subsubsection{Confirming Results via Visual Inspection}
\label{app:support-eval}

In this section, we provide complementary details to Sec.~\ref{sec:autofrg-control-rules}. 

\descr{Experiment Setup.} We follow the \fla{} approach from \task~\ref{chal:effective-filter-rules} to test the effectiveness of filter rules using visual inspection. From our dataset of \siteruletotal{}, we sample \supportevaltotal{} sites to achieve a confidence level of 95\% with a confidence interval of 5\%. We set up our Chrome browser with \adhigh{} and ABP. We toggle off all \fl{s} within ABP. 
For each site from our sample, we visit it and count the number of ads that \adhigh{} detects and their locations.  We scroll down as far as possible; if the site infinitely scrolls, we limit our count to 10 ads. We then apply the filter rules that \toolcontrol{} generated for that site using ABP, which lets users add custom filter rules. Next, we reload the page and see if ads are still displayed, and repeat the process as before with scrolling. Furthermore, we take note of different kinds of breakage: (1) missing images and text (not exact counts but whether they were noticeably missing); and (2) if the layout of the page is broken, functionality such as infinite scrolling is broken, etc... Once done, we remove the custom filter rules and repeat the process for another site. Importantly, since a human can't notice the exact number of missing images and text, our evaluation is more binary: the rules either blocked all ads or not and whether the rules caused breakage.

\descr{The Effectiveness of \toolcontrol{}.} We provide our results in Table~\ref{tab:all-results} col.~\confirmcol.
Our filter rules can block all ads for 85\% (230/\supportevaltotal) of sites without any breakage (rows 1 and 2). If we only consider breakage caused by missing images and text, this number goes up to 87\% (236/\supportevaltotal). 

\descr{Breakage Analysis.} We observe that 4\% (10/\supportevaltotal) of sites had breakage. For instance, four sites had their layouts broken (although the content is still visible), and one site had the scrolling functionality broken. Recall that this kind of breakage is currently not considered by \tool{}. Next, we found that three sites had some images missing. One of the reasons \tool{} did not detect this was because the images were served with ``$<$amp-img$>$'' instead of the standard ``$<$img$>$'' tag. This can be easily addressed by updating how we retrieve $C_A$ in Sec.~\ref{sec:adhigh}. Lastly, we find two sites like (\textit{gazeta.ru}) that intentionally caused the whole site to go blank after detecting their ads were blocked. Currently, \tool{} does not deal with this type of aggressive circumvention. 

\descr{Blocking Non-Transparency Ads.} \toolcontrol{} can generate filter rules that even block ads that do not have ad transparency logos. We observe that our filter rules could block all ads for 90\% (44/49) of sites that also served non-transparency ads. We surmise this is because a site will use the same approach (or JS) to serve ads with and without ad transparency logos.

\major{\descr{How Precise is \adhigh{} at Detecting Ads?} We count a total of 1040 ads that were detected by \adhigh{}. We found five false positives (\ie{} not ads), giving us a 99\% precision in ads. When we consider it in terms of sites, this affected 2\% of sites (5/\supportevaltotal). False positives can appear due to social widgets like Twitter and SoundCloud with play buttons similar to AdChoice logos. Although we note that this does not always happen for every embedded social widget.}

\begin{figure}[t!]
	\centering
	\includegraphics[width=1\columnwidth]{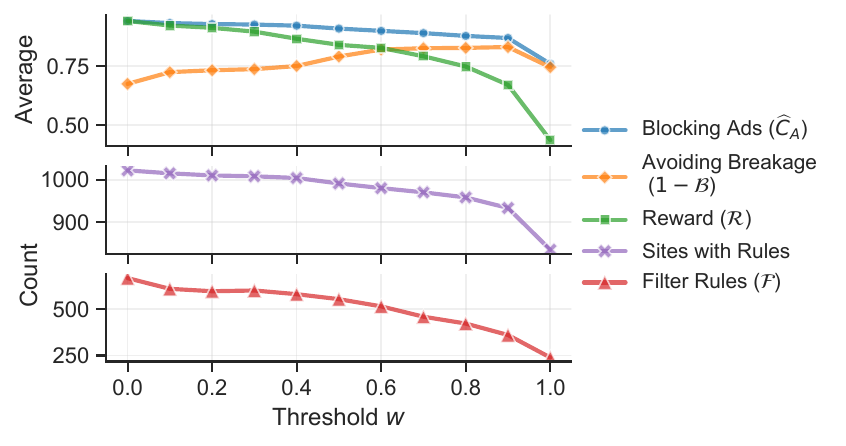}
	\vspace{-10pt}
	\caption{{\small \textbf{Effect of Breakage Threshold $\mathbf{w}$.} We run \tool{} on our entire \fulldata{} using a range of values for $w \in [0, 0.1, ..., 1]$ and we show its effect on the generated filter rules and on the trade-off between ads blocked and page breakage.
	Additional details, for each value of $w$, are provided in \appref~\ref{app:effects-of-w} and Fig.~\ref{fig:app-effects-of-w}.}
	}
	   \vspace{-10pt}
	\label{fig:effects-of-w}
    
\end{figure}

\begin{figure*}[t!]
\centering
	\subfigure[\toolcontrol{}: $w=0$]{
		 \includegraphics[width=.65\columnwidth]{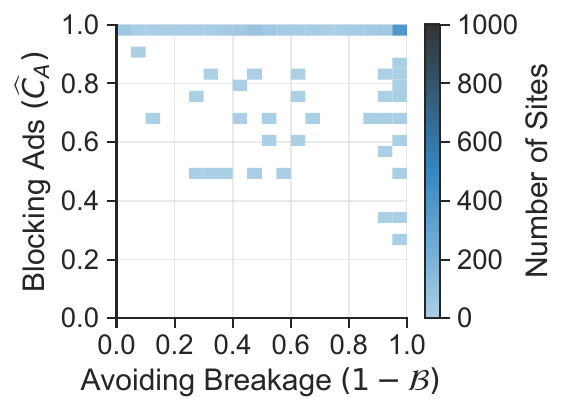}
        \label{fig:effects-of-w-0}
	}
	\subfigure[\toolcontrol{}: $w=0.5$]{
		 \includegraphics[width=.65\columnwidth]{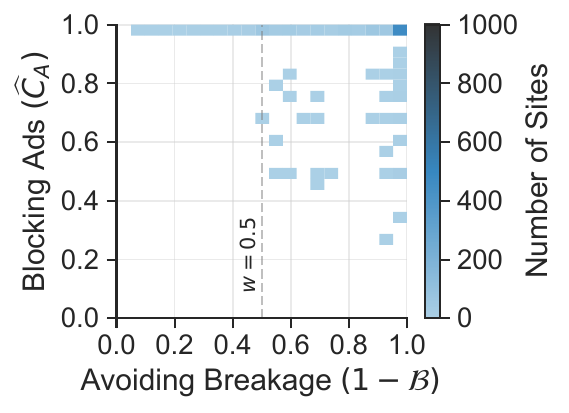}
        \label{fig:effects-of-w-05}
	}
	\subfigure[\toolcontrol{}: $w=1$]{
		 \includegraphics[width=.65\columnwidth]{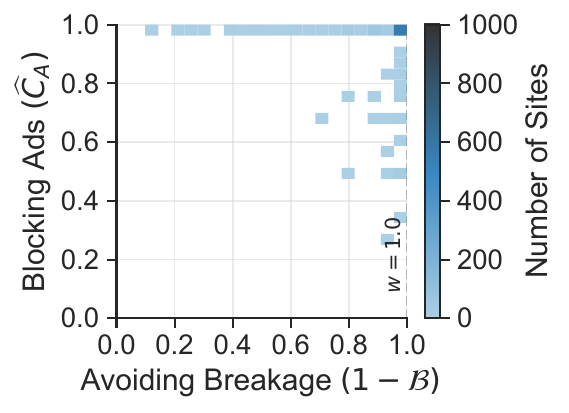}
        \label{fig:effects-of-w-1}
	}
 \vspace{-5pt}
\caption{\small \textbf{\toolcontrol{} across Different $\mathbf{w}$ (Top--5K).} We run \tool{} on our entire \fulldata{} using a range of $w \in [0, 0.5, 1]$ and visualize the effectiveness based on the trade-off of blocking ads \vs{} avoiding breakage. As $w$ increases, there are more sites in \bestbin. Lower $w$ denotes that the user does not care about breakage, which causes less exploration of the action space for rules that fall in the \bestbin{}. }
\label{fig:app-effects-of-w}
\end{figure*}

\subsubsection{Validating the Capture of Visible Images and Text}
\label{app:eval-annotation}
\major{
We validate our methodology of capturing the number of visible images ($C_I$) and text ($C_T$) for a given site, as described in Sec.~\ref{app:visit-site-real}.
To do so, we randomly sample 100 sites from \wdata{} and modify our custom JS in the following ways. For images that we identify, we add a blue solid border; for text, we append ``(AutoFR)''. For each site, we visit the site and inject the modified JS before taking a screenshot. We then visually inspect the 100 screenshots to see whether the images and text were captured; correctly capturing content should be visible to the user and not part of ads. 

We observe our methodology has 100\% precision in capturing visible images and text. This is not surprising, as our methodology relies on common approaches to display images (using $img$ tags, and ``background-url'') and text (we only consider HTML nodes with the type ``TEXT\_NODE''~\cite{TextNodes}).

Next, we evaluate the images and text that were missed. First, we utilize the screenshots to find the locations of visible content that were missed and keep track of their counts. Then, we visit the site using a Chrome browser, inject the JS using the Chrome Developer Tools, and inspect the HTML DOM to determine the reason for missed content. We find that we miss visible images and text because they are rendered using $<svg>$ or pseudo-elements. However, the majority of this missed content is small icons for social media sharing (\eg{} Facebook, Twitter), top menus, and footers. 
If we consider the missed images as false negatives, we get a recall of 95\% for capturing visible images. 
For capturing visible text, we get a recall of 99\%. Future improvements to \tool{} can consider $<svg>$ for $C_I$ by modifying the JS. However, for pseudo-elements, we would need to modify the browser to capture these images. Fortunately, these are often used for non-trivial images such as small icons.
}

\subsubsection{Exploring the Effects of Threshold $\mathbf{w}$}
\label{app:effects-of-w}

\begin{table}[t!]
    \footnotesize
	\centering
	\begin{tabularx}{\linewidth}{p{0.5cm} X}
	    \toprule
		$\mathbf{w}$ & \textbf{Filter Rules for} \textit{womenshealthmag.com}
		\\
		\midrule
        0 & \parbox{7cm}{$||doubleclick.net$\textcaret{}, 
        $||googlesyndication.com$\textcaret{}, \\ $||hearstapps.com$\textcaret{}}  \\
        \midrule
        0.1--0.5 & \parbox{7cm}{$||doubleclick.net$\textcaret{}, \\
        $||hearstapps.com$\textcaret{}}  \\
        \midrule
        0.6--0.9 & \parbox{7cm}{$||doubleclick.net$\textcaret{},
        $||assets.hearstapps.com$\textcaret{},\\
        $||amazon\text{-}adsystem.com$\textcaret{}, $||googletagmanager.com$\textcaret{}}  \\
        
        \midrule
        1 & \parbox{7cm}{$||doubleclick.net$\textcaret{},\\
        $||assets.hearstapps.com/moapt/moapt\text{-}hdm.latest.js$,
        $||assets.hearstapps.com/moapt/moapt\text{-}bidder\text{-}pb.\text{*}.js$, \\
        $||amazon\text{-}adsystem.com$\textcaret{},  $||googletagmanager.com$\textcaret{}}  \\
	    \bottomrule
	\end{tabularx}
	\caption{\small \textbf{Effects of $w$.} We detail how $w$ changes the generated rules for one site, $womenshealthmag.com$. As $w$ increases, some rules will no longer be outputted, such as $googlesyndication.com$ from 0 to 0.1. New rules, like $googletagmanager.com$, may be discovered, from 0.5 to 0.6. While some rules will become more specific, like $assets.hearstapps.com/moapt/\text{*}.js$ rules at 1.}
	\label{tab:effects-of-w-one-site}
	\vspace{-10pt}
\end{table}

\major{
This appendix complements Sec.~\ref{sec:autofrg-control-rules} and further explores the effects of $w$.
 In Fig.~\ref{fig:effects-of-w}, we showed how $w$ affects \tool{}'s performance by using our \fulldata{} \ie{} we applied the filter rules to their corresponding site snapshots.
As the threshold $w$ increases, there is less breakage, and the ad blocking decreases slowly. Ad blocking is the highest at $w=0$: indeed, you can always block ads if you are willing to break the entire page. The reward is also the highest at $w=0$, as there is no constraint on breakage. Once we restrict breakage, we see the trade-off that users must make: higher $w$ (\eg{} $\geq 0.8$) will reduce breakage and the number of rules generated, even causing some sites not to have any rules. Thus, the trade-off also depends on the individual sites. Some sites, like $histats.com$, will have rules no matter the $w$, while others can have no rules after a certain $w$, \eg{} $dailyherald.com$ has no rules after $w \geq 0.5$. This is expected, as \tool{} does not control the effectiveness of a rule on a site but  only learns it.}

\descr{How does the Trade-off Change as $\mathbf{w}$ Increases?} Fig.~\ref{fig:app-effects-of-w} illustrates the trade-off as $w$ increases on the entire \fulldata{} for each individual $w$ value. First, for low $w$'s, we notice more breakage. This is not surprising as the user does not care about breakage. As $w$ increases, we can see that the filter rules adhere to the threshold and mostly stay within it (\ie{} being on the right side of $w$). However, interestingly, we observe that there are more sites that are in the \bestbin{} of the plots (\ie{} the top-right corner). This is because as the user cares more about breakage, \tool{} is exploring more of the action space (\ie going down the hierarchy), and thus more chances of candidate rules that are in the \bestbin{}.

\descr{How do Filter Rules Change as $\mathbf{w}$ Increases?} Next, Table~\ref{tab:effects-of-w-one-site} deep dives into an example of how $w$ changes the output of \toolcontrol{} for one site. First, filter rules can go from being part of the output to no longer part of the output, as shown with the transition of $||googlesyndication.com$\textcaret{} from $w=0$ to $w=0.1$. Conversely, new rules may appear as $w$ increases, as evident with $||amazon\text{-}adsystem.com$\textcaret{} between $w=0.5$ and $w=0.6$. Lastly, we observe that as $w$ increases, rules will be more specific, as shown with the progression of how $||hearstapps.com$\textcaret{} changes from eSLD in $w=0$, to FQDN in $w=0.6$, then to a rule with a path in $w=1$.